\documentclass{vgtc}                          % final (conference style)
\ifpdf%                                % if we use pdflatex
  \pdfoutput=1\relax                   % create PDFs from pdfLaTeX
  \usepackage{graphicx}                % allow us to embed graphics files
  \DeclareGraphicsExtensions{.pdf,.png,.jpg,.jpeg} % for pdflatex we expect .pdf, .png, or .jpg files
\else%                                 % else we use pure latex
  \usepackage{graphicx}                % allow us to embed graphics files
  \DeclareGraphicsExtensions{.eps}     % for pure latex we expect eps files
\fi%

\graphicspath{{figures/}{pictures/}{images/}{./}} % where to search for the images
\usepackage{wrapfig}
\usepackage{microtype}                 % use micro-typography (slightly more compact, better to read)
\PassOptionsToPackage{warn}{textcomp}  % to address font issues with \textrightarrow
\usepackage{textcomp}                  % use better special symbols
\usepackage{amsmath}
\usepackage{amsfonts}
\usepackage{mathptmx}                  % use matching math font
\usepackage{times}                     % we use Times as the main font
\usepackage{cite}                      % needed to automatically sort the references
\usepackage{tabu}                      % only used for the table example
\usepackage{tabularx}
\usepackage{booktabs}                  % only used for the table example
\usepackage{caption}
\usepackage{subcaption}
\usepackage{multirow}
\usepackage[hidelinks]{hyperref}
\usepackage{tikz}
\usepackage{textcomp}
\usepackage{hyperref}
\usepackage{lipsum}

\usepackage[nolist]{acronym}

\newcommand\copyrightnoticeieee{%
\begin{tikzpicture}[remember picture,overlay]
\node[anchor=south,yshift=10pt] at (current page.south) {\fbox{\parbox{\dimexpr\textwidth-\fboxsep-\fboxrule\relax}{\footnotesize \textcopyright 2024 IEEE. Personal use of this material is permitted. Permission from IEEE must be obtained for all other uses, in any current or future media, including reprinting/republishing this material for advertising or promotional purposes, creating new collective works, for resale or redistribution to servers or lists, or reuse of any copyrighted component of this work in other works. DOI: }}};
\end{tikzpicture}%
}

\onlineid{1136}

\vgtccategory{Research}

\vgtcpapertype{algorithm/technique}

\title{GBOT: Graph-Based 3D Object Tracking for \\Augmented Reality-Assisted Assembly Guidance}

\author{Shiyu Li\thanks{e-mail: shiyu.li@tum.de}\\ %   %
\and Hannah Schieber\thanks{e-mail: hannah.schieber@fau.de}
\and Niklas Corell\thanks{e-mail: niklas.corell@fau.de}
\and Bernhard Egger\thanks{e-mail: bernhard.egger@fau.de}
\and Julian Kreimeier\thanks{e-mail: julian.kreimeier@tum.de}
\and Daniel Roth\thanks{e-mail: daniel.roth@tum.de} \phantom{\thanks{\url{https://github.com/roth-hex-lab/gbot}}}\\ %
\and \parbox{2in}{\scriptsize \centering Technical University of Munich\\ School of Medicine and Health\\ Department Clinical Medicine\\Machine Intelligence in Orthopedics\\ Clinic for Orthopedics and Sports Orthopedics$^{\text{*,†,¶,‖}}$}
\and \parbox{2in}{\scriptsize \centering Human-Centered Computing and Extended Reality \\ 
 Friedrich-Alexander-Universität Erlangen-Nürnberg (FAU) \\ Erlangen, Germany$^{\text{*,†,‡}}$}
\parbox{2in}{\scriptsize \centering Lehrstuhl für Graphische Datenverarbeitung (LGDV) \\
Friedrich-Alexander Universität (FAU) Erlangen-Nürnberg \\
Erlangen, Germany$^{\text{§}}$}
}

\teaser{
  \centering
  \includegraphics[width=0.9\linewidth]{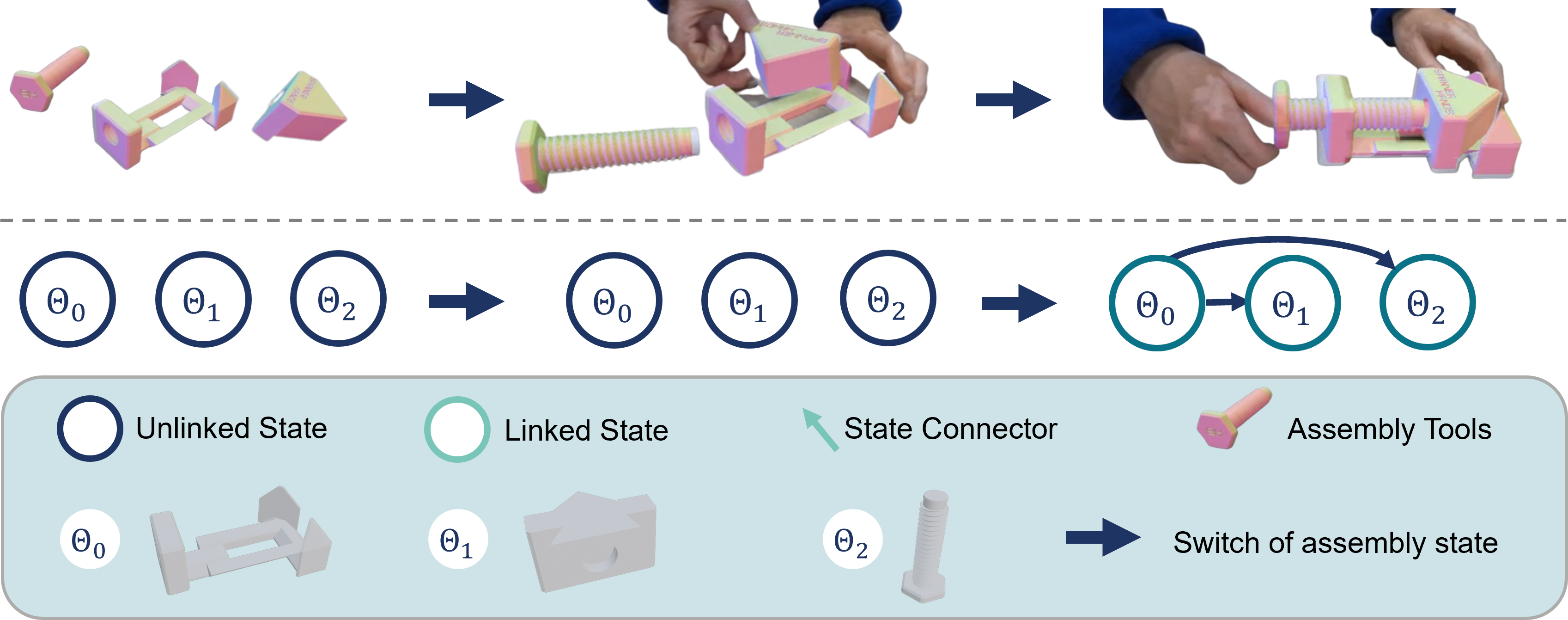}
  \caption{\textbf{
  Our graph-based object tracking algorithm can track objects in their individual assembly states.}
  We apply kinematic links between objects in different assembly states. % The assembly state can be automatically acquired by object poses for \ac{ar}-assisted assembly tasks.
  For augmented reality based assembly tasks, the assembly state can be automatically detected by the constantly tracked object pose.
  }
  \label{fig:teaser}
}

\abstract{
Guidance for assemblable parts is a promising field for augmented reality. Augmented reality assembly guidance requires 6D object poses of target objects in real time. Especially in time-critical medical or industrial settings, continuous and markerless tracking of individual parts is essential to visualize instructions superimposed on or next to the target object parts. In this regard, occlusions by the user's hand or other objects and the complexity of different assembly states complicate robust and real-time markerless multi-object tracking. 
To address this problem, we present Graph-based Object Tracking (GBOT), a novel graph-based single-view RGB-D tracking approach. The real-time markerless multi-object tracking is initialized via 6D pose estimation and updates the graph-based assembly poses. The tracking through various assembly states is achieved by our novel multi-state assembly graph. We update the multi-state assembly graph by utilizing the relative poses of the individual assembly parts. Linking the individual objects in this graph enables more robust object tracking during the assembly process. For evaluation, we introduce a synthetic dataset of publicly available and 3D printable assembly assets as a benchmark for future work. Quantitative experiments in synthetic data and further qualitative study in real test data show that GBOT can outperform existing work towards enabling context-aware augmented reality assembly guidance. Dataset and code will be made publically available\textsuperscript{**}.
} % end of abstract

\CCScatlist{
  
   \CCScatTwelve{Computing methodologies}{Artificial intelligence}{Computer vision}{}
    \CCScatTwelve{Computing methodologies}{Computer graphics}{Graphics systems and interfaces}{Mixed / augmented reality}
}

\usepackage{graphicx}
\usepackage{xcolor}
\definecolor{added}{RGB}{30,144,255}
\definecolor{adapted}{RGB}{200, 100, 100}
\definecolor{spelling}{RGB}{212, 140, 142}

\usepackage{soul}
\soulregister\cite7
\soulregister\ref7
\soulregister\pageref7
\soulregister\section7
\soulregister\subsection7
\soulregister\caption7
\soulregister\acrlong7
\soulregister\ac7
\soulregister\acs7
\soulregister\acp7
\soulregister\figure7

\begin{document}
\begin{acronym}[Bspwwww.]  % Längstes Kürzel in der nachfolgenden
                       % Liste um die Breite der Spalte für die
                       % Abkürzungen zu bestimmen.

%% Eintrag: \acro{Referenzname}[Kürzel]{Langform}
%% Im Text wird die Abkürzung dann mit \ac{Referenzname} benutzt.
%Az
\acro{ar}[AR]{Augmented Reality}
\acro{add}[ADD]{Average Distance Error}
\acro{ate}[ATE]{absolute trajectory error}
\acro{bvip}[BVIP]{blind or visually impaired people}
% C
\acro{cnn}[CNN]{convolutional neural network}
%F
\acro{fov}[FoV]{field of view}
%G
\acro{gan}[GAN]{generative adversarial network}
\acro{gcn}[GCN]{graph convolutional Network}
\acro{gnn}[GNN]{Graph Neural Network}
\acro{gbot}[GBOT]{Graph-based Object Tracking}
%H
\acro{hmi}[HMI]{Human-Machine-Interaction}
\acro{hmd}[HMD]{head-mounted display}
% DoF
\acro{dof}[DoF]{Degrees of Freedom}
\acro{mr}[MR]{mixed reality}
% I
\acro{iot}[IoT]{internet of things}
\acro{icp}[ICP]{Iterative Closest Point}
% L
\acro{llff}[LLFF]{Local Light Field Fusion}
\acro{bleff}[BLEFF]{Blender Forward Facing}

\acro{lpips}[LPIPS]{learned perceptual image patch similarity}
%N
\acro{nerf}[NeRF]{neural radiance fields}
\acro{nvs}[NVS]{novel view synthesis}
% M
\acro{mlp}[MLP]{multilayer perceptron}
\acro{mrs}[MRS]{Mixed Region Sampling}

%O
\acro{or}[OR]{Operating Room}
%P
\acro{pbr}[PBR]{physically based rendering}
\acro{psnr}[PSNR]{peak signal-to-noise ratio}
\acro{pnp}[PnP]{Perspective-n-Point}
%Q
%R
%
\acro{sus}[SUS]{system usability scale}
\acro{ssim}[SSIM]{similarity index measure}
\acro{sfm}[SfM]{structure from motion}
\acro{slam}[SLAM]{simultaneous localization and mapping}

%T
\acro{tp}[TP]{True Positive}
\acro{tn}[TN]{True Negative}
\acro{thor}[thor]{The House Of inteRactions}
%U
\acro{ueq}[UEQ]{User Experience Questionnaire}
%V
\acro{vr}[VR]{virtual reality}
%W
\acro{who}[WHO]{World Health Organization}
\acro{ycb}[YCB]{Yale-CMU-Berkeley}
\acro{yolo}[YOLO]{you only look once}

\end{acronym}

%% The ``\maketitle'' command must be the first command after the
%% ``\begin{document}'' command. It prepares and prints the title block.

%% the only exception to this rule is the \firstsection command
\firstsection{Introduction}
\maketitle
\copyrightnoticeieee
%% \section{Introduction} %for journal use above \firstsection{..} instead
Accurate and real-time estimation of 6D object pose and robust object tracking play a crucial role in enhancing \ac{ar} assembly guidance through immersive 3D visualization. This task becomes especially challenging in dynamic scenarios involving multi-object tracking, where assembly states change and occlusions occur~\cite{zauner_authoring_2003,kleinbeck_artfm_2022,su_deep_2019}. \ac{ar}-guided assemblies can be widely applied to everyday life~\cite{su_deep_2019,wu_augmented_2016}, industrial settings~\cite{eswaran_augmented_2023,radkowski_hololens_2017,bottani_augmented_2019}, or in the medical context~\cite{kleinbeck_artfm_2022}. Different visualization approaches employing optical or video see-through devices, such as \ac{ar} \acp{hmd}~\cite{kleinbeck_artfm_2022,blattgerste_comparing_2017} or smartphones~\cite{yan_augmented_2022,alves_comparing_2019}, have been explored. There is evidence, that \ac{ar} assembly guidance can outperform pictorial instructions \cite{blattgerste_comparing_2017,wu_augmented_2016}, especially in-situ visualizations outperform side-by-side or in-view visualizations \cite{laviola_-situ_2023, blattgerste_-situ_2018}. Furthermore, some approaches tackle hand occlusions \cite{yan_augmented_2022} or authoring instructions \cite{chidambaram_processar_2021}. 

To enable \ac{ar}-based assembly guidance in medical or industrial settings, real-time markerless tracking is one crucial element. To achieve markerless tracking, the utilization of 6D object pose estimation~\cite{wang_gdr-net_2021,peng_pvnet_2019,tekin_real-time_2018,wang_normalized_2019} or 6D object tracking~\cite{prisacariu_pwp3d_2012,stoiber_iterative_2022} is essential. Most deep-learning methods combine object detection or semantic segmentation with a consecutive pose estimation~\cite{peng_pvnet_2019,wang_gdr-net_2021,jantos_poet_nodate}. This consecutive combination can be computationally costly and is often not real-time suitable. 6D object tracking instead is often more efficient but requires an initial pose for each object. Moreover, the focus of tracking or pose estimation is often on a single object ~\cite{stoiber_iterative_2022,rambach_6dof_2018} instead of multiple objects in various assembly states or requires textured objects~\cite{park_multiple_2008}.

To address this challenge, we present \acf{gbot}. A real-time capable graph-based tracking approach enabling multi-object tracking in multi-state assembly tasks. To automatically initialize the tracking prior to the assembly tasks, we harness 6D pose estimation using our YOLOv8Pose. After initialization, the tracking updates the objects' poses constantly. To enable tracking over different assembly steps, we define a graph describing the kinematic links between each assembly pair in different steps. \ac{gbot} can dynamically switch between different assembly states while existing graph-based approaches~\cite{stoiber_multi-body_2022, schmidt2015dart} require re-initialization per state. Each assembly step is defined by the relative pose between the objects. If the individual parts are assembled in their relative pose to each other, the linked parts are considered as a module. The tracking of the module is eased by constraining links which regulate the individual objects' \ac{dof}. This enables a more efficient tracking compared to tracking individual assembly parts. 

To train and validate our approach, we generated synthetic data of 3D assembly assets. The generated validation data contains scenes with standard (normal) conditions, hand occlusion, dynamic light, and motion blur. To address the sim-to-real gap, we recorded real data without ground truth for qualitative evaluation. Our evaluation compares \ac{gbot} with a state-of-the-art tracking approaches \cite{stoiber_iterative_2022, stoiber_srt3d_2022, li2023more} and our 6D pose estimation. Our approach shows robustness and real-time capability for object tracking towards \ac{ar} assembly guidance.  
%In the end of this paper, we shows a simple example of our tracking algorithm using Hololens.
In summary, we contribute:

\begin{itemize}
% \item A graph-based object tracking for multi-state assembly including assembly state identification
\item A real-time multi-object assembly graph tracking driven by 6D pose estimation for multi-state assembly including assembly state identification.
\item An annotated synthetic dataset and unlabeled real test data of publicly available and 3D printable assembly assets as a quantitative and qualitative benchmark for \ac{ar} assembly guidance.
\end{itemize}

\section{Related work}

\subsection{Instance-level 6D Pose Estimation}

% Introduction sentence for 6d pose
6D object pose estimation predicts the six \ac{dof}, namely rotation and translation that define an object's position in a camera coordinate system. Pose estimation can be categorized into two main areas: category-level 6D pose estimation, which involves class-wise pose identification~\cite{wang_normalized_2019,zaccaria_self-supervised_2023} and instance-level 6D pose estimation~\cite{xiang_posecnn_2017,amini_YoloPose_2023,he_ffb6d_2021}, i.e., identifying the pose for each individual instance of a class.

Instance-level 6D object pose assumes that 3D object models are known during inference. Furthermore, instance-level 6D object pose estimation can be distinguished into one-stage~\cite{he_ffb6d_2021,amini_YoloPose_2023,xiang_posecnn_2017} and two-stage~\cite{jantos_poet_nodate,peng_pvnet_2019} approaches. One-stage approaches like PoseCNN~\cite{xiang_posecnn_2017} or FFB6D~\cite{he_ffb6d_2021} rely on semantic segmentation, while SingleShotPose~\cite{tekin_real-time_2018} builds upon the object detector YOLO~\cite{redmon_you_2016}. Using the extracted features, they either regress the pose directly~\cite{xiang_posecnn_2017}, feed extracted points into \ac{pnp}~\cite{tekin_real-time_2018} or utilize Least Squares Fitting algorithm~\cite{he_ffb6d_2021}. Amini et al.~ \cite{amini_YoloPose_2023} introduce YoloPose which directly regresses keypoints in an image and presents a learnable module to replace \ac{pnp}.

Moreover, two-stage approaches~\cite{peng_pvnet_2019,wang_gdr-net_2021} apply an off-the-shelf object detector and subsequently estimate the 6D pose. These methods restrict real-time capability due to computational cost. Depending on the choice of the object detection or semantic segmentation backbone, as well as the algorithm employed for 6D pose estimation, one-stage approaches may exhibit greater efficiency, making them better suited for real-time applications.

\subsection{Object Tracking}

In contrast to 6D pose estimation, object tracking assumes the 6D pose in the initial frame to be known. Object tracking can be divided into region-based~\cite{prisacariu_pwp3d_2012,tjaden_region-based_2018} and depth-based~\cite{comport_real-time_2006,issac_depth-based_2016,choi_rgb-d_2013,pomerleau_review_2015,chen_object_1992,besl_method_1992} approaches. Tracking algorithms can suffer from losing track due to heavy occlusion. To address this, DeepIM~\cite{li2018deepim} introduced the re-initialization by PoseCNN~\cite{xiang_posecnn_2017}.

Region-based methods rely on color images to predict the probability that a pixel belongs to the object or to the background based on statistics. PWP3D~\cite{prisacariu_pwp3d_2012} is the first real-time region-based approach around 20 Hz. 6D object poses are updated based on pixel-wise optimization. Based on this work, Tjaden et al.~\cite{tjaden_region-based_2018} proposed RBOT, which uses a Gauss-Newton optimization scheme to accelerate the optimization. To make the object tracking more efficient, a sparse tracker SRT3D~\cite{stoiber_srt3d_2022} is introduced using second-order Newton optimization with Tikhonov regularization. Their tracking algorithm reaches about 200 Hz considering occlusion without the need of GPU acceleration.

Depth-based methods use depth cameras to track objects. The most common tracking approach is the \ac{icp} framework~\cite{radkowski_hololens_2017,rusinkiewicz_efficient_2001}. Other approaches introduce particle filters~\cite{choi_rgb-d_2013} or Gaussian filters~\cite{issac_depth-based_2016}. 

Besides approaches that only rely on depth or regions, Stoiber et al.~\cite{stoiber_iterative_2022} propose the hybrid approach ICG. ICG considers visual appearance and adds both region and depth modality on the framework reaching around 788 Hz for a single object. Moreover, Stoiber et al.~\cite{stoiber_fusing_2023} extend ICG to ICG$+$ by adding texture modality using visual features like SIFT or ORB. Although this further improves the results, it also extends the computational time. Additionally, Stoiber et al.~\cite{stoiber_multi-body_2022} present a multi-body tracking framework Mb-ICG for kinematic structures. They interpret kinematic structures as graphs and formulate the 6~\ac{dof} pose variations of bodies as the vertices of the graph. Their multi-body tracking with kinematic structures considers objects with different regions defined as fixed and movable but is constrained to one object state, which cannot be applied for tracking assembly parts. For tracking initialization, they manually aligned objects to the predefined poses which limits its real AR application with users.  % with different regions.

\subsection{Datasets for Object Pose and Tracking}

To evaluate 6D pose estimation or tracking algorithms, datasets with manifold objects and various scenarios are crucial. LINEMOD~\cite{hinterstoisser_model_2013}/LINEMOD occluded~\cite{brachmann_learning_2014} and YCB-Video (YCBV) contain textured objects. T-Less~\cite{hodan_t-less_2017} provides texture-less objects which is challenging for feature-based approaches. Most existing datasets~\cite{hinterstoisser_model_2013,brachmann_learning_2014,hodan_t-less_2017,xiang_posecnn_2017} contain multiple objects but do not consider dynamic assembly objects with hand occlusion. Therefore, assembly assets with a subset of assembly parts~\cite{su_ikea_2021,NEURIPS2022_b645d1a0} can be considered as a 6D pose estimation dataset for \ac{ar}-assisted guidance. IKEA-Manual\cite{wang2022ikea} is a assembly dataset with 3D pose of each parts, i.e., rotation of the 3D part but the translation of each parts are not investigated. The IKEA assembly dataset \cite{su_ikea_2021} addresses this but is not publicly available. For hand pose estimation, Schoonbeek et al.~\cite{schoonbeek2024industreal} introduce an assembly dataset. However, it focuses on hand poses instead of object poses.

The RBOT dataset~\cite{tjaden_region-based_2018} focuses on single object tracking and includes a selection of twelve objects from LINEMOD dataset and five from the Rigid Pose dataset~\cite{pauwels_real-time_2013}. RBOT is a semi-real dataset with rendered models on real background images. Regarding tracking of kinematic objects, RTB~\cite{stoiber_multi-body_2022} proposed a semi-real dataset with robotic kinematic objects. Another tracking dataset is OPT~\cite{wu_poster_2017}. It contains single 2D and 3D objects. OPT only contains a single object per frame. In addition, Choi Dataset~\cite{choi2013rgb} introduces four kitchen objects and provides RGB-D data both in synthetic scenes with
the ground truth object trajectories and real scenes for qualitative evaluation. Moreover, BCOT dataset~\cite{li2022bcot} proposed a marker-less high-precision 3D Object Tracking Benchmark using calibrated binocular cameras and object poses are optimized by the re-projection constraints in all views.

 Moreover, the use of synthetic data is common for 6D pose estimation as capturing real-world data is time-consuming and error-prone due to calibration errors. In this regard, Denninger et al.~\cite{denninger_blenderproc_2020} propose Blenderproc, a Blender and python based library to generat synthetic images. Additionally, other tools like Unity or Unreal can also be used to generate synthetic data~\cite{qiu_unrealcv_2016,martinez-gonzalez_unrealrox_2021,borkman_unity_2021,schieber2023indoor}.

\begin{figure*}[t!]
    \centering
    \includegraphics[width=0.85\textwidth]{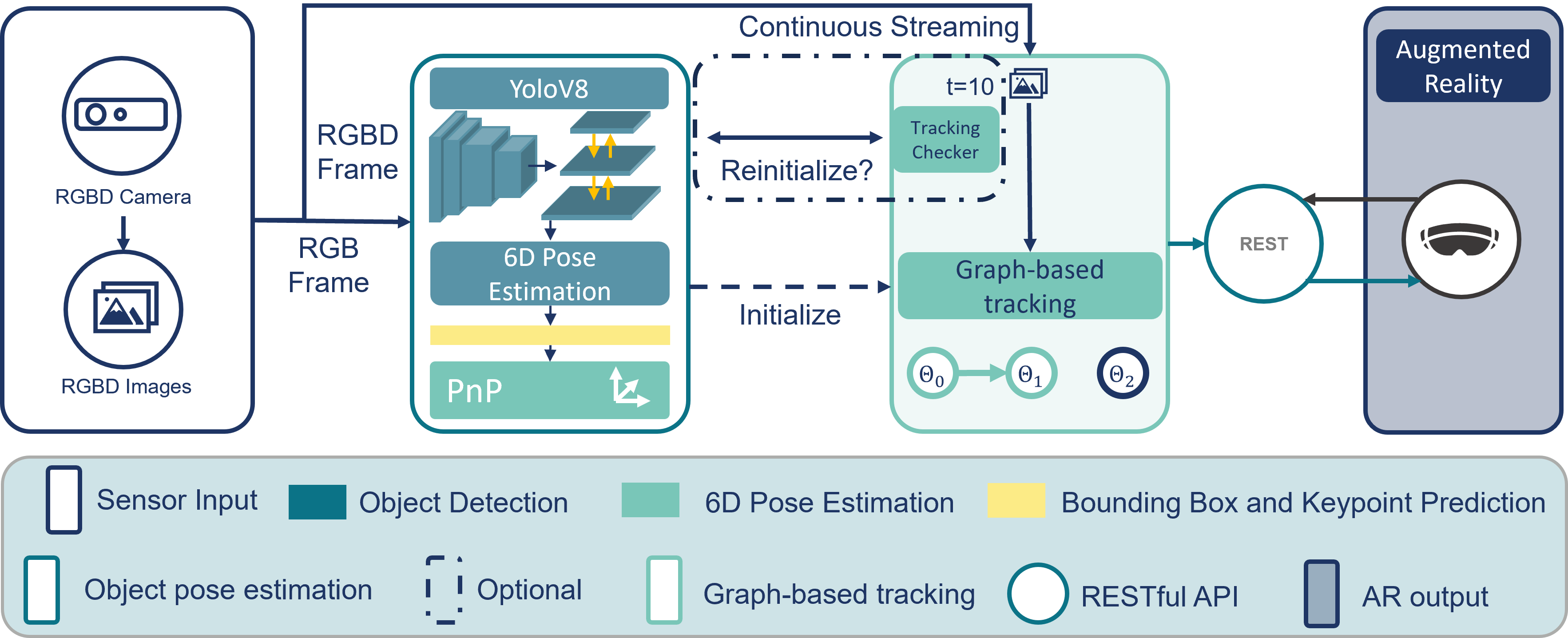}
     \caption{\textbf{\acf{gbot} framework.} We utilize RGB for our initialization and RGB-D to enable our continuous 6D pose object pose tracking. Our YOLOV8pose aquires bounding boxes and keypoints after non maximum suppression and \acf{pnp} recovers the 6D object poses. YOLOV8pose initializes the graph-based tracking. Additionally, we enable GBOT to re-initialize the tracker every 10th frame~\cite{li2018deepim}. The tracker updates the estimated poses and uses links between assembled objects. These poses can be published in real time to an \ac{ar} \ac{hmd} via a RESTful API. }
    \label{fig:pose_architecture}
\end{figure*}

\subsection{Pose Estimation and Tracking in AR}

Utilizing markers~\cite{zauner_authoring_2003}, object detection~\cite{kleinbeck_artfm_2022}, pose estimation~\cite{su_deep_2019}, or tracking~\cite{park_multiple_2008} enables, for example, the visualization of assembly instructions in \ac{ar}~\cite{zauner_authoring_2003,kleinbeck_artfm_2022}. Zauner et al.~\cite{zauner_authoring_2003} proposed hierarchical structures for authoring \ac{ar} assembly instructions. To track the objects, visual markers were used. Rambach et al.~\cite{rambach_6dof_2018} introduced a 6\ac{dof} single object tracking based on 3D Scans for \ac{ar}. 

Moreover, Park et al.~\cite{park_multiple_2008} proposed a method of multi-object tracking. For object tracking they rely on local features requiring textured objects. Additionally, Wu et al.~\cite{wu_augmented_2016} apply template matching and \ac{icp} to track objects in various assembly states. To monitor assembly states they use a tree-like structure, with the 
closing and combining states which define the connecting components of the assembly parts in a graph. However, their template-based approach is limited in the hand occlusion scenes and with the rise of deep learning, this approach has been outperformed by correspondence-based approaches driven by \acp{cnn}. 
Su et al.~\cite{su_deep_2019} proposed a deep-learning-based multi-state object pose estimation for \ac{ar} assembly tasks, focusing on pose estimation of static objects. Additionally, Kleinbeck et al.~\cite{kleinbeck_artfm_2022} utilized synthetic data and deep-learning based object detection to visualize assembly steps. Moreover, Wang et al.~\cite{wang2018active} proposed a tree-like assembly graph for \ac{ar} guidance. However, the object detector does not provide the six~\ac{dof} information for the individual objects and it is not suitable for 3D user interface. Liu et al.~\cite{liu2020tga} train a object detector with an attention module for recognizing objects in multiple scenes. Similarly, Stanescu et al.~\cite{stanescu2023state} train a variation of YOLO to identify object in multiple states using a state vector in their object detector.

%
%%%%% 
% I would write a general summary here, so with CNNs there is effort made for isntance level pose estimation, tracking is more robust and faster and in AR we want all of this but tracking is commonly done by markers or relies on local features
%% to bring this all together one would need to consider .... which is provided by our approach

\section{Methodology}
\ac{gbot} can be split into two parts: First, 6D object pose estimation is used to initialize the tracking approach. Second, we implement multi-object tracking and identify the assembly states using our multi-state assembly graph. If the tracking is lost, GBOT + re-init, can enable re-initialized again by object pose estimation. The overall framework is depicted in \autoref{fig:pose_architecture}. % We apply 6D pose estimation for the first frame of tracking as initialization. Then we use a graph-based tracking approach to track objects in different assembly states.
To train and validate our approach we leverage synthetic data generation. Real data is only used for qualitative tests.

\subsection{Initial Object Pose Estimation}

For our 6D pose estimation, we build upon the state-of-the-art object detector YOLOv8~\cite{jocher_yolo_2023} and due to real-time capability, design it as a one-stager approach~\cite{amini_YoloPose_2023}. To detect the objects, we extend the detection output to not only bounding boxes but also keypoints for 6D pose estimation~\cite{amini_YoloPose_2023}. We detect keypoints directly on the objects surfaces instead the corners of 3D bounding boxes~\cite{amini_YoloPose_2023}. Our architecture is illustrated in \autoref{fig:pose_architecture}. The backbone of the YOLOv8 model is CSPDarknet as feature extractor, which is followed by the C2f module. The C2f module feeds to three prediction heads, which learn to predict the object keypoints for the input image. To ensure solid prediction results for small object, we set the input image size to $1280 \times 720$ px. After the keypoints and bounding boxes are detected, they are fed into RANSAC \ac{pnp} to recover the object pose. 

\subsubsection{Keypoints Selection}

To define surface keypoints on each object, we apply Farthest Point Sampling~\cite{amini_YoloPose_2023,peng_pvnet_2019} which initializes a keypoint set on the object surfaces and adds overall $N$ points. In terms of $N$, three keypoints are the minimum for \ac{pnp}~\cite{marchand_pose_2016}, but this number is not robust with occlusions or symmetric objects limiting the pose estimation. The maximum amount of keypoints used is 24~\cite{amini_YoloPose_2023}. Generally, more keypoints slightly improve the accuracy but also increase the computational cost~\cite{peng_pvnet_2019}. Due to our varying object sizes, we use 17 as an economic trade-off between keypoint quantity, visibility, and computational time.

\subsubsection{6D Pose Prediction}

\ac{pnp} is the problem of solving 6D object pose given a set of $N$ 3D points of object models and corresponding prediction 2D key points. The output of the object detector, i.e., the bounding boxes and keypoints, is processed by RANSAC \ac{pnp} to recover the 6D object poses. \ac{pnp} is not robust to outliers if the key points are not accurate. Therefore RANSAC can be applied with \ac{pnp} to make the pose estimation more reliable. To train the network we apply the loss proposed by YOLOv8 for keypoint regression.~\cite{jocher_yolo_2023}:

\begin{equation}
\begin{split}
L_{total} = \sum_{i,j,k}(\lambda_{cls}L_{cls} + \lambda_{box}L_{box} + \lambda_{pose}L_{pose} + \\ \lambda_{kobj}L_{kobj} + \lambda_{focal}L_{focal})
\end{split}
\end{equation}
where $L_{total}$ represents the total training loss, $L_{cls}$ represents the loss for class labels, $L_{pose}$ represents the key points loss for pose estimation and $L_{kobj}$ represents the keypoint objectiveness loss. Additionally, VariFocal loss~\cite{zhang_varifocalnet_2021} $L_{focal}$ is applied ~\cite{zhang_varifocalnet_2021,jocher_yolo_2023} considering the overlap of the bounding boxes. $\lambda_{cls} = 0.5$, $\lambda_{box} = 7.5$, $\lambda_{kpts} = 12$, $\lambda_{kobj} = 1.0$, and $\lambda_{focal} = 1.5$ are hyperparameters to balance the loss for the individual tasks, i.e., object detection (class and bounding box), and keypoints.

The pose loss is defined as L2 loss between the predicted value and ground truth value~\cite{wang_densefusion_2019,kendall_geometric_2017}. 

\begin{equation}\label{eqn:pose_loss}
{L}_{pose} =|| \pmb{K}_{pred} - \pmb{K}_{gt} ||_2
\end{equation}
where $\pmb{K}_{pred}$ is the predicted 2D keypoints by deep learning model and $\pmb{K}_{gt}$ is the ground truth 2D keypoints.

\subsection{Graph-based Object Tracking}

6D object pose estimation can be used to constantly detect individual objects, but is computationally costly impeding real-time capabilities. Object tracking can provide real-time pose information, but needs a pose for initialization. We use 6D pose estimation for object tracking initialization. A graph-based object tracking is based on updates of object poses at new frames in time~\cite{fan_deep_2022}. Most of the tracking algorithms (e.g., \cite{stoiber_multi-body_2022,tjaden_region-based_2018}) define a probabilistic model based on an energy function or a pose variation vector. We use the energy function defined as the negative logarithmic probability following Stoiber et al.~\cite{stoiber_multi-body_2022}. Our tracking in particular extends their graph-based approach \cite{stoiber_multi-body_2022} which uses kinematic links between different objects to ease the tracking process. In contrast to their work, we update those links in real-time according to the prior known assembly graph. Furthermore, we do not rely on deep learning models detecting assembled parts as new parts \cite{su_deep_2019}. Instead, we specifically consider the spatial combinations of existing objects. \autoref{fig:pose_architecture} shows the structure of our graph-based object tracking pipeline. We use a RGB-D sensor for object detection and tracking providing object poses via a RESTful API. With the API, multiple \ac{ar} visualization devices can request the object poses. % The poses are send to an \ac{ar} \ac{hmd}  for in-situ visualization.

\subsubsection{Multi-State Assembly Graph}

To track objects through different assembly states, we build upon the multi-body kinematic tracking approach from Stoiber et al. \cite{stoiber_multi-body_2022} which is limited to one object state, and extend it to be applicable for assembly parts. Through the integration of the assembly graph following the kinematic link between the individual objects, our approach is able to track assembly parts in multiple states. After each assembly state, there are new links or constraints between objects which reduce the degrees of movement. Occlusions by other parts challenge even state-of-the-art graph-free tracking algorithms. Through the use of an assembly graph, the tracked and connected parts can be treated as one new tracking object. By linking assembled objects in real time to one object, we aim to overcome the tracking limitation of existing approaches with static graphs.

\begin{figure}[t!]
    \centering
             \begin{subfigure}{0.62\columnwidth}
            \centering
            \includegraphics[width=\textwidth]{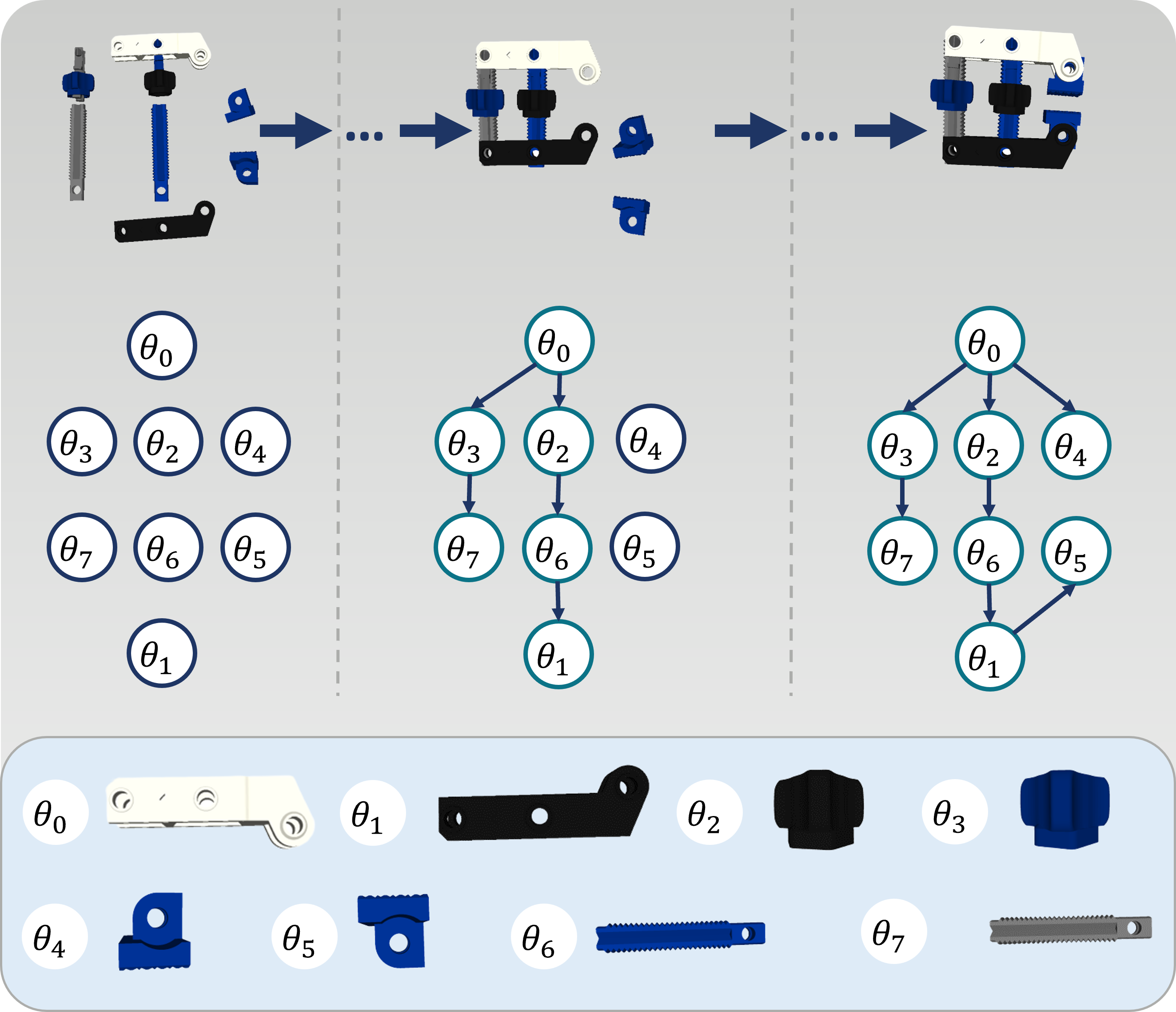}
            \caption{Multi-state Assembly graph}
            \label{fig:state_graph}
        \end{subfigure}
         \begin{subfigure}{0.34\columnwidth}
            \centering
            \includegraphics[width=\textwidth]{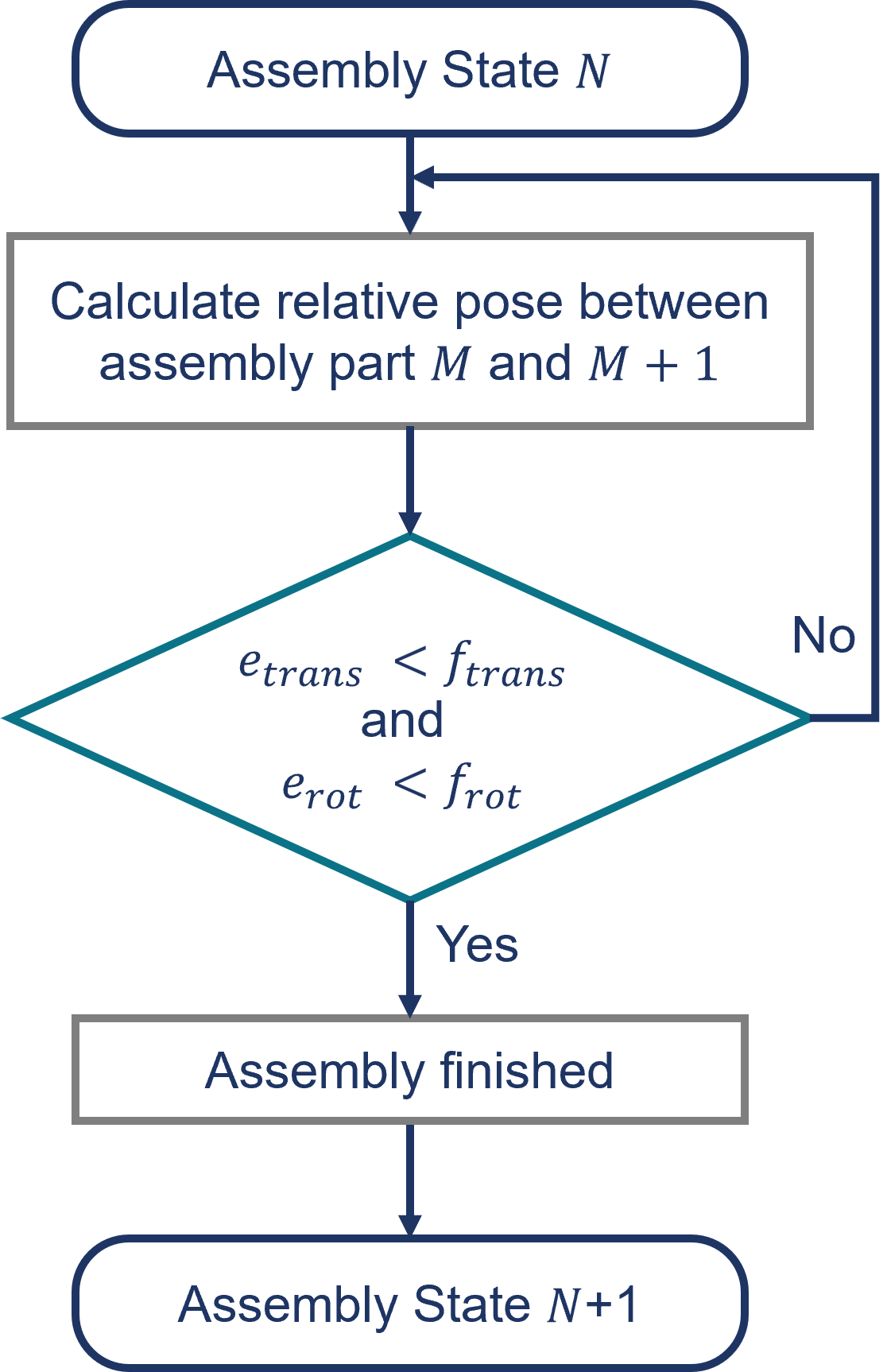}
            \caption{Switch of assembly state}
            \label{fig:state_switch}
        \end{subfigure}
    \caption{\textbf{Multi-state assembly graph and criteria for the switch of assembly states.} If the difference between the calculated relative pose between two assembly parts and the defined ground truth pose is smaller than the offset of translation and rotation, the algorithm will switch to the next assembly state.}
    \label{fig:state_chart}
\end{figure}

For each assembly state, we defined a tree-like assembly state graph, see \autoref{fig:state_graph}. Multi-state assembly graph is defined as multiple tree-like state graph which can be switched in the order of the assembly sequence. In the first state, each part is tracked individually without links. In the end, assembly parts are finished as a complete asset. Our multi-state assembly graph is formulated as:

% \begin{equation}
% \pmb{A}_{t} = 
% \begin{bmatrix}
%     \pmb{T}_{11} & \pmb{T}_{12} & \pmb{T}_{13} & \dots  & \pmb{T}_{1n} \\
%     \pmb{T}_{21} & \pmb{T}_{22} & \pmb{T}_{23} & \dots  & \pmb{T}_{2n} \\
%     \vdots & \vdots & \vdots & \ddots & \vdots \\
%     \pmb{T}_{n1} & \pmb{T}_{n2} & \pmb{T}_{n3} & \dots  & \pmb{T}_{nn} \\
% \end{bmatrix}
% \end{equation}
% where $\pmb{A}_{t}$ is the assembly graph in assembly step t, $\pmb{T}_{ij}$ is the relative pose defined by the transformation matrix between two assembly parts, with $\pmb{T}_{ij} = \pmb{T}_{ji}^{-1} $.
\begin{equation}
{[\pmb{R}|\pmb{t}]}_n =\mathcal{F}_{state}\{\pmb{T}_{ij}, I_n,...,I_{n-k+1},I_{n-k}; {[\pmb{R}|\pmb{t}]}_0| \beta \}
\end{equation}
where  $\pmb{R}$ and $\pmb{t}$ represents rotation and translation in 6D object pose, $\mathcal{F}_{state}$ represents the tracker of current assembly state, $\pmb{T}_{ij}$ is the relative pose defined by the transformation matrix between two assembly parts, $\beta$ refers to the model parameters and $I$ represents RGB-D images.

We defined the relative pose for each assembly pair by a transformation matrix~\cite{stoiber_multi-body_2022}. The individual parts are linked to the base part. Between two assembly parts, we defined the kinematic links which limit the relative movement for the assembly pairs. During the assembly process, the assembly state graph is updated based on 6D pose information. The base part has six~\ac{dof} while the other parts are kinetically constrained and linked to the base part.

\begin{figure*}[th]
    \centering\includegraphics[width=0.9\textwidth]{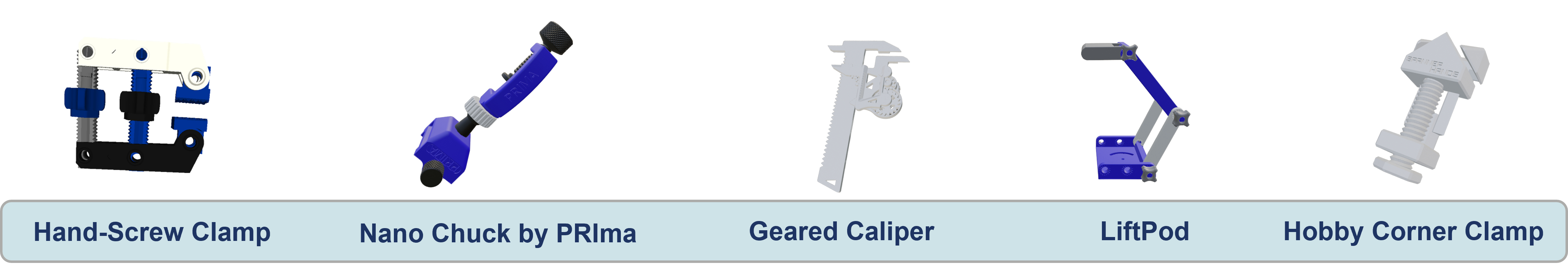}
    \caption{\textbf{An overview of all five assembly assets included in the \ac{gbot}~dataset.}}
    \label{fig:gbot_object}
\end{figure*}

\begin{figure}[t]
    \centering
    \includegraphics[width=\columnwidth]{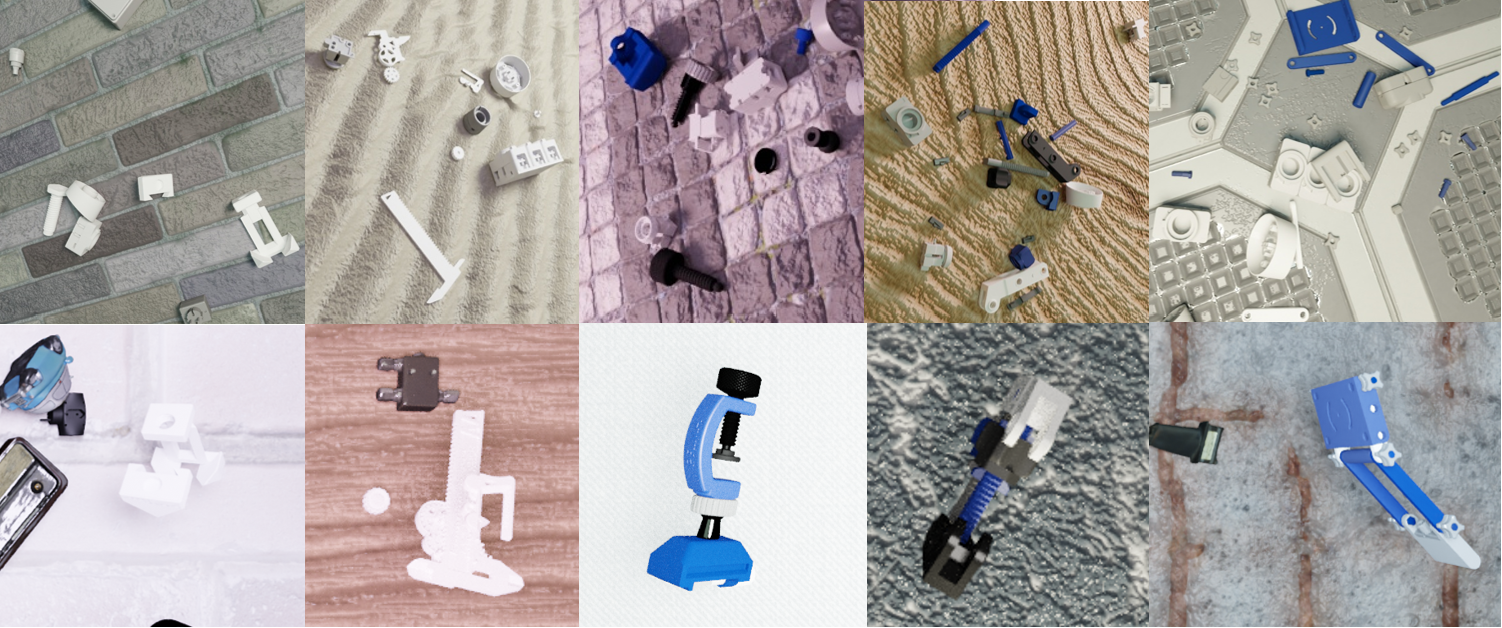}
    \caption{\textbf{Our synthetic training images.} Clustered scenes with 3D printing parts for the assembly parts are generated. To add domain randomization, we add objects from the T-less~\cite{hodan_t-less_2017} dataset, varying lighting conditions, and randomized backgrounds. \label{fig:synthetic}
}   
\end{figure}

\subsubsection{Determining Assembly States}

To switch from one assembly state to another during the assembly process, Su et al. ~\cite{su_deep_2019} used deep learning methods. However, there are many possible connections of assembly parts, so it is challenging to use deep learning methods to detect every possible combination. Therefore, we utilize the knowledge of the relative pose between two assembly parts to acquire the assembly state. We define the switch condition from one state to another by the frame parts of the assembly assets instead of the connectors like screws. These objects are selected to provide robustness towards occlusion scenarios.

A priori, we defined the ground truth assembly pose for each assembly pair. This ground truth could be also retrieved from CAD models or existing instructions. During the assembly process, we measure the relative pose between connecting parts. If the relative pose between two parts (compared to the ground truth poses) is smaller than the tracking error, the specific assembly state is assumed to be finished. We calculate the error of translation and rotation separately due to the difference of symmetrical and unsymmetrical objects. The errors are formulated as follows:

\begin{equation}
e_{trans} = || \pmb{t}_{pred} -\pmb{t}_{gt} ||
\end{equation}
where $\pmb{t}_{pred}$ is the predicted translation and $\pmb{t}_{gt}$ is the defined ground truth translation for two assembly parts.
\begin{equation}
e_{rot} = || arccos(\frac{tr(\pmb{R}_{pred} \pmb{R}_{gt}^T))-1}{2}) ||
\end{equation}
where $e_{rot}$ calculates the absolute value of angle offset between predicted rotation matrix $\pmb{R}_{pred}$ and ground truth $\pmb{R}_{gt}$.

\autoref{fig:state_switch} shows the procedure for switching between assembly states. As an empirically feasible threshold we chose 3~cm for translation offset and 10 degree for rotation offset.

Before the tracking process begins, object-specific trackers are set for different assembly states. Once an assembly step is detected to be finished, the trackers are automatically switched to the new state. The object poses for pre-assembly parts are used to initialize tracking the assembly as a whole in the next step.

\subsection{The \ac{gbot} Dataset}
Currently, there is no real-world or synthetic dynamic assembly dataset available. We utilized synthetic data to enable an easy extension of the dataset in the future. For real-world scenes, annotating moving poses frame-by-frame as ground truth data is challenging. Synthetic data is a common approach for acquiring ground truth data~\cite{schieber2023indoor}. To generate training data, we used publicly available 3D models. However, one can also use 3D scans or CAD models. The dataset consists  of five 3D printing assembly assets from Thingiverse\footnote{\href{https://www.thingiverse.com/thing:4614448}{Liftpod}, \href{https://www.thingiverse.com/thing:5178901}{Nano Chuck by PRima}, \href{https://www.thingiverse.com/thing:2403756}{Hand-Screw Clamp}, \href{https://www.thingiverse.com/thing:1024366}{Hobby Corner Clamp}, \href{https://www.thingiverse.com/thing:3006884/files}{Geared Caliper}} to test our algorithm, see \autoref{fig:gbot_object}. We used Blenderproc~\cite{denninger_blenderproc_2020} to generate RGB-D data as well as ground truth poses with Azure Kinect intrinsic camera model. To bridge the sim-to-real gap, we apply domain randomization, i.e., varying background textures, different lighting conditions, and distracting objects~\cite{schieber2023indoor}. 

To robustly track the object poses during the assembly, we focus on instance-level objects that can be continuously detected. On the one hand, screws and small connectors can be easily occluded by hands and other parts during assembly. On the other hand, there are multiple same screws which lead to only a category-level problem while we focus on the instance-level problem. Therefore, we concentrate on instance-level assembly parts, see Tab.~\ref{tab:obj_tracking}.

\begin{figure*}[t]
        \centering
        \begin{subfigure}{0.195\textwidth}
            \centering
            \includegraphics[trim=0cm 9cm 30cm 7cm, clip, keepaspectratio, width=\textwidth]{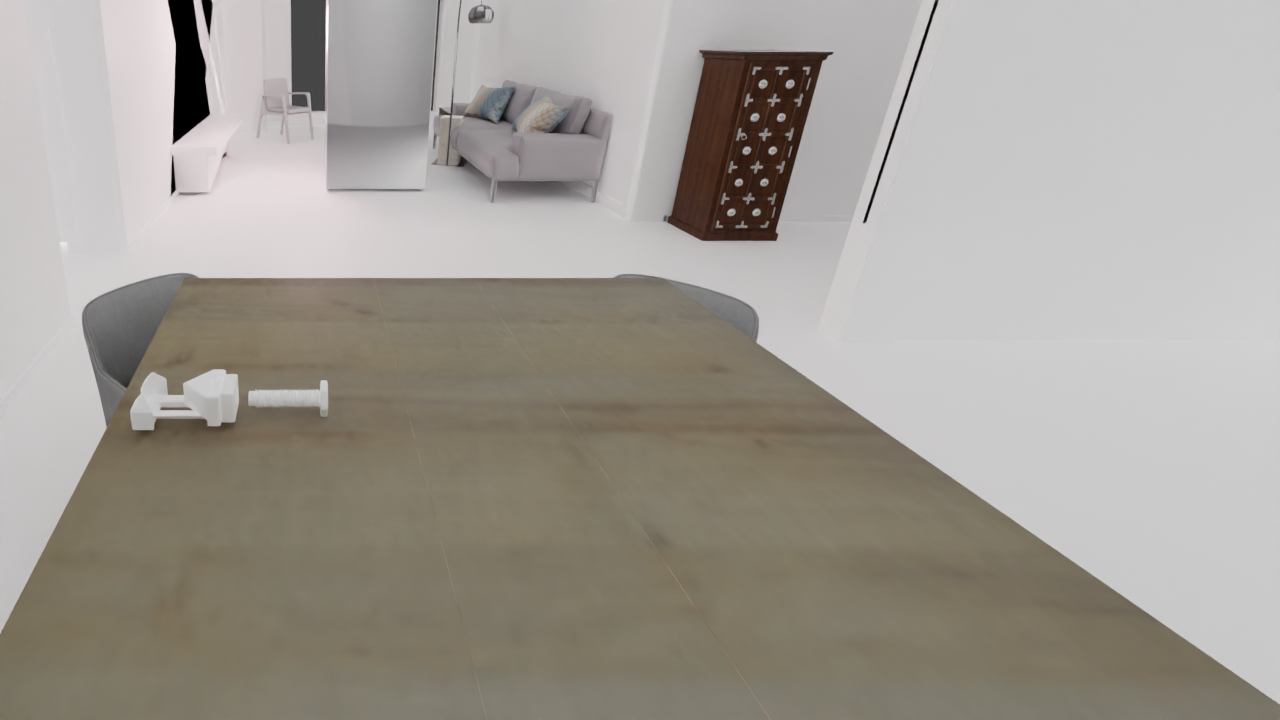}
            \caption{Normal scene}
            \label{fig:mean and std of net14}
        \end{subfigure}
         \hfill
        \begin{subfigure}{0.195\textwidth}  
            \centering 
            \includegraphics[trim=0cm 9cm 30cm 7cm, clip, keepaspectratio, width=\textwidth]{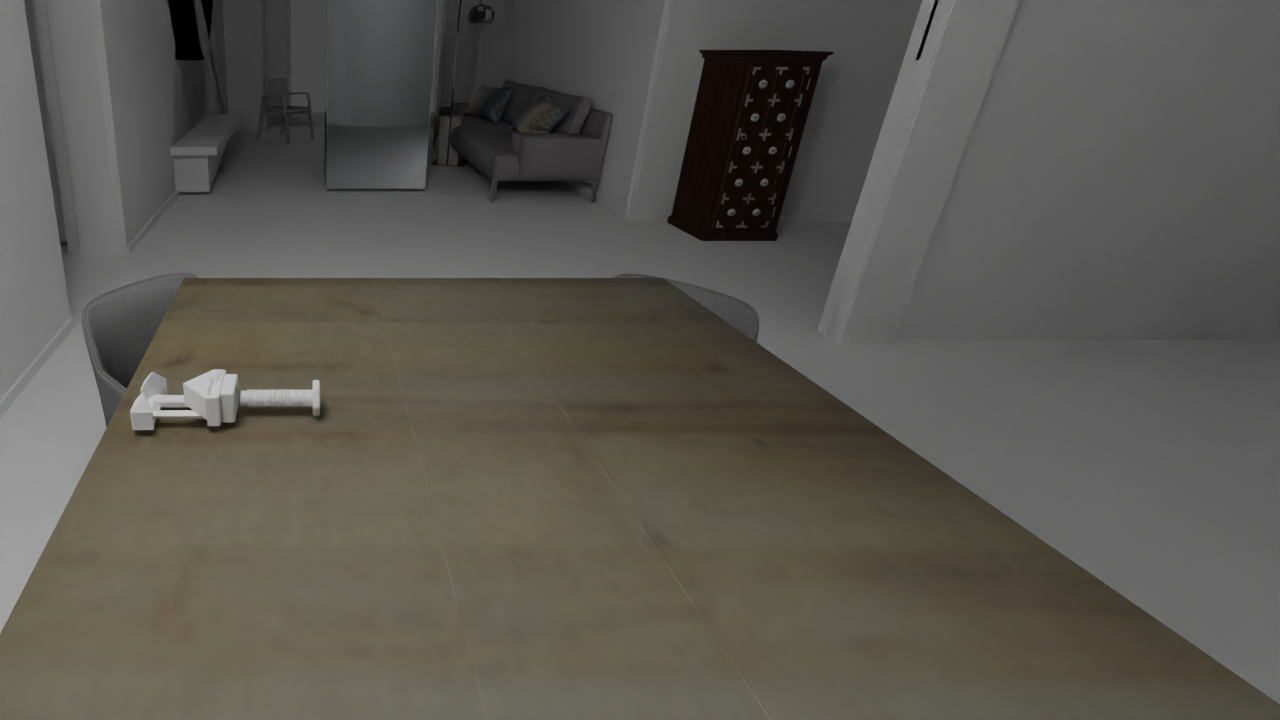}
            \caption{Scene with dynamic light}
            \label{fig:mean and std of net24}
        \end{subfigure}
         \hfill
        \begin{subfigure}{0.195\textwidth}   
            \centering 
            \includegraphics[trim=0cm 8cm 30cm 8cm, clip, keepaspectratio, width=\textwidth]{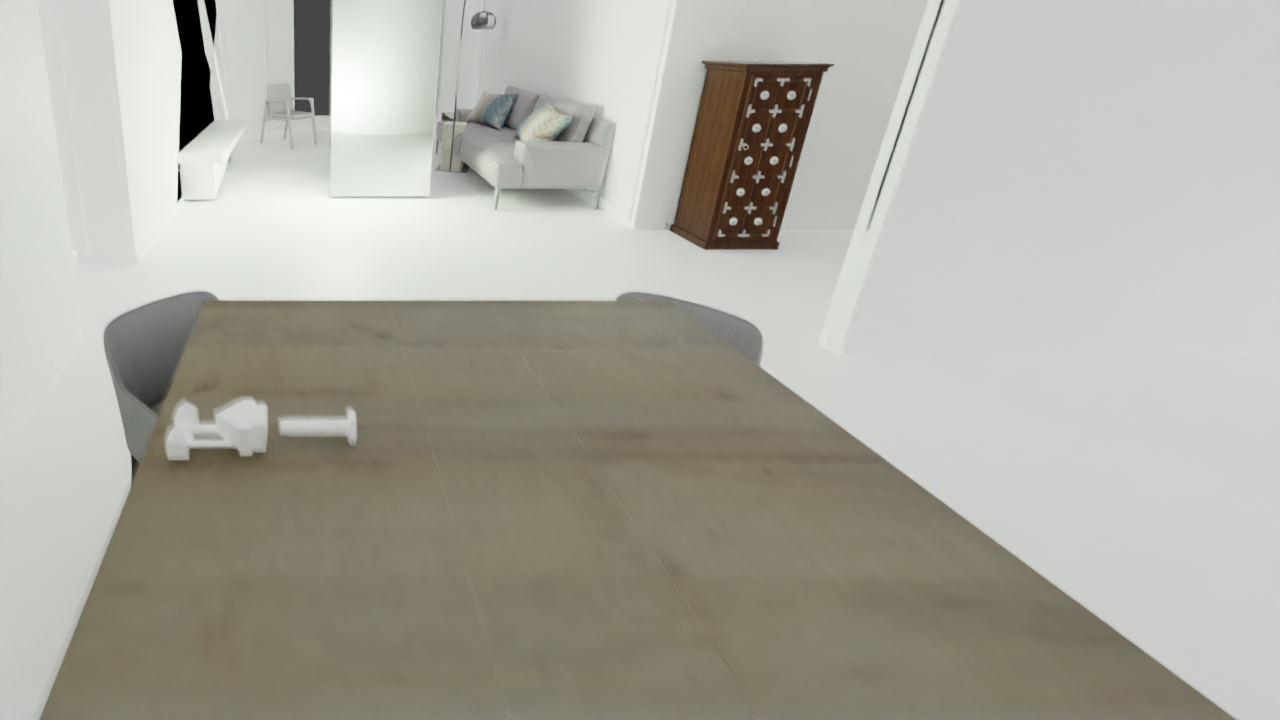}
            \caption{Scene with motion blur}
            \label{fig:mean and std of net34}
        \end{subfigure}
        \hfill
        \begin{subfigure}{0.195\textwidth}   
            \centering 
            \includegraphics[trim=0cm 9cm 30cm 7cm, clip, keepaspectratio, width=\textwidth]{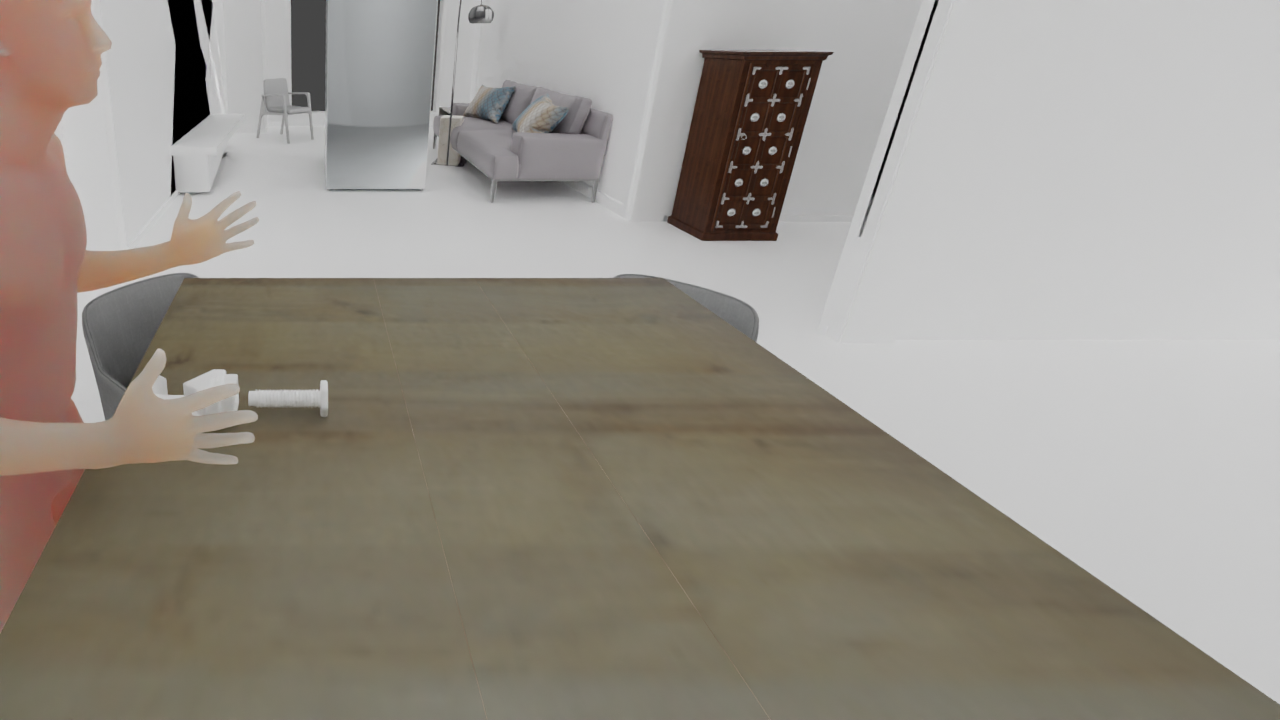} 
            \caption{Scene with hand occlusion}
            \label{fig:mean and std of net44}
        \end{subfigure}
        \hfill
        \begin{subfigure}{0.195\textwidth}   
            \centering 
            \includegraphics[trim=10cm 2.8cm 10cm 7cm, clip, keepaspectratio, width=\textwidth]{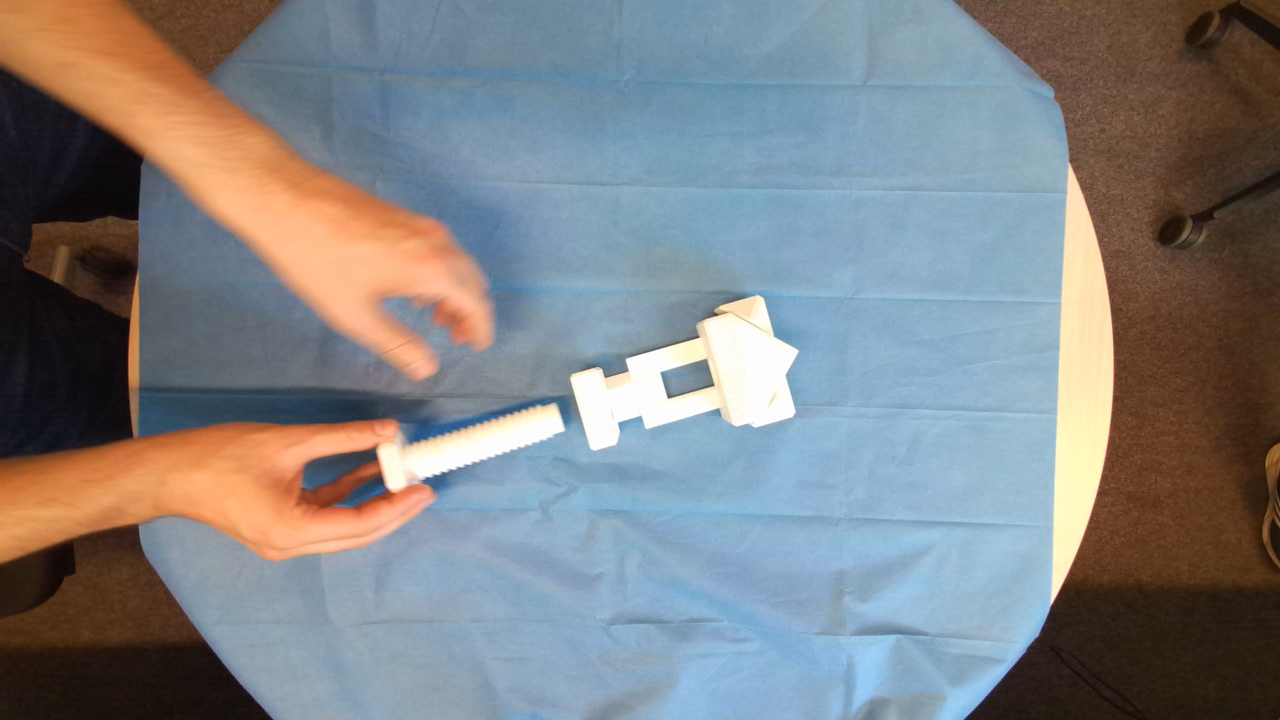} 
            \caption{Real scene}
            \label{fig:mean and std of net44}
        \end{subfigure}
        \caption{\textbf{Synthetic and real scenes with different light conditions, motion blur, and occlusion:} We make ablation studies regarding different light conditions, motion blur, and hand occlusion as real data restrictions.}      
        \label{fig:syn_test}
\end{figure*}

\begin{table}[t!]
  \begin{center}
    \caption{\textbf{Objects for evaluation}: The following instances' poses are to be estimated and tracked. Symmetric objects are labelled in \textbf{bold}.}
    \label{tab:obj_tracking}
    \begin{tabular}{l|p{0.5\linewidth}} \toprule % <-- Alignments: 1st column left, 2nd middle and 3rd right, with vertical lines in between
      \textbf{Assembly assets} & \textbf{Objects} \\
      \midrule 
      Hobby Corner Clamp & Clamp Base, \textbf{Clamp Bolt}, Clamp Jaw\\
      Geared Caliper & Fix, Move Bottom, Move Top Vernier \\
      Nano Chuck by PRIma & \textbf{Balljoint}, Base, \textbf{Headplate}, \textbf{Nut}, \textbf{Screw}, Vise base, \textbf{Vise screw}, Vise slider \\
      Hand-Screw Clamp & Jaw 1, Jaw 2, \textbf{Knob 1}, \textbf{Knob2}, Pad, \textbf{Thread 1}, \textbf{Thread 2}\\
      LiftPod & \textbf{Arm first}, \textbf{Arm last}, \textbf{Bar}, \textbf{Base plate}, Clamp frame, Clamp slider, \textbf{Sleeve} \\ \bottomrule
    \end{tabular}
  \end{center}
\end{table}

\subsubsection{Training Set}

Our training set contains at least 15K images for each assembly asset. The images feature assembled and unassembled objects with 50\% of each variant. To improve domain randomization, we generated synthetic data with different textures, scenes and overlays, distracting objects from T-less~\cite{hodan_t-less_2017}, backgrounds, and varying light conditions, see  \autoref{fig:synthetic}. As ground truth, we export the poses of each object, semantic segmentation mask, and bounding boxes. 

\subsubsection{Test Set}

The test set contains images with $1280\times720$ px resolution in a total number of $13800=(900+400+500+750+900)*4$, divided in five assembly sequences. The sequences are normal, different light conditions, motion blur and hand-occlusion, see \autoref{fig:syn_test}. The normal scene contains objects during the assembly process with an ambient scattered light source. During the sequence, the objects move and get assembled. For capturing, the camera view is fixed and the camera pose is set up to always cover all the assembly parts. The second variant adds dynamic point light based on the first scene. The light point moves randomly in the volume of a sphere to create dynamic light effects. In the third variant blur effects are rendered of moving objects to test the robustness of the tracking algorithm with regard to fast motion. The exposure time is set to one second to capture the blur effects during assembly. For hand occlusion, we use the SMPL human model~\cite{denninger_blenderproc_2020, loper_smpl_2015}.

Moreover, we record five assembly sequences with Azure Kinect from a top-down view as real test cases. For the real scene, we do not have ground truth poses due to the limitation of annotations. We only use the real scenes for qualitative evaluation.

\section{Evaluation}

\subsection{Metrics}

To evaluate the 6D pose accuracy we follow the state-of-the-art metric \ac{add}~\cite{hinterstoisser_model_2013}.  \ac{add} denotes the average distance error $e_\textrm{\ac{add}}$ for unsymmetrical objects and $e_\textrm{\ac{add}-S}$ of symmetric objects:

\begin{equation}
\label{eq:e00}
e_\textrm{\ac{add}} = \frac{1}{n_v}\sum_{i=1}^{n_v}|| (\pmb{R}_{pred} \mathbf{x} + \pmb{T}_{pred}) - (\pmb{R}_{gt} \mathbf{x} + \pmb{T}_{gt}) ||,  
\end{equation}

\begin{equation}
e_\text{\ac{add}\textnormal{-}S} = \frac{1}{n_\textrm{v}}\sum_{i=1}^{n_\textrm{v}} \min_{j\in[n_\textrm{v}]}|| (\pmb{R}_{pred} \mathbf{x}_i + \pmb{T}_{pred}) - (\pmb{R}_{gt} \mathbf{x}_j + \pmb{T}_{gt}) ||
\label{eq:e01}
\end{equation}
where $e_\textrm{\ac{add}}$ represents the average distance between predicted model points and ground truth model points, $\mathbf{x}$ is a vertex point from the 3D object mesh written in object coordinates, $n_\textrm{v}$ is the number of vertices of object mesh, the average distance $e_\textrm{\ac{add}-S}$ for symmetric objects is computed using the closest point distance between $\mathbf{x}_i$ and~$\mathbf{x}_j$.

The 6D pose is considered to be correct if the average distance is smaller than a predefined threshold. \textrm{\ac{add}}/\textrm{\ac{add}-S} are computed based on the formula below:

\begin{equation}\label{eq:e02}
	s_i = \frac{1}{ n_\textrm{f}}\sum_{j=1}^{n_\textrm{f}} \max\Big(1 - \frac{e_{j}}{e_\textrm{t}}, 0\Big),
\end{equation}
with $s_i \in {\textrm{\ac{add}}, \textrm{\ac{add}-S}}$,  $n_\textrm{f}$ the number of frames, and $e_\textrm{t}$ an error threshold which is defined as 10~cm in our case.

However, these metrics cannot measure the translation and rotation error intuitively. Especially in the use case of \ac{ar}, the error measured in length and angle need to be defined for our assembly state acquirement, so we defined the metrics as follows:

\begin{equation}
e_{ave\_ trans} = \frac{1}{n_f}\sum_{i=1}^{n_f}|| \pmb{t}_{pred} - \pmb{t}_{gt} ||
\end{equation}

\begin{equation}
e_{ave\_ rot} = \frac{1}{n_f}\sum_{i=1}^{n_f}|| arccos(\frac{tr(\pmb{R}_{pred} \pmb{R}_{gt}^T))-1}{2}) ||
\end{equation}
where $e_{ave\_ trans}$ is the translation error measured in cm and $e_{ave\_ rot}$ is the rotation error measured in degree. $\pmb{R}_{gt}$ denotes the ground truth rotation and $\pmb{t}_{gt}$ the ground truth translation. $\pmb{R}_{pred}$ describes the predicted rotation and $\pmb{t}_{pred}$ the predicted rotation. 

Additionally, we report the average inference runtime measured in ms for each algorithm.

\subsection{Implementation Details}

Our object pose estimation algorithm extends YOLOv8~\cite{jocher_yolo_2023}, is implemented in PyTorch for 6D pose estimation, and uses NVIDIA TensorRT to accelerate and optimize inference performance. YOLOv8Pose inference engine, our tracking, and the RESTful API are implemented in C++ 17. For the RESTful API we use Restbed~\cite{restbed}. Via our RESTful API, individual output devices can access our tracking data. The code of GBOT is publicly available on GitHub to ensure reproducibility and reusability for further future study.

We use a workstation with an Intel(R) Core(TM) i9-10980XE CPU and NVIDIA GeForce RTX 3090 GPU with 24GB VRAM. For GBOT we additionally followed the implementation of a re-initialization, denoted as GBOT + re-init in the following. We added re-initialization every 10th frame~\cite{li2018deepim} if the tracking offset is larger than 5cm compared with YOLOv8Pose.

\subsection{Evaluation on GBOT Dataset}

For evaluation we used the test split our \ac{gbot} dataset with four conditions (normal, dynamic light, motion blur, hand occlusion). We compare our YOLO-based 6D pose estimation, denoted as YOLOv8Pose, the state-of-the-art tracking approaches~\cite{stoiber_iterative_2022,stoiber_srt3d_2022} and our multi-state assembly tracking approach \ac{gbot}. We do not compare our approach with Mb-ICG ~\cite{stoiber_multi-body_2022}, because it needs a re-initialization per assembly step. Mb-ICG is a single-state approach, which cannot be applied to the multi-state assembly sequences. For YOLOv8Pose we conducted two training strategies. First, we trained it only on unassembled objects and evaluated its performance. Then, we trained YOLOv8Pose on the full training split including assembled objects. We evaluate the performance separately for assembled assets of Hobby Corner Clamp, Geared Caliper, Nano Chuck by PRIma, Hand-Screw Clamp and Liftpod, see~\autoref{tab:obj_tracking}.

\subsection{Quantitative Evaluation}

\begin{table}[t!]
    \centering
    \caption{\textbf{Average runtime in ms of the compared methods.}}
    \begin{tabular}{l|c} \toprule
        Approach &  {Avg. runtime [ms]} \\ \midrule
        \textbf{6D pose estimation} &  \\
        ~~YOLOv8Pose (ours) & 63.60\\ \midrule 
        \textbf{Tracking} &  \\
        ~~SRT3D \cite{stoiber_srt3d_2022} & 11.87\\
        ~~ICG \cite{stoiber_iterative_2022} & 27.75\\
        ~~ICG+SRT3D \cite{stoiber_srt3d_2022,stoiber_iterative_2022,li2023more} & 38.75\\
        ~~GBOT~(ours)  & 28.32\\ \midrule
        \textbf{6D pose estimation + Tracking} & \\
        ~~\ac{gbot} + re-init~(ours) & 36.89\\ \bottomrule
    \end{tabular}
    \label{tab:runtime}
\end{table}

\begin{table*}[t!]
\caption{\textbf{Results on the \ac{gbot}~dataset:} Rotational errors are only evaluated for unsymmetrical objects. We report \ac{add}(S)  ($\uparrow$) with $10cm$ threshold, the translation error in centimeters (cm) ($e_{ave\_{trans}}$ denoted as $e_{trans}$, $\downarrow$) and the rotation error ($e_{ave\_ rot}$ denoted as $e_{rot}$, $\downarrow$) in degrees. The best results among all methods are labeled in bold. In this evaluation, tracking is initialized with ground truth pose only for the first frame and the pose is not reinitialized afterwards. In the last column, we show the tracking results of tracking re-initialization by pose estimation. GBOT outperforms the state-of-the-art tracking approaches, GBOT re-init and YOLOv8Pose.}
    \label {tab:quan_eval}
\centering
   \resizebox{\textwidth}{!}{    
        \begin{tabular}{ll|ccc|ccc|ccc|ccc|ccc|ccc |ccc}\toprule
        \multicolumn{2}{c|}{Approach} & 
        \multicolumn{6}{c|}{{\centering 6D pose estimation}}& 
        \multicolumn{12}{c|}{{\centering Tracking}}&
        \multicolumn{3}{c}{{\centering 6D pose estimation + tracking}} \\ \midrule
        \multicolumn{2}{c|}{Modality} & 
        \multicolumn{9}{c|}{{\centering RGB}}& 
        \multicolumn{12}{c}{{\centering RGB-D}} \\ \midrule
        \multirow{3}{*}{Asset} & \multirow{3}{*}{Condition} & \multicolumn{3}{c|}{YOLOv8Pose} & \multicolumn{3}{c|}{YOLOv8Pose} &\multicolumn{3}{c|}{SRT3D \cite{stoiber_srt3d_2022}} &\multicolumn{3}{c|}{ICG \cite{stoiber_iterative_2022}} &\multicolumn{3}{c|}{ ICG+SRT3D~\cite{stoiber_srt3d_2022,stoiber_iterative_2022,li2023more}} & \multicolumn{3}{c|}{GBOT~(ours)} & \multicolumn{3}{c}{GBOT + re-init (ours)} \\ 
        &  & \multicolumn{3}{c|}{(ours, without assembly)} & \multicolumn{3}{c|}{(ours, with assembly)} & \multicolumn{3}{c|}{}  & \multicolumn{3}{c|}{}  & \multicolumn{3}{c|}{}  & \multicolumn{3}{c|}{}  &  \multicolumn{3}{c}{} \\ 
         & & {\small $\ac{add}(S) \uparrow$} & $e_{trans} \downarrow$ & $e_{ rot} \downarrow$  & {\small $\ac{add}(S) \uparrow$} & $e_{trans} \downarrow$ & $e_{ rot} \downarrow$  & {\small $\ac{add}(S) \uparrow$} & $e_{trans} \downarrow$  & $e_{ rot} \downarrow$ & {\small $\ac{add}(S) \uparrow$} & $e_{trans} \downarrow$  & $e_{ rot} \downarrow$ & {\small $\ac{add}(S) \uparrow$} & $e_{trans} \downarrow$  & $e_{ rot} \downarrow$ & {\small $\ac{add}(S) \uparrow$} & $e_{trans} \downarrow$  & $e_{ rot} \downarrow$ & {\small $\ac{add}(S) \uparrow$} & $e_{trans} \downarrow$  & $e_{ rot} \downarrow$\\ \midrule
        \multirow{4}{1.5cm}{Hobby Corner Clamp}  
          & Normal   & 50.2 & 17.0 & 38.9 & 91.5 & 9.3 & 3.8 & 89.8 & 4.7 & 25.8 & \textbf{100.0} & \textbf{0.1} & 38.7 & \textbf{100.0} & 1.3 & 38.7 & \textbf{100.0} & 0.6 & \textbf{2.1} & 99.5 & 0.7 & 2.7\\
          & Dynamic     & 56.7 & 18.0 & 37.4 & 99.0 & 2.5 & 4.8 & 88.4 & 3.7 & 25.0 & \textbf{100.0} & 1.5 & 46.8 & \textbf{100.0} & 2.0 & 47.8 & \textbf{100.0} & 0.6 & 23.2 & \textbf{100.0} & \textbf{0.5} & \textbf{3.5}\\
          & Hand     & 18.3 & 31.6 & 133.5 & 45.4 & 54.1 & 97.1 & 66.6 & 8.8 & 37.1 & 68.4 & 7.4 & 59.4 & 68.4 & 7.5 & \textbf{58.3} & 81.9 & 5.7 & 90.4 & \textbf{90.6} & \textbf{4.4} & 84.7\\
          & Blur    & 50.4 & 18.0 & 38.1 & 97.3 & 4.0 & 4.3 & 88.9 & 5.0 & 30.4 & \textbf{100.0} & \textbf{0.5} & 2.2 & \textbf{100.0} & 1.0 & 3.0 & \textbf{100.0} & 0.6 & \textbf{2.1} & 99.9 & 0.6 & \textbf{2.1}\\  \midrule 
        \multirow{4}{1.5cm}{Geared Caliper}  
          & Normal   & 99.3 & 2.3 & 42.0 & 99.6 & 1.4 & 10.1 & 90.6 & 1.8 & 5.4 & \textbf{100.0} & \textbf{0.2} & 2.6 & \textbf{100.0} & 0.3 & 3.0 & \textbf{100.0} & \textbf{0.2} & \textbf{2.1} & \textbf{100.0} & 0.5 & 3.3 \\
          & Dynamic     & 93.3 & 3.8 & 41.0 & 99.9 & 1.3 & 9.3 & 92.7 & 1.4 & 9.9 & \textbf{100.0} & \textbf{0.2} & 2.5 & \textbf{100.0} & 0.4 & 3.6 & \textbf{100.0} & \textbf{0.2} & \textbf{2.3} & \textbf{100.0} & 0.5 & 3.6\\
          & Hand    & 98.2 & 2.7 & 38.8 & 99.2 & 2.4 & 12.5 & 96.5 & 0.9 & \textbf{4.4} & 85.4 & 3.0 & 30.0 & 85.5 & 3.1 & 30.0 & 85.4 & 3.0 & 30.0 & \textbf{99.6} & \textbf{0.8} & 7.0\\
          & Blur     & 99.1 & 2.2 & 39.4 & 99.6 & 1.4 & 9.9 & 98.9 & 0.9 & 8.7 & \textbf{100.0} & \textbf{0.2} & 2.5 & \textbf{100.0} & 0.3 & 2.8 & \textbf{100.0} & \textbf{0.2} & \textbf{2.2} & \textbf{100.0} & 0.5 & 3.6\\ \midrule 
        \multirow{4}{1.5cm}{Nano Chuck by PRIma}  
          & Normal   & 19.6 & 19.6 & 139.0 & 71.1 & 8.8 & 4.8 & 74.1 & 6.6 & 13.8 & 89.8 & 2.1 & 16.7 & 89.4 & 2.3 & 16.5 & \textbf{99.8} & \textbf{0.6} & 7.2 & 93.8 & 2.4 & \textbf{3.6}  \\
          & Dynamic    & 18.5 & 26.4 & 144.8 & 72.6 & 7.6 & 4.6 & 63.4 & 10.3 & 15.6 & 87.3 & 3.9 & 15.3 & 87.3 & 4.1 & 15.8 & \textbf{96.0} & \textbf{2.4} & 20.0 & 92.9 & 2.5 &  \textbf{3.7}\\
          & Hand     & 17.7 & 30.1 & 120.7 & 65.9 & 11.0 & \textbf{4.7} & 61.4 & 13.6 & 15.3 & 76.5 & 6.0 & 18.3 & 75.8 & 6.1 & 18.1 & 72.9 & 8.7 & 14.5 & \textbf{87.8} & \textbf{3.1} & 7.3\\
          & Blur     & 19.7 & 20.4 & 150.6  & 70.9 & 9.9 & 5.2 & 61.9 & 11.6 & 15.2 & 91.6 & 1.9 & 11.4 & 91.5 & 2.1 & 11.1 & \textbf{95.7} & \textbf{0.7} & 25.1 & 92.7 & 3.0 & \textbf{4.7}\\ \midrule 
        \multirow{4}{1.5cm}{Hand-Screw Clamp}  
          & Normal   & 45.8 & 17.3 & 66.4 & 73.5 & 7.7 & 17.6 & 86.5 & 4.7 & 8.9 & 96.0 & 1.2 & 1.1 & 95.9 & 1.4 & 2.3 & \textbf{98.8} & \textbf{0.4} & \textbf{0.9} & 83.7 & 3.0 & 4.7\\
          & Dynamic     & 38.1 & 17.5 & 64.8 & 82.0 & 6.7 & 18.4 & 86.4 & 5.6 & 27.0 & 95.9 & 1.7 & 2.1 & 95.9 & 2.2 & 3.4 & \textbf{98.8} & \textbf{0.6} & \textbf{1.7} & 91.6 & 2.1 & 6.1\\
          & Hand     & 33.4 & 18.7 & 80.9 & 67.2 & 14.2 & 39.8 & 60.1 & 14.3 & 37.7 & 73.4 & 6.9 & 53.4 & 73.1 & 7.1 & 53.4 & 68.7 & 7.9 & 61.9 & \textbf{83.9} & \textbf{4.9} & \textbf{27.2}\\
          & Blur     & 44.3 & 16.7 & 66.1 & 85.6 & 6.5 & 18.6 & 86.5 & 5.6 & 34.6 & 95.7 & 3.0 & 6.6 & 95.7 & 3.2 & 7.7 & \textbf{98.6} & \textbf{1.2} & \textbf{1.1} & 91.1 & 3.0 & 4.8 \\ \midrule 
        \multirow{4}{1.5cm}{LiftPod}  
          & Normal   & 26.9 & 24.3 & 119.6 & 69.1 & 13.4 & 6.7 & 68.9 & 11.5 & 60.0 & \textbf{100.0} & 0.3 & 2.3 & \textbf{100.0} & 0.7 & 3.9 & \textbf{100.0} & \textbf{0.2} & \textbf{1.5} & 84.3 & 5.5 & 11.4\\
          & Dynamic     & 28.4 & 26.8 & 127.8 & 64.3 & 11.6 & 7.7 & 56.0 & 14.4 & 6.8 & 85.1 & 7.7 & 42.9 & 85.0 & 8.1 & 45.4 & \textbf{100.0} & \textbf{0.3} & \textbf{1.6} & 80.8 & 5.5 & 13.1\\
          & Hand     & 25.6 & 29.3 & 116.7 & 62.6 & 19.6 & 19.9 & 39.1 & 25.6 & 60.6 & 51.7 & 17.6 & 74.9 & 51.8 & 17.8 & 76.8 & 51.7 & 17.6 & 74.9 & \textbf{70.0} & \textbf{10.1} & \textbf{7.2}\\
          & Blur     & 31.1 & 24.3 & 123.7 & 68.3 & 13.9 & 6.9 & 76.9 & 11.9 & 5.2 & \textbf{100.0} & 0.3 & 2.9 & \textbf{100.0} & 0.8 & 3.4 & \textbf{100.0} & \textbf{0.2} & \textbf{1.4} & 83.1 & 5.7 & 10.3\\ \midrule 
         Mean & Overall  & 45.7 & 18.4 & 86.5 & 79.2 & 10.4 & 15.3 & 76.7 & 8.1 & 22.4 & 89.8 & 3.3 & 21.6 & 89.8 & 3.6 & 22.3 & \textbf{92.4} & \textbf{2.6} & 18.3 & 91.3 & 3.0 & \textbf{10.7}\\
        \bottomrule
        \end{tabular}}
\end{table*}

\begin{figure*}[t!]
        \captionsetup[subfigure]{labelformat=empty}
        \centering
        \begin{subfigure}{0.12\textwidth}   
            \centering 
            \includegraphics[trim=16cm 9cm 14cm 3cm,clip,width=2.4cm,keepaspectratio]{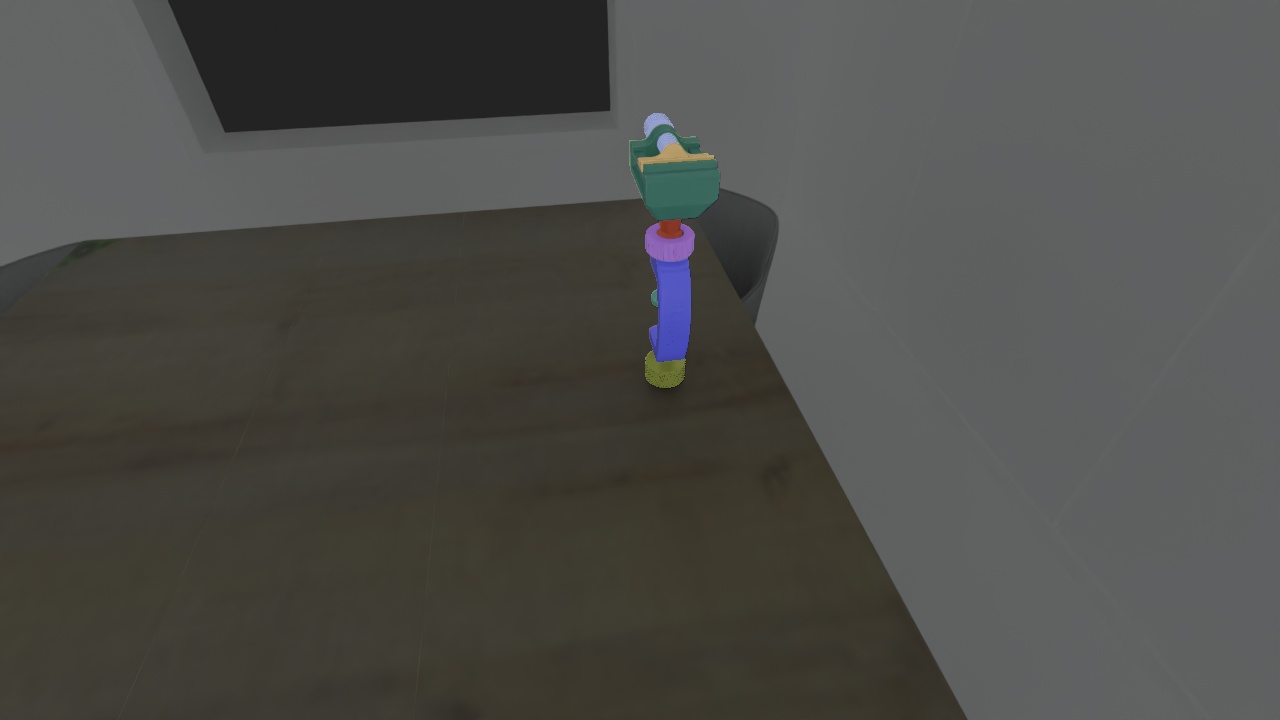}
            %\caption{NanoChuckbyPrima (Ground truth)}
        \end{subfigure}
        \hfill
        \begin{subfigure}{0.12\textwidth}   
            \centering 
            \includegraphics[trim=16cm 9cm 14cm 3cm,clip,width=2.4cm,keepaspectratio]{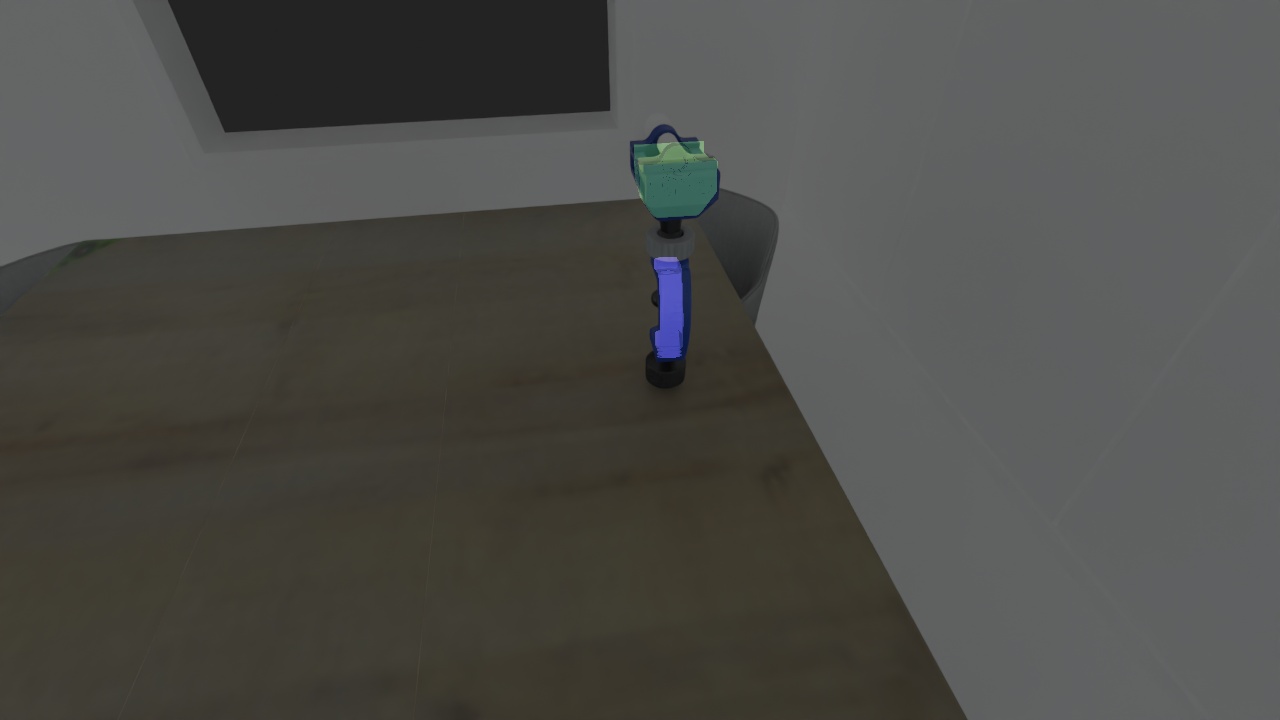}
            %\caption{NanoChuckbyPrima (YOLOv8pose (w/o a))}
        \end{subfigure}
        \hfill
        \begin{subfigure}{0.12\textwidth}   
            \centering 
            \includegraphics[trim=16cm 9cm 14cm 3cm,clip,width=2.4cm,keepaspectratio]{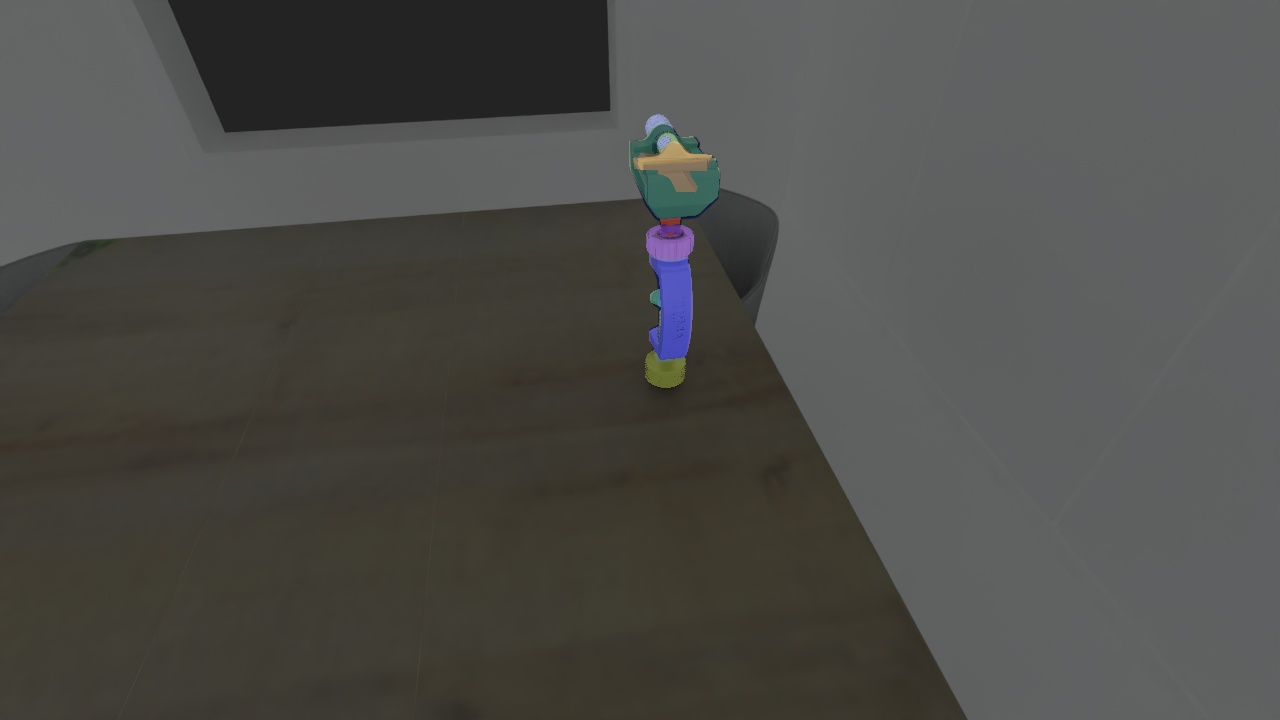}
            %\caption{NanoChuckbyPrima (YOLOv8pose (w a))}
        \end{subfigure}
        \hfill
        \begin{subfigure}{0.12\textwidth}   
            \centering 
            \includegraphics[trim=16cm 9cm 14cm 3cm,clip,width=2.4cm,keepaspectratio]{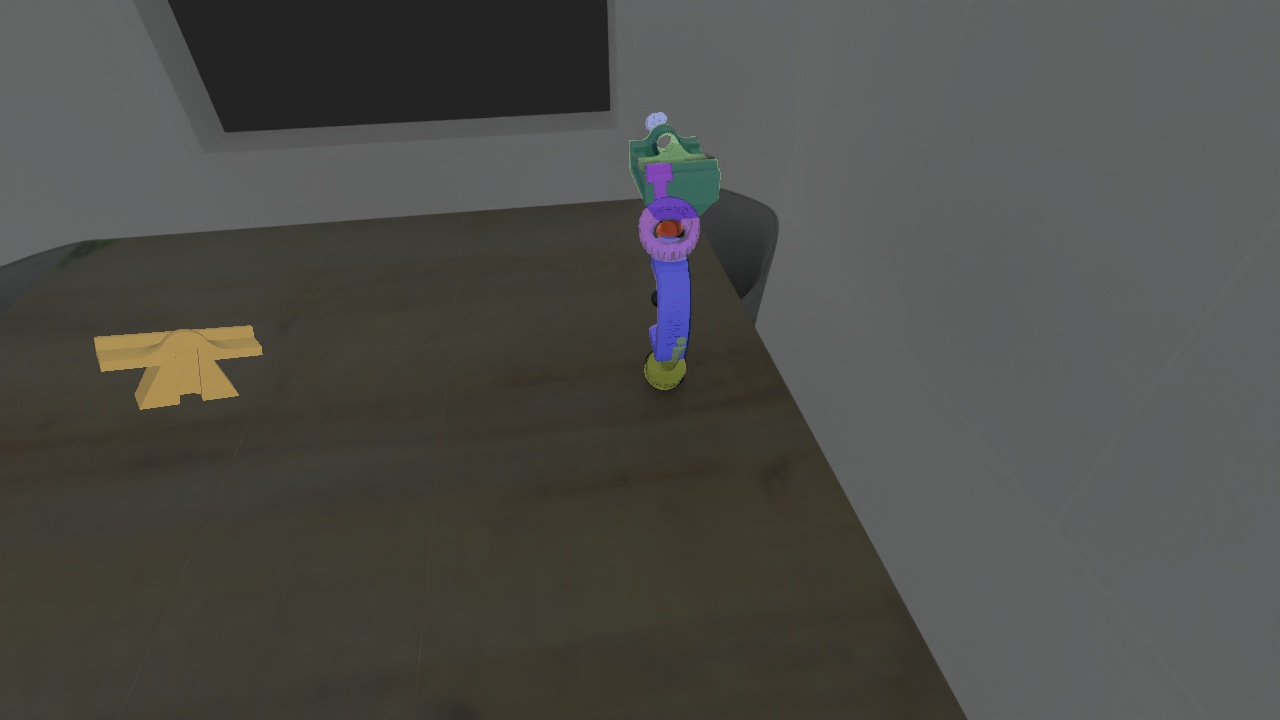}
            %\caption{NanoChuckbyPrima (SRT3D (w a))}
        \end{subfigure}
        \hfill
        \begin{subfigure}{0.12\textwidth}   
            \centering 
            \includegraphics[trim=16cm 9cm 14cm 3cm,clip,width=2.4cm, keepaspectratio]{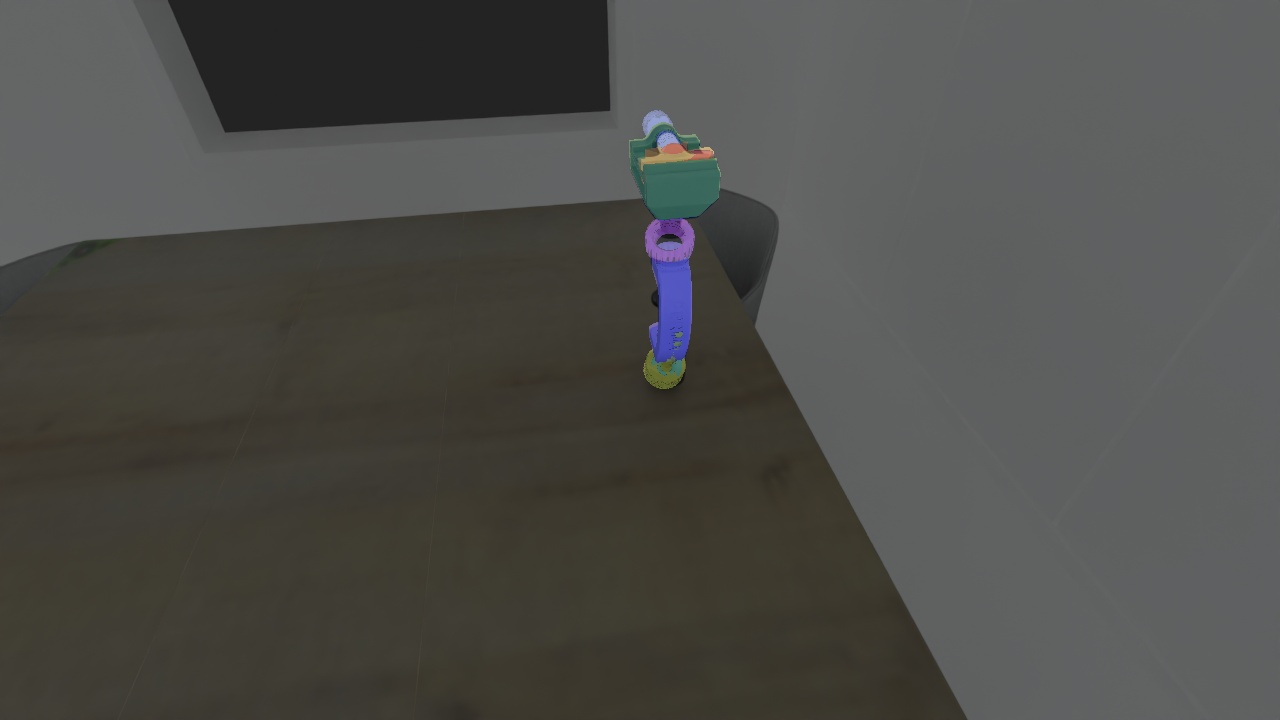}
            %\caption{NanoChuckbyPrima (ICG)}
        \end{subfigure}
        \hfill
        \begin{subfigure}{0.12\textwidth}   
            \centering 
            \includegraphics[trim=16cm 9cm 14cm 3cm,clip,width=2.4cm, keepaspectratio]{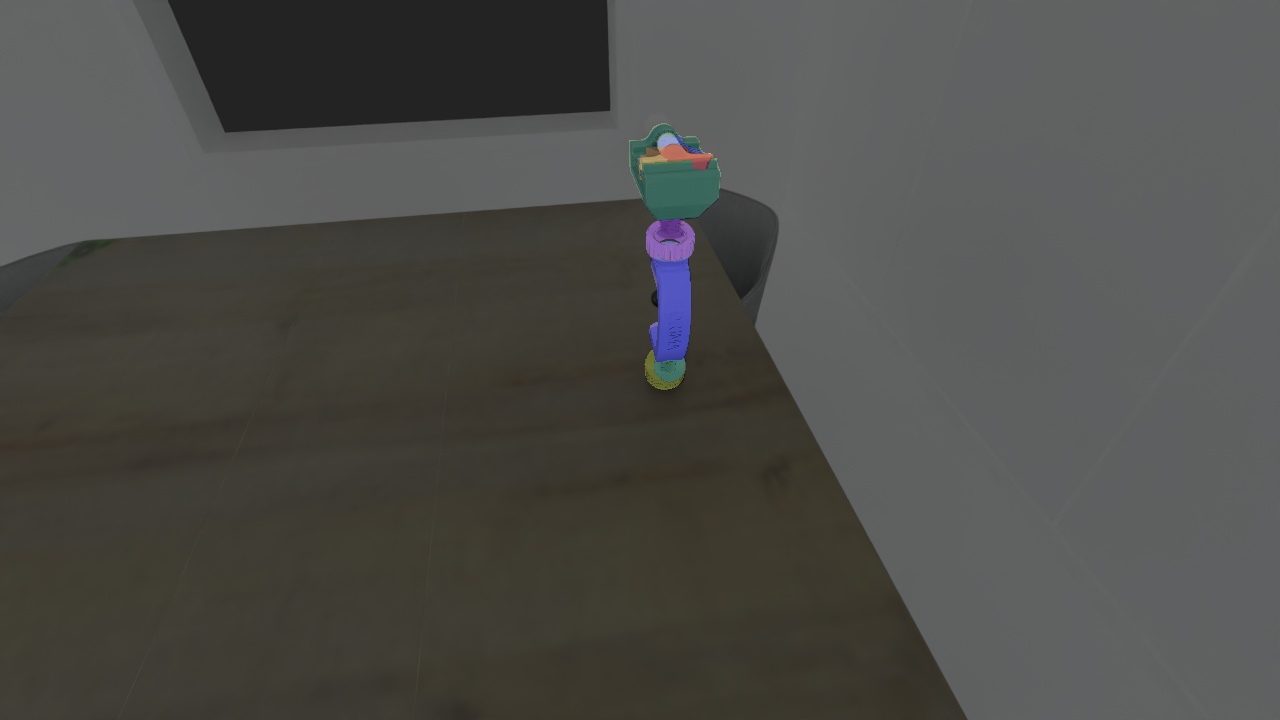}
            %\caption{NanoChuckbyPrima (ICG+SRT3D)}
        \end{subfigure}
        \hfill
        \begin{subfigure}{0.12\textwidth}   
            \centering 
            \includegraphics[trim=16cm 9cm 14cm 3cm,clip,width=2.4cm, keepaspectratio]{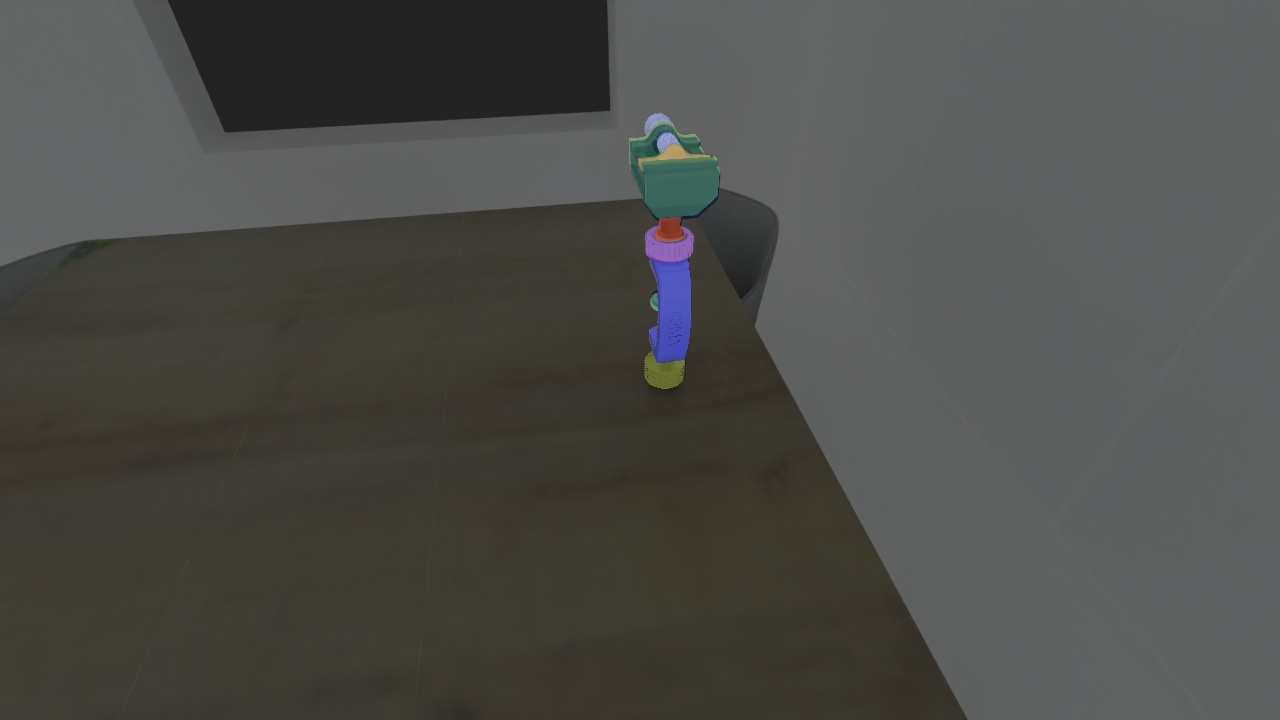} 
           % \caption{NanoChuckbyPrima (GBOT)}
        \end{subfigure}
        
        \begin{subfigure}{0.12\textwidth}   
            \centering 
            \includegraphics[trim=10cm 5cm 22cm 8cm,clip,width=2.4cm, keepaspectratio]{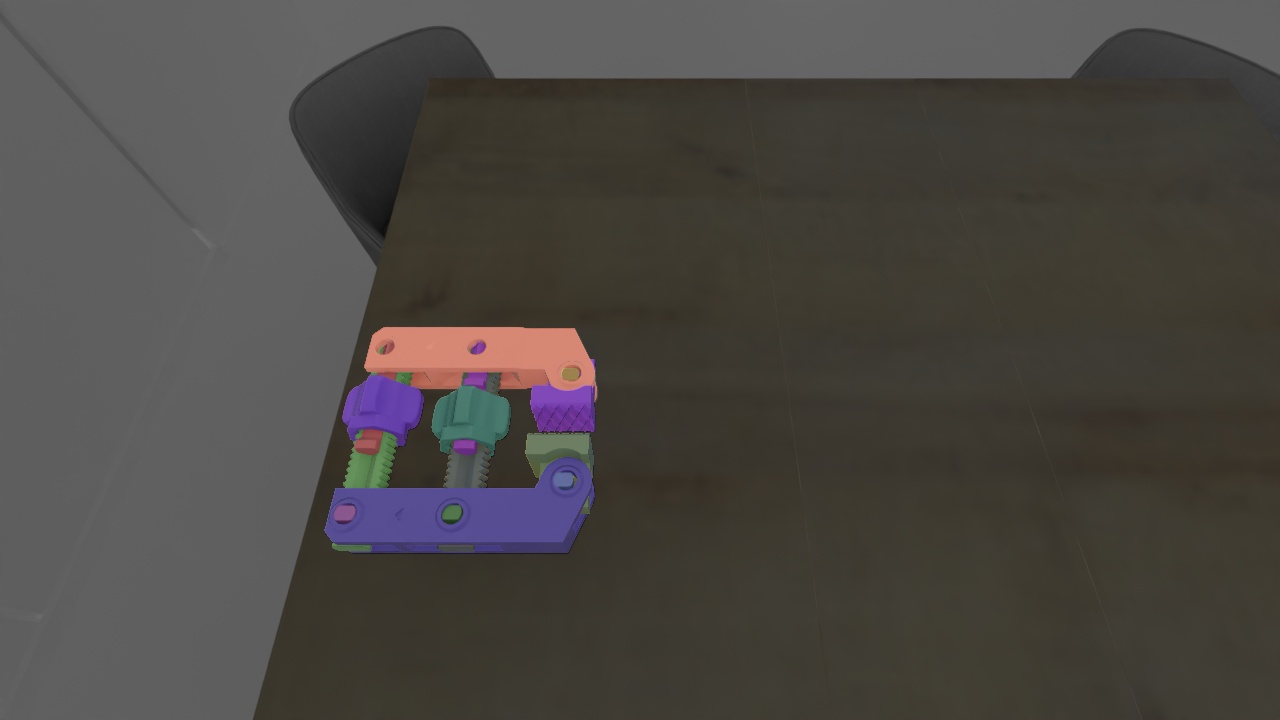}
           % \caption{HandScrewClamp (Ground truth)}
        \end{subfigure}
        \hfill
        \begin{subfigure}{0.12\textwidth}   
            \centering 
            \includegraphics[trim=10cm 5cm 22cm 8cm,clip,width=2.4cm, keepaspectratio]{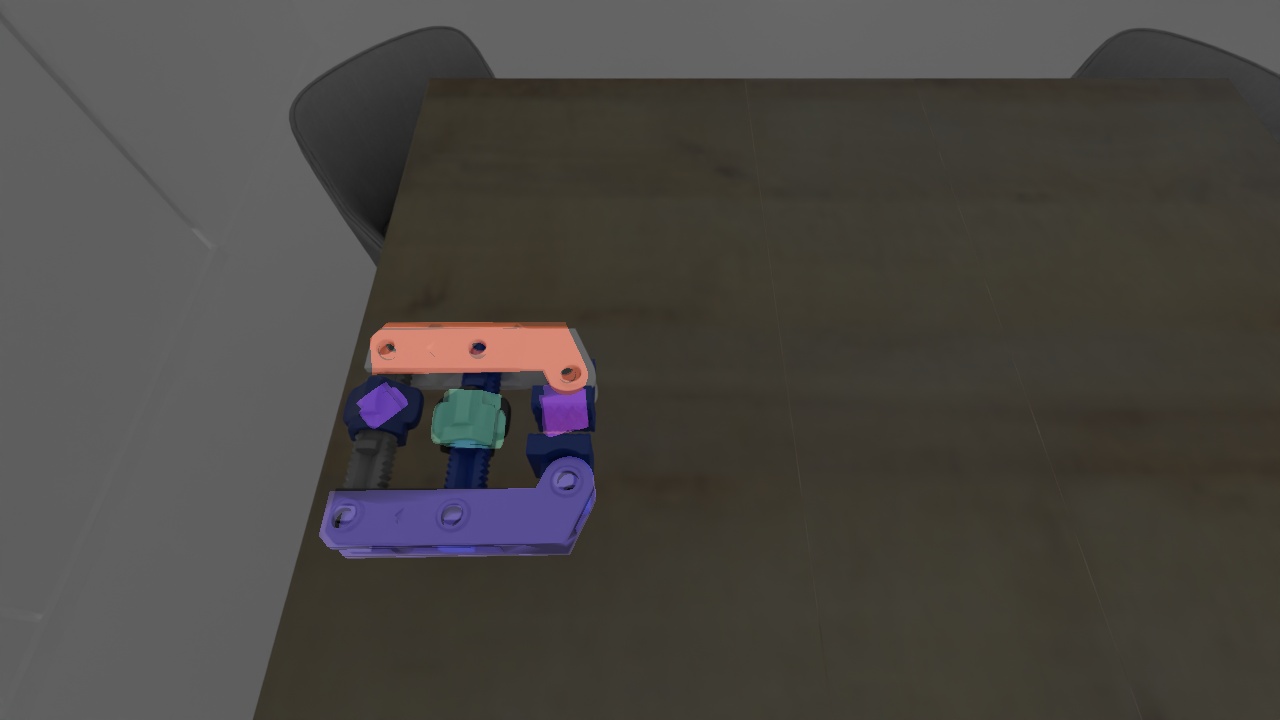}
            %\caption{HandScrewClamp (YOLOv8pose (w/o a))}
        \end{subfigure}
        \hfill
        \begin{subfigure}{0.12\textwidth}   
            \centering 
            \includegraphics[trim=10cm 5cm 22cm 8cm,clip,width=2.4cm, keepaspectratio]{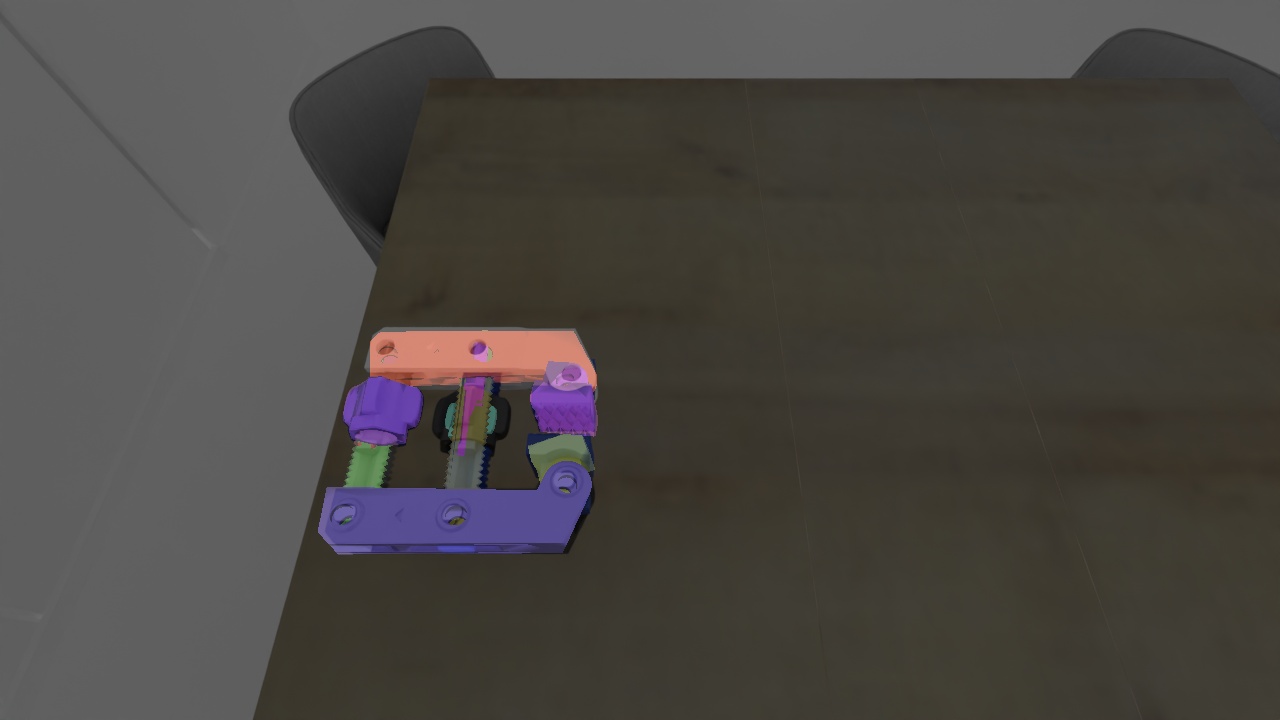}
            %\caption{HandScrewClamp (YOLOv8pose (w a))}
        \end{subfigure}
        \hfill
        \begin{subfigure}{0.12\textwidth}   
            \centering 
            \includegraphics[trim=10cm 5cm 22cm 8cm,clip,width=2.4cm, keepaspectratio]{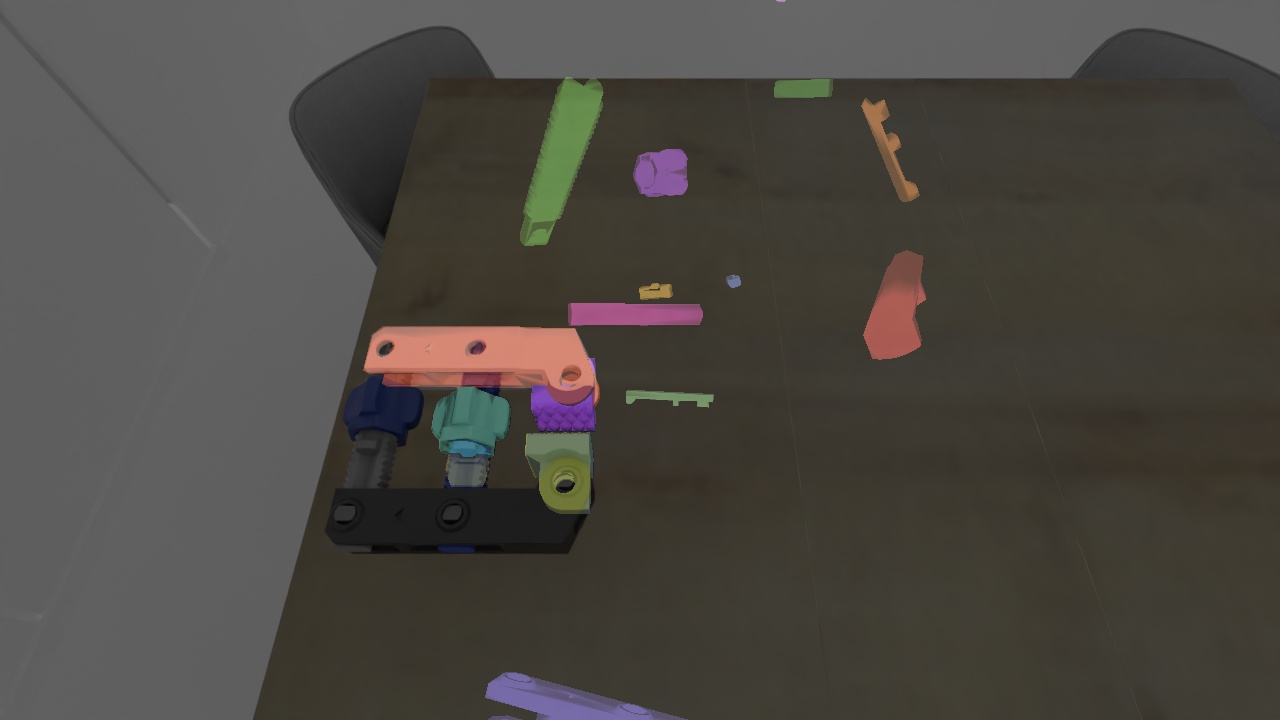}
            %\caption{HandScrewClamp (SRT3D)}
        \end{subfigure}
        \hfill
        \begin{subfigure}{0.12\textwidth}   
            \centering 
            \includegraphics[trim=10cm 5cm 22cm 8cm,clip,width=2.4cm, keepaspectratio]{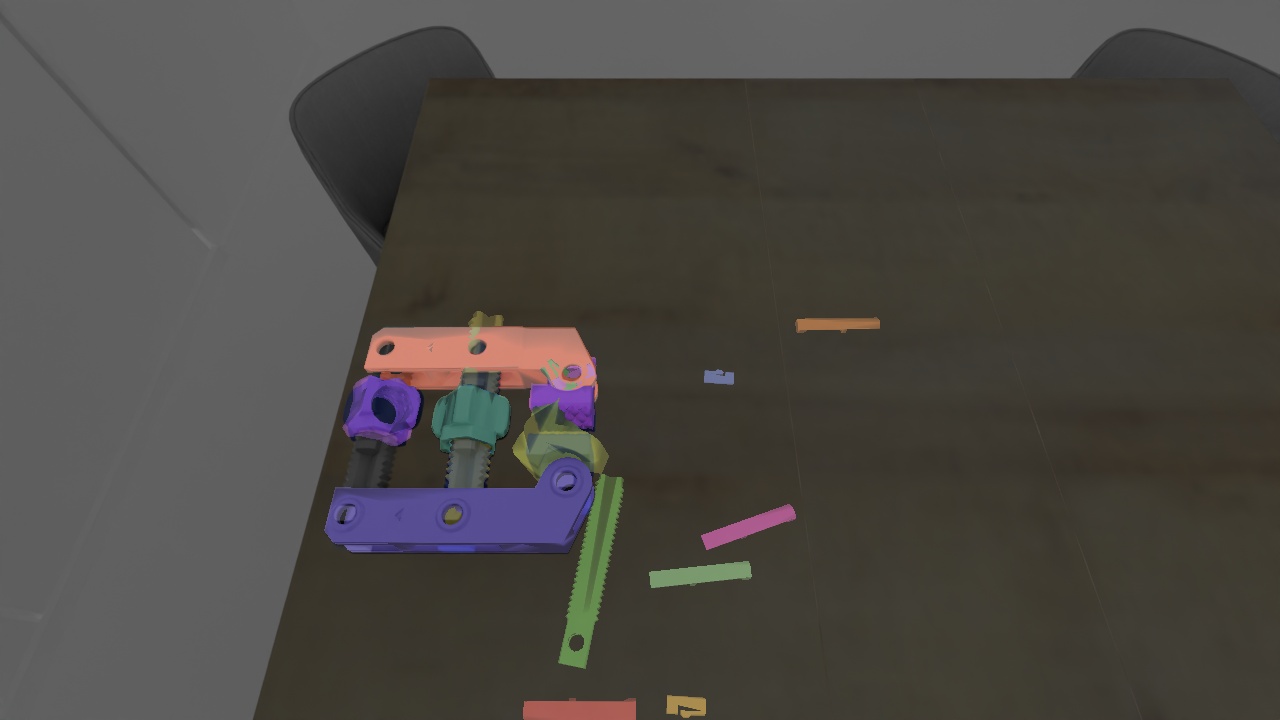}
            %\caption{HandScrewClamp (ICG)}
        \end{subfigure}
        \hfill
        \begin{subfigure}{0.12\textwidth}   
            \centering 
            \includegraphics[trim=10cm 5cm 22cm 8cm,clip,width=2.4cm, keepaspectratio]{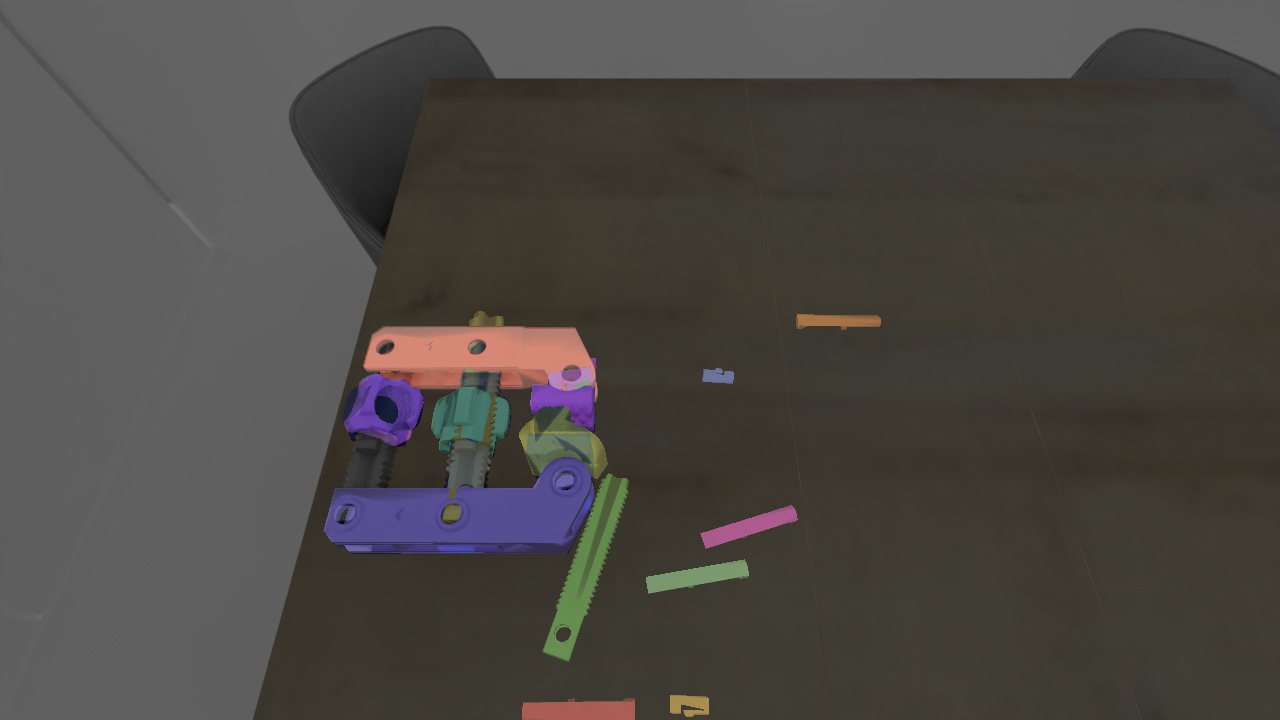}
            %\caption{HandScrewClamp (ICG)}
        \end{subfigure}
        \hfill
        \begin{subfigure}{0.12\textwidth}
            \centering 
            \includegraphics[trim=10cm 5cm 22cm 8cm,clip,width=2.4cm, keepaspectratio]{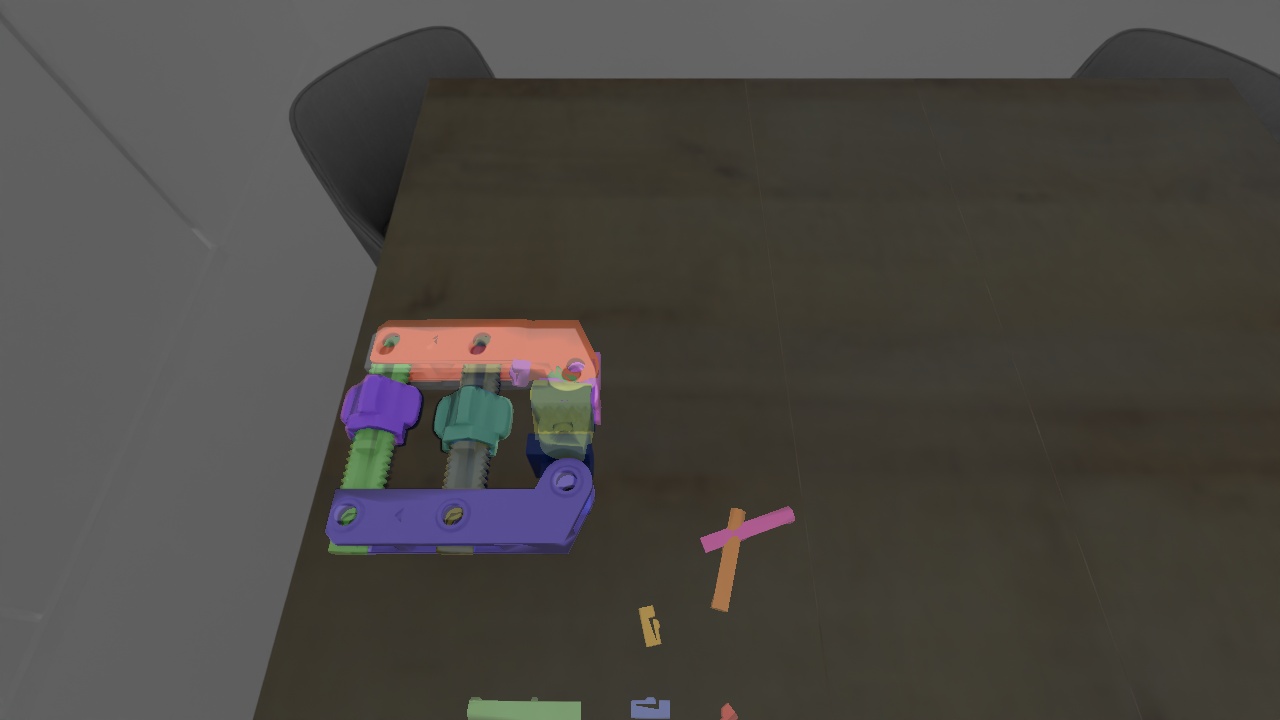} 
            %\caption{HandScrewClamp (GBOT)}
        \end{subfigure}
        
        \begin{subfigure}{0.12\textwidth}   
            \centering 
            \includegraphics[trim=10cm 15cm 24.3cm 0cm,clip,width=2.4cm, keepaspectratio]{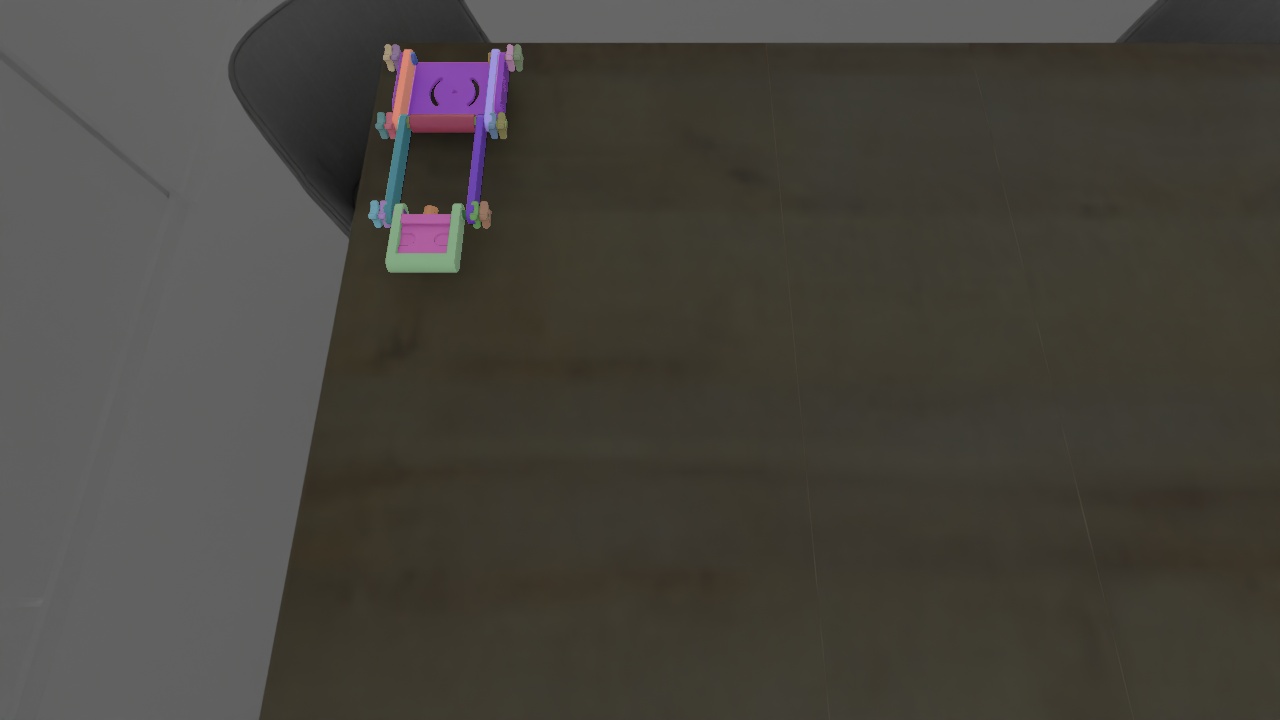}
            \caption{Ground truth}
        \end{subfigure}
        \hfill
        \begin{subfigure}{0.12\textwidth}   
            \centering 
            \includegraphics[trim=10cm 15cm 24.3cm 0cm,clip,width=2.4cm, keepaspectratio]{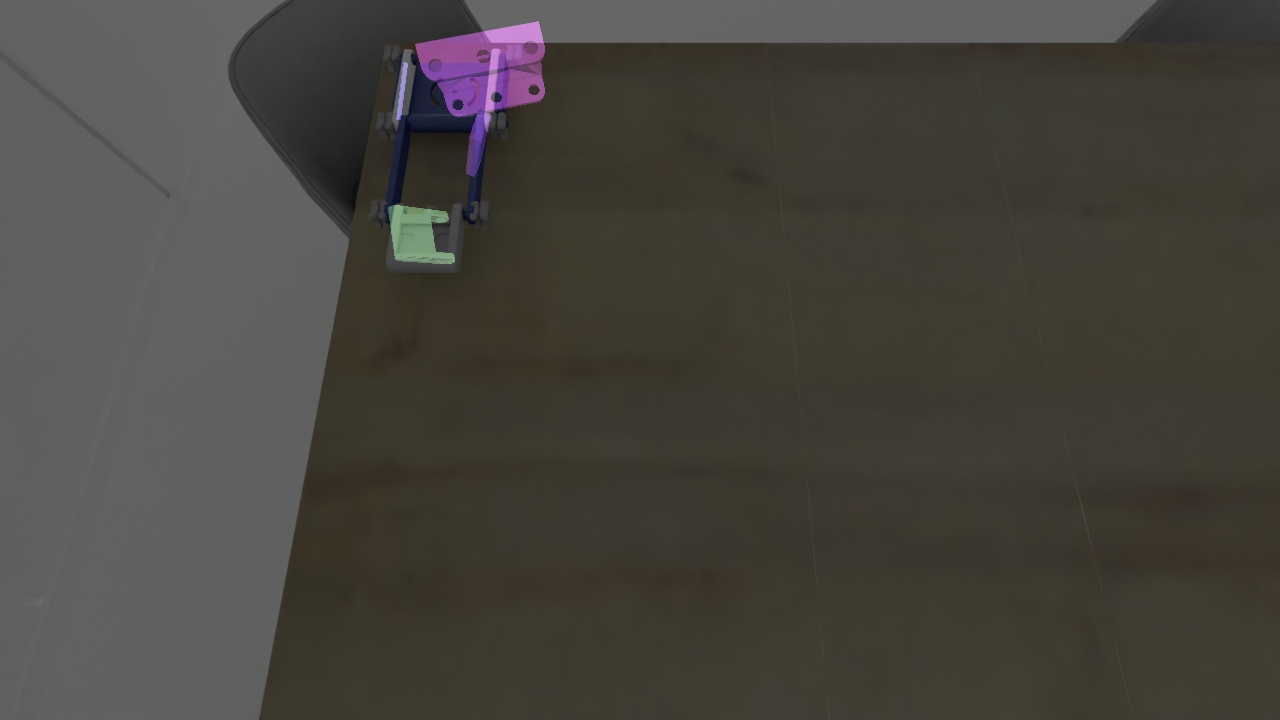}
            \caption{YOLOv8pose (w/o a)}
        \end{subfigure}
        \hfill
        \begin{subfigure}{0.12\textwidth}   
            \centering 
            \includegraphics[trim=10cm 15cm 24.3cm 0cm,clip,width=2.4cm, keepaspectratio]{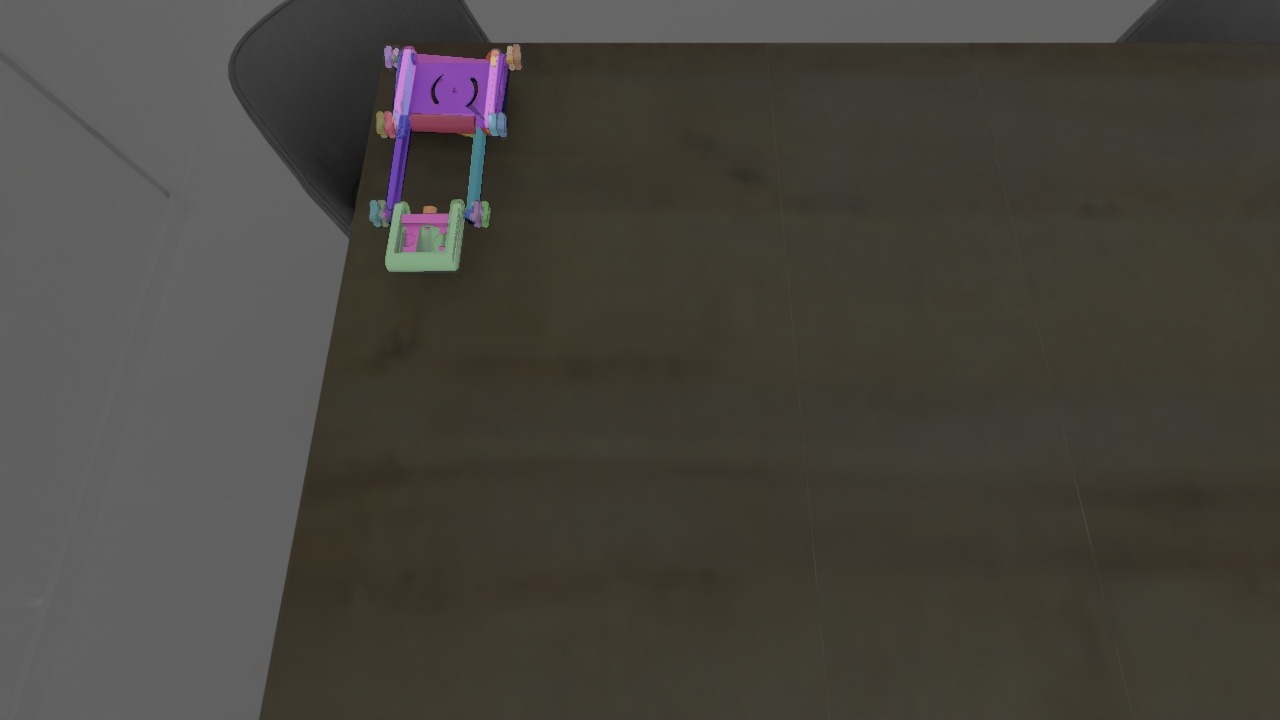}
            \caption{YOLOv8pose (w a)}
        \end{subfigure}
        \hfill
        \begin{subfigure}{0.12\textwidth}   
            \centering 
            \includegraphics[trim=10cm 15cm 24.3cm 0cm,clip,width=2.4cm, keepaspectratio]{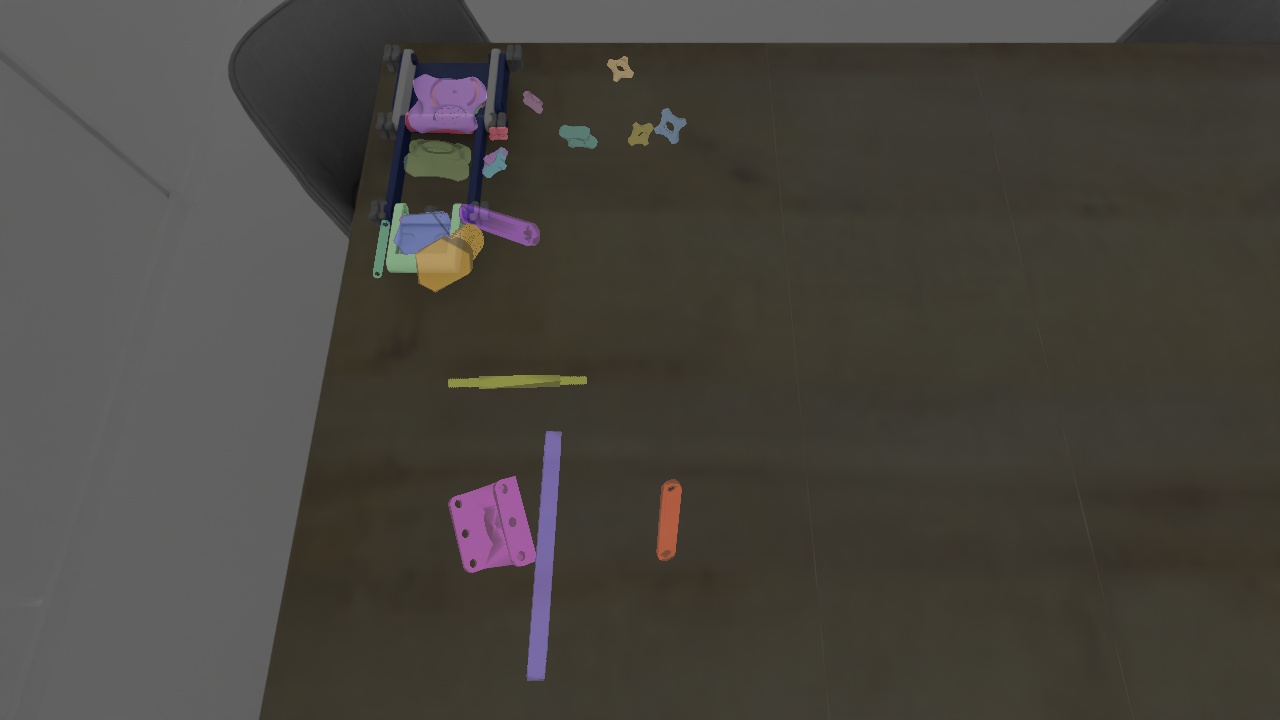}
            \caption{SRT3D}
        \end{subfigure}
        \hfill
        \begin{subfigure}{0.12\textwidth}   
            \centering 
            \includegraphics[trim=10cm 15cm 24.3cm 0cm,clip,width=2.4cm, keepaspectratio]{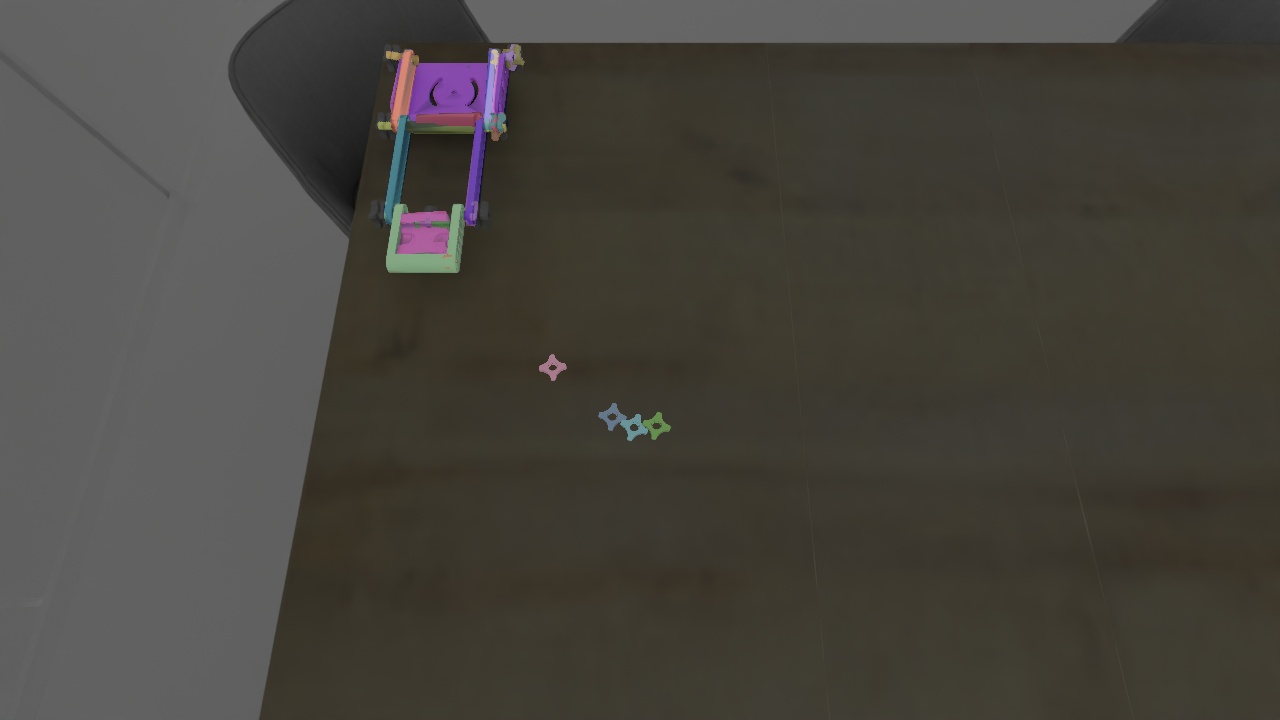}
            \caption{ICG}
        \end{subfigure}
        \hfill
        \begin{subfigure}{0.12\textwidth}   
            \centering 
            \includegraphics[trim=10cm 15cm 24.3cm 0cm,clip,width=2.4cm, keepaspectratio]{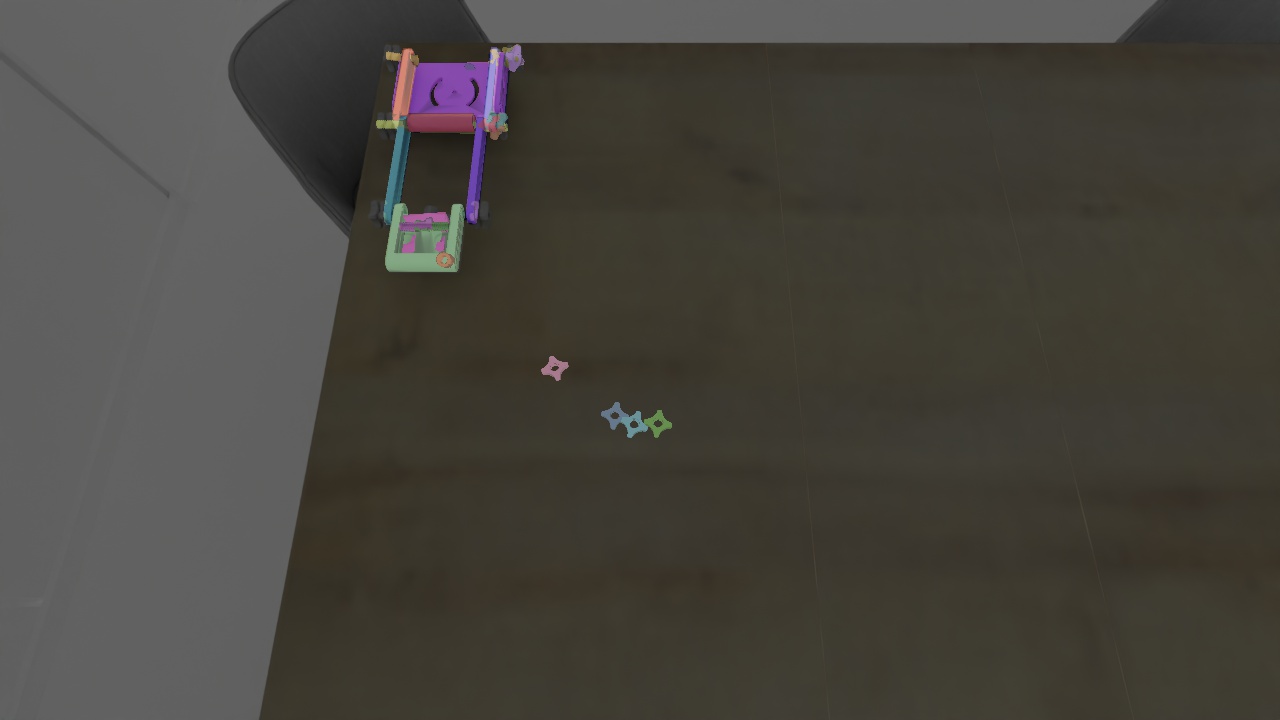}
            \caption{ICG+SRT3D}
        \end{subfigure}
        \hfill
        \begin{subfigure}{0.12\textwidth}   
            \centering 
            \includegraphics[trim=10cm 15cm 24.3cm 0cm,clip,width=2.4cm, keepaspectratio]{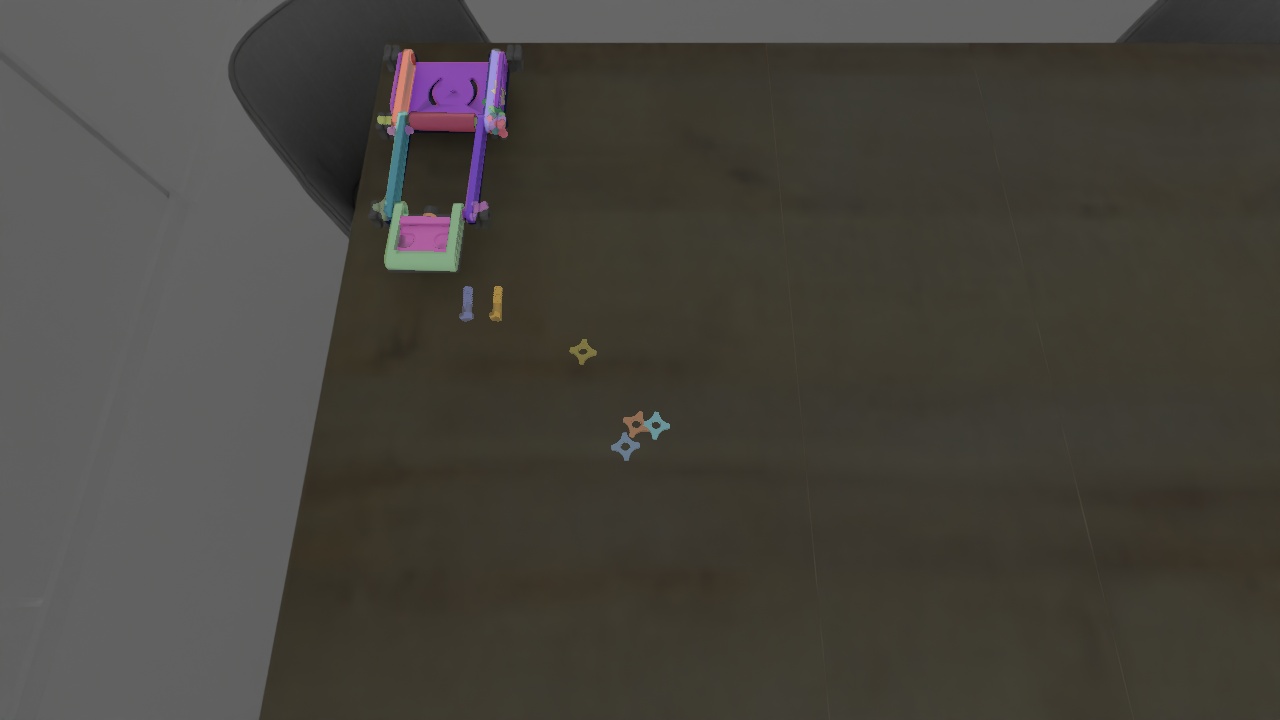} 
            \caption{GBOT}
        \end{subfigure}
        \caption{\textbf{Qualitative evaluation on the GBOT synthetic dataset.} We compared on the three assembly tools, Nano Chuck by Prima, Hand-Screw Clamp, and Liftpod (top to bottom). Tracked objects are colored individually. It becomes apparent that with a progressing assembly state, GBOT is more aware of the tracking than the state-of-art trackers.}      
        \label{fig:qualitative_eval}
\end{figure*}

\begin{figure*}[t!]
        \captionsetup[subfigure]{labelformat=empty}
        \centering
        
        \begin{subfigure}{0.16\textwidth}   
            \centering 
            \includegraphics[trim=5cm 3cm 10cm 0cm,clip, width=2.8cm, keepaspectratio]{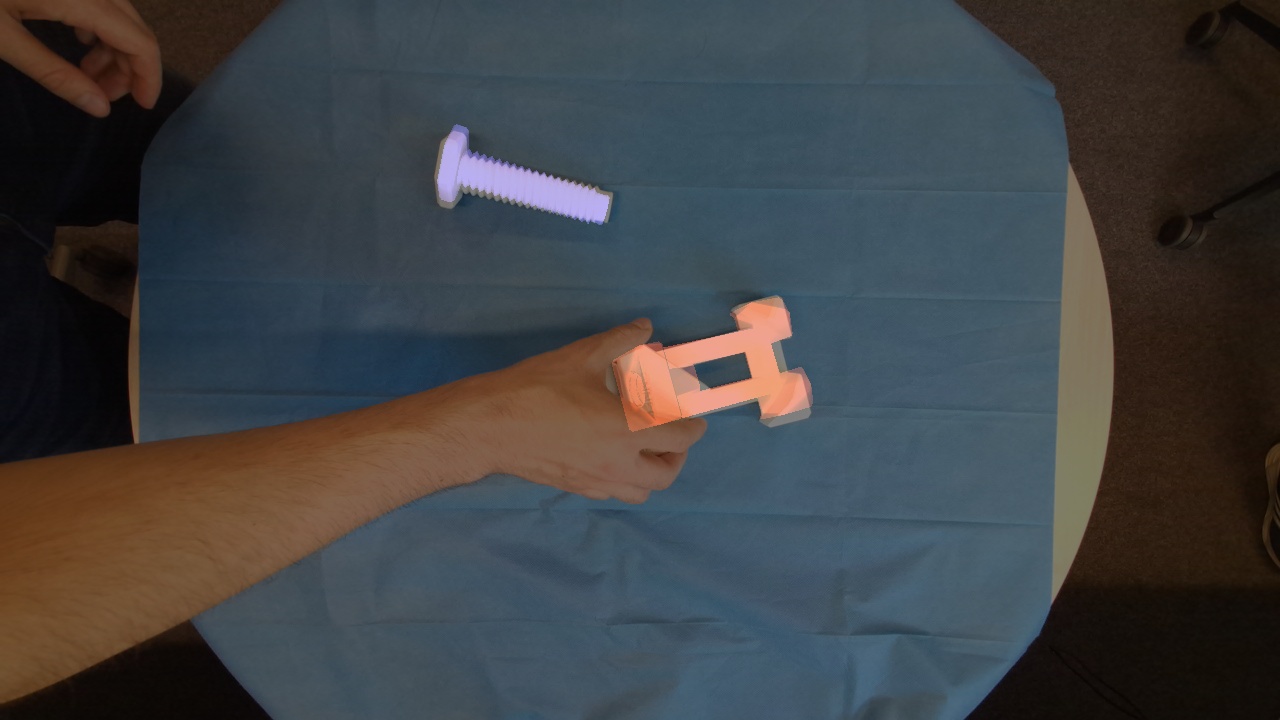}
            \caption{YOLOv8Pose (w a)}
        \end{subfigure}
        \hfill
        \begin{subfigure}{0.16\textwidth}   
            \centering 
            \includegraphics[trim=5cm 3cm 10cm 0cm,clip, width=2.8cm, keepaspectratio]{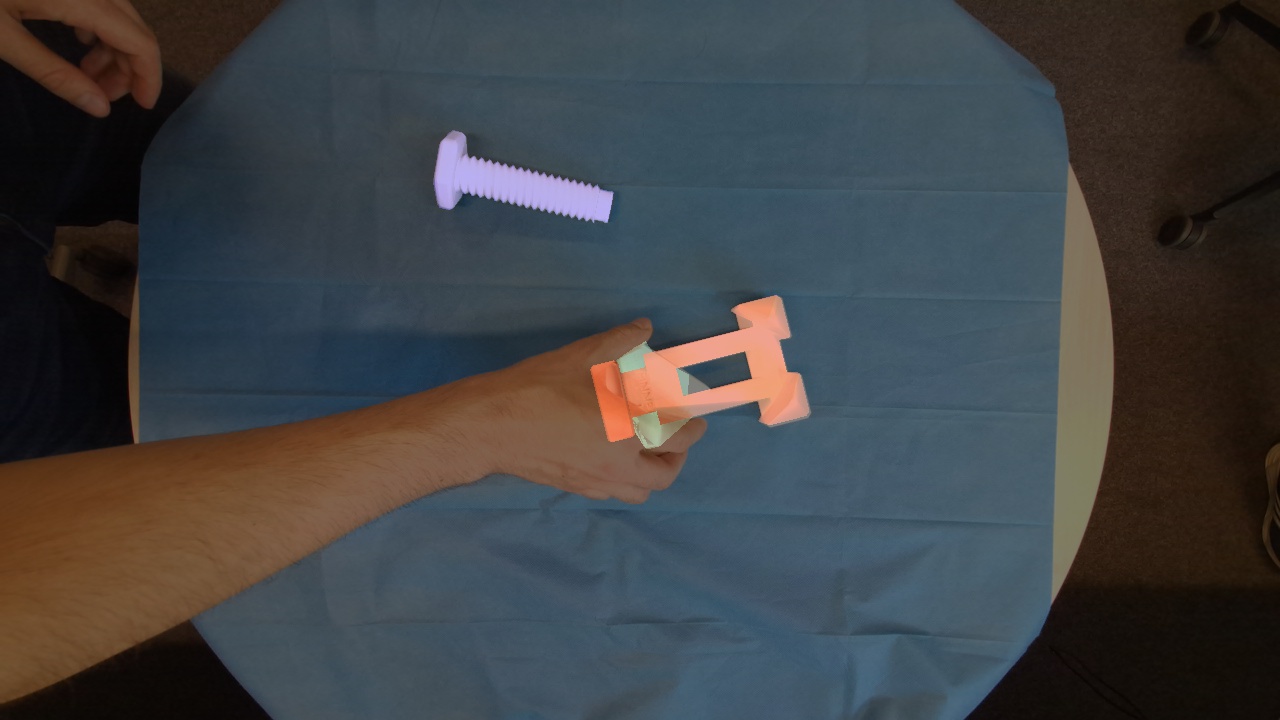}
            \caption{SRT3D}
        \end{subfigure}
        \hfill
        \begin{subfigure}{0.16\textwidth}   
            \centering 
            \includegraphics[trim=5cm 3cm 10cm 0cm,clip, width=2.8cm, keepaspectratio]{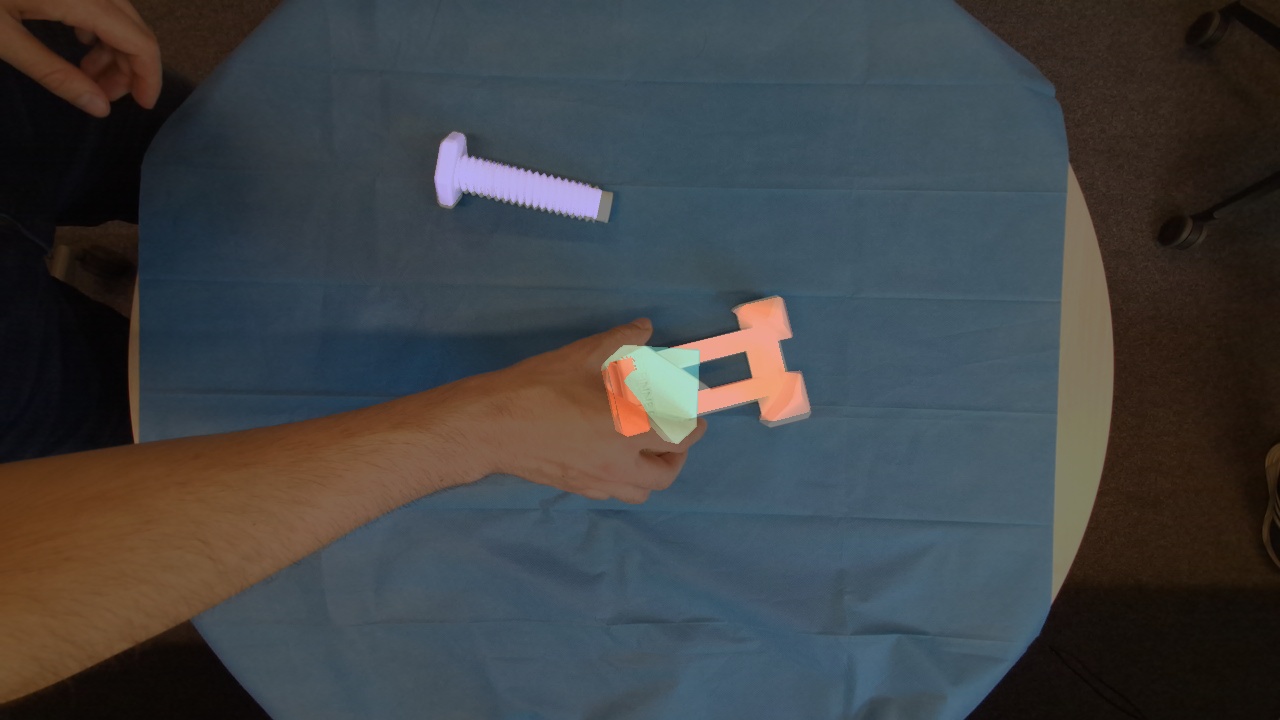}
            \caption{ICG}
        \end{subfigure}
        \hfill
        \begin{subfigure}{0.16\textwidth}   
            \centering 
            \includegraphics[trim=5cm 3cm 10cm 0cm,clip, width=2.8cm, keepaspectratio]{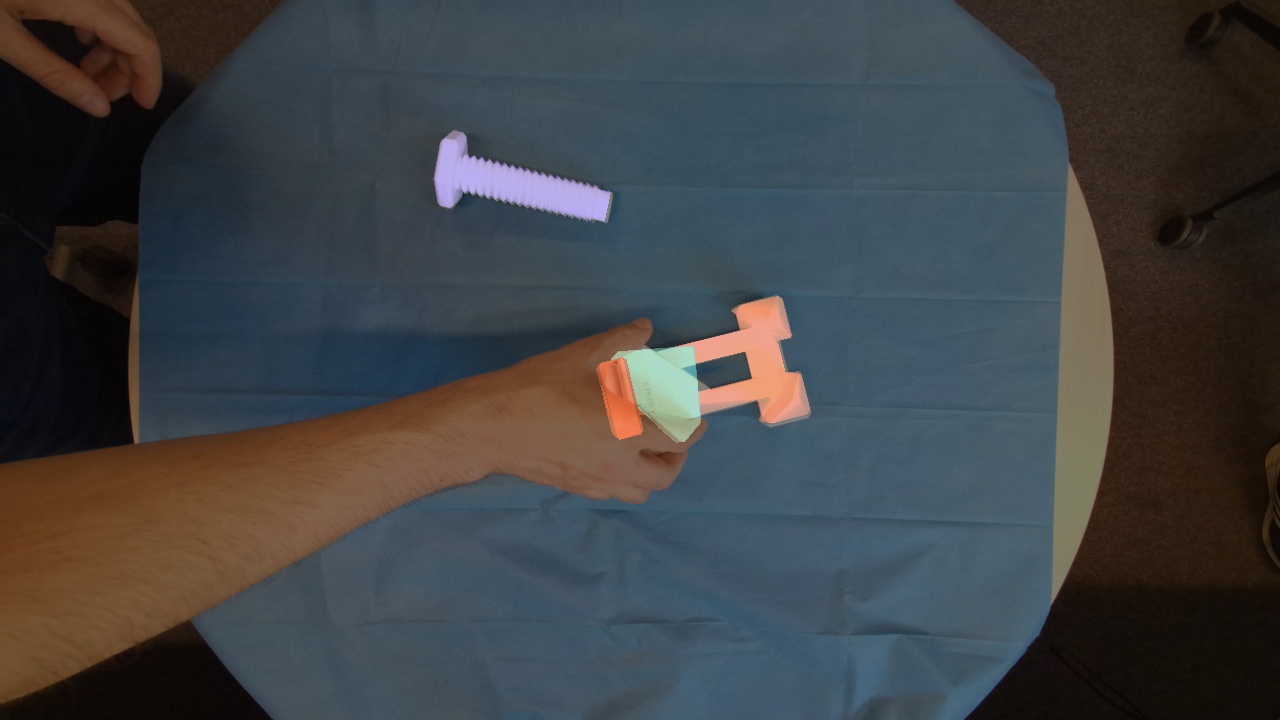}
            \caption{ICG+SRT3D}
        \end{subfigure}
        \hfill
        \begin{subfigure}{0.16\textwidth}   
            \centering 
            \includegraphics[trim=5cm 3cm 10cm 0cm,clip, width=2.8cm, keepaspectratio]{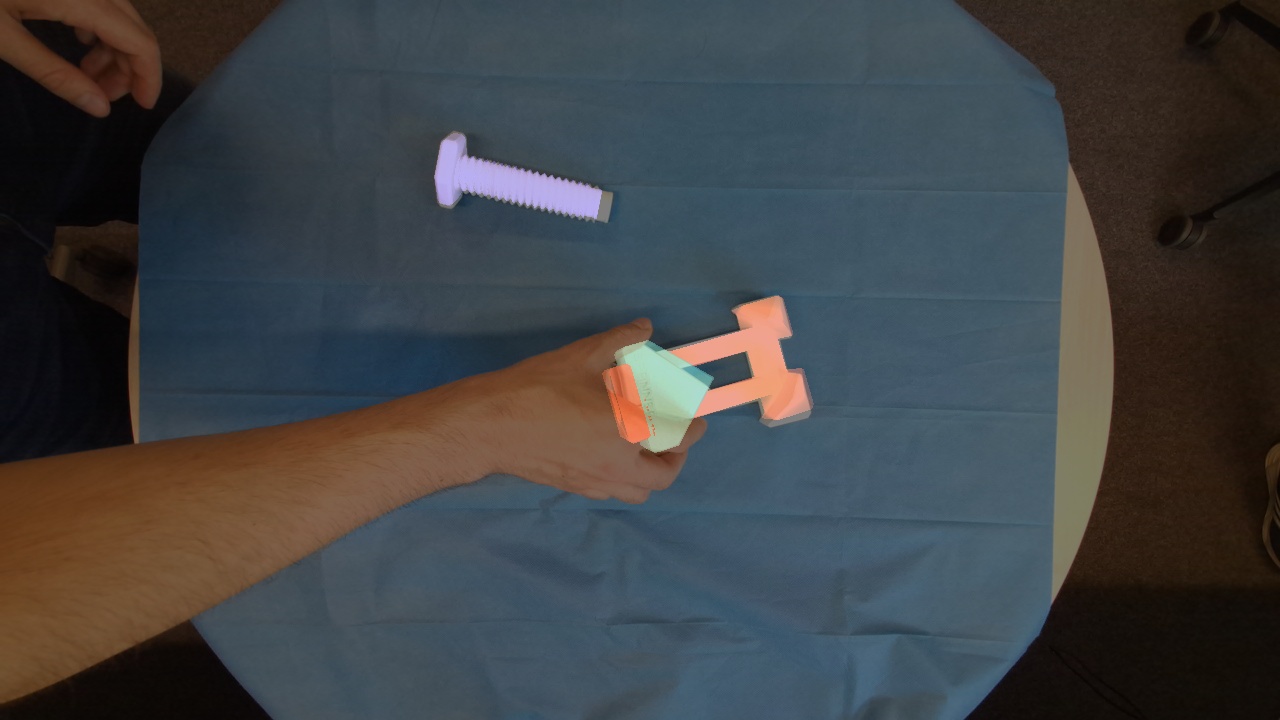}
            \caption{GBOT}
        \end{subfigure}
        \hfill
        \begin{subfigure}{0.16\textwidth}   
            \centering 
            \includegraphics[trim=5cm 3cm 10cm 0cm,clip, width=2.8cm, keepaspectratio]{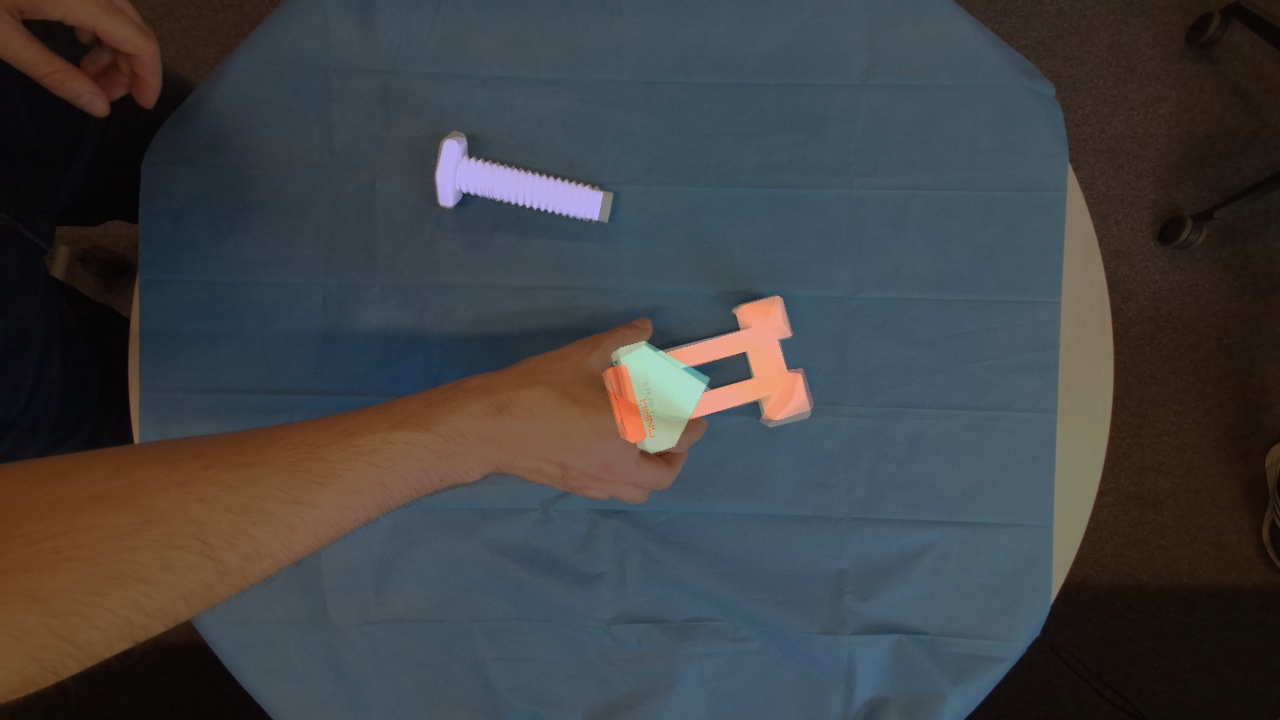} 
            \caption{GBOT + re-init}
        \end{subfigure}
        \caption{\textbf{Qualitative evaluation of GBOT compared to YOLOv8Pose, SRT3D, ICG, ICG+SRT3D and GBOT + re-init on a real scene.} We compared the assembly tool Hobby Corner Clamp with on the different methods. We show the tracked objects in individual colors. YOLOv8Pose cannot detect and estimate the pose of occluded assembly objects, while the tracking algorithm is still able to update object poses. It becomes apparent that with a progressing assembly state, GBOT is more aware of the tracking than the state-of-art trackers SRT3D, ICG and ICG+SRT3D.
        }      
        \label{fig:qualitative_eval_real}
\end{figure*}

The evaluation of the four different conditions of the \ac{gbot} dataset in \autoref{tab:quan_eval} shows that the overall accuracy of the tracking approaches is higher compared to the deep learning-based approach. 

The first assembly asset includes three assembly parts. The more parts an asset has, the more challenging the tracking is, see \autoref{tab:quan_eval}. For the Hobby Corner Clamp with three parts, GBOT achieves 100.0 for \ac{add}/\ac{add}-S in three conditions. The average translation error was in the best case $0.6$ cm and the rotation error was smaller than $3^{\circ}$. The last two assembly assets have more complex assembly parts than the first asset. Objects like bars or arms contain geometric ambiguity in some special views. 

The quantitative results show that \ac{gbot} outperforms SRT3D, ICG and ICG+SRT3D. Different from the results on a single-state dataset~\cite{xiang_posecnn_2017, li2023more}, our experiment shows that ICG+SRT3D~\cite{li2023more} does not outperform ICG when using multi-state assembly data. We achieved higher \ac{add}/\ac{add}-S values of 92.4, lower translation error of 2.6 cm and rotation error of $18.3^{\circ}$ on average. GBOT + re-init, updates the tracking regularly by 6D pose estimation, has the accuracy between YOLOv8Pose and GBOT in synthetic data, with \ac{add}/\ac{add}-S values of 91.3. We can observe beneficial results on occluded scenes as indicated by the evaluation on the hand conditions for all assembly objects. On real-scenes we also observed that objects which suffer from heavy occlusion during the assembly can benefit from the re-initialization, see \autoref{fig:failcases}.

Regarding the runtime, our YOLOv8Pose is about 35~ms slower than our object tracking, see \autoref{tab:runtime}. SRT3D is the fastest algorithm with 11.87~ms in average considering only RGB tracking. ICG is 15.88~ms slower than SRT3D due to introduction of depth modality. ICG+SRT3D takes extra 11~ms for pose refinement. The latency of our GBOT is about 0.57~ms larger than the state of art object detector ICG. GBOT + re-init, takes overall 36.89~ms per frame. It needs less than 3~ms for one single object. Overall, GBOT and the extension GBOT + re-init are both suitable for real-time applications.

%%%%%% Bernhard's comments on the evaluation %%%%%%

\subsection{Qualitative Evaluation}

\begin{figure}[t!]
    \centering

    \begin{subfigure}{0.495\columnwidth}
        \centering
        \includegraphics[trim=8cm 2cm 6cm 3cm,clip, height=2.5cm, keepaspectratio]{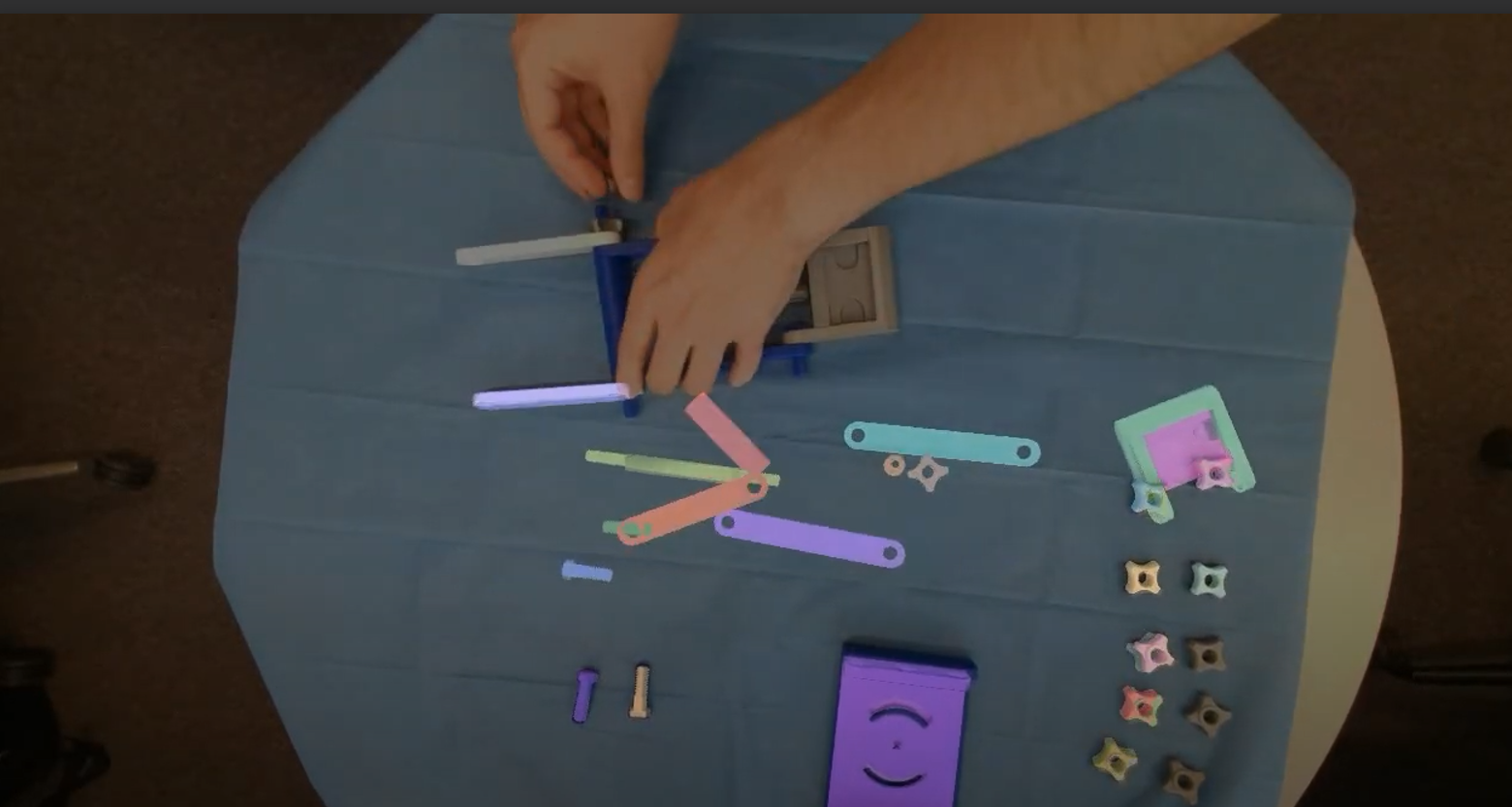}
        \caption{Fail case (GBOT + re-init)}
    \end{subfigure}
    \hfill
    \begin{subfigure}{0.495\columnwidth}
        \centering
        \includegraphics[trim=8cm 2cm 6cm 3cm,clip, height=2.5cm, keepaspectratio]{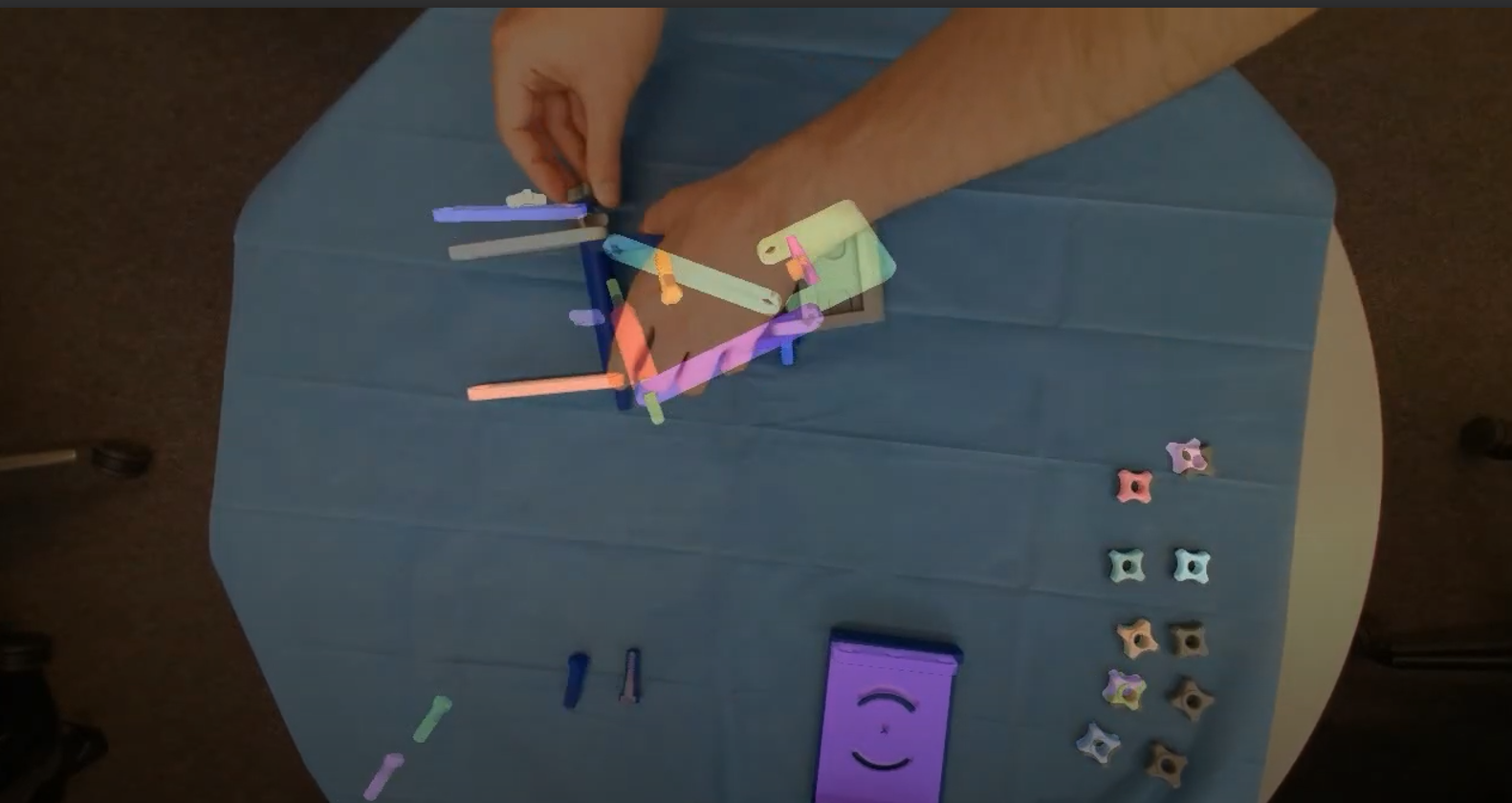}
        \caption{Fail case (GBOT)}
    \end{subfigure}

        \begin{subfigure}{0.495\columnwidth}
        \centering
        \includegraphics[trim=3cm 0cm 5cm 0cm,clip, height=2.5cm, keepaspectratio]{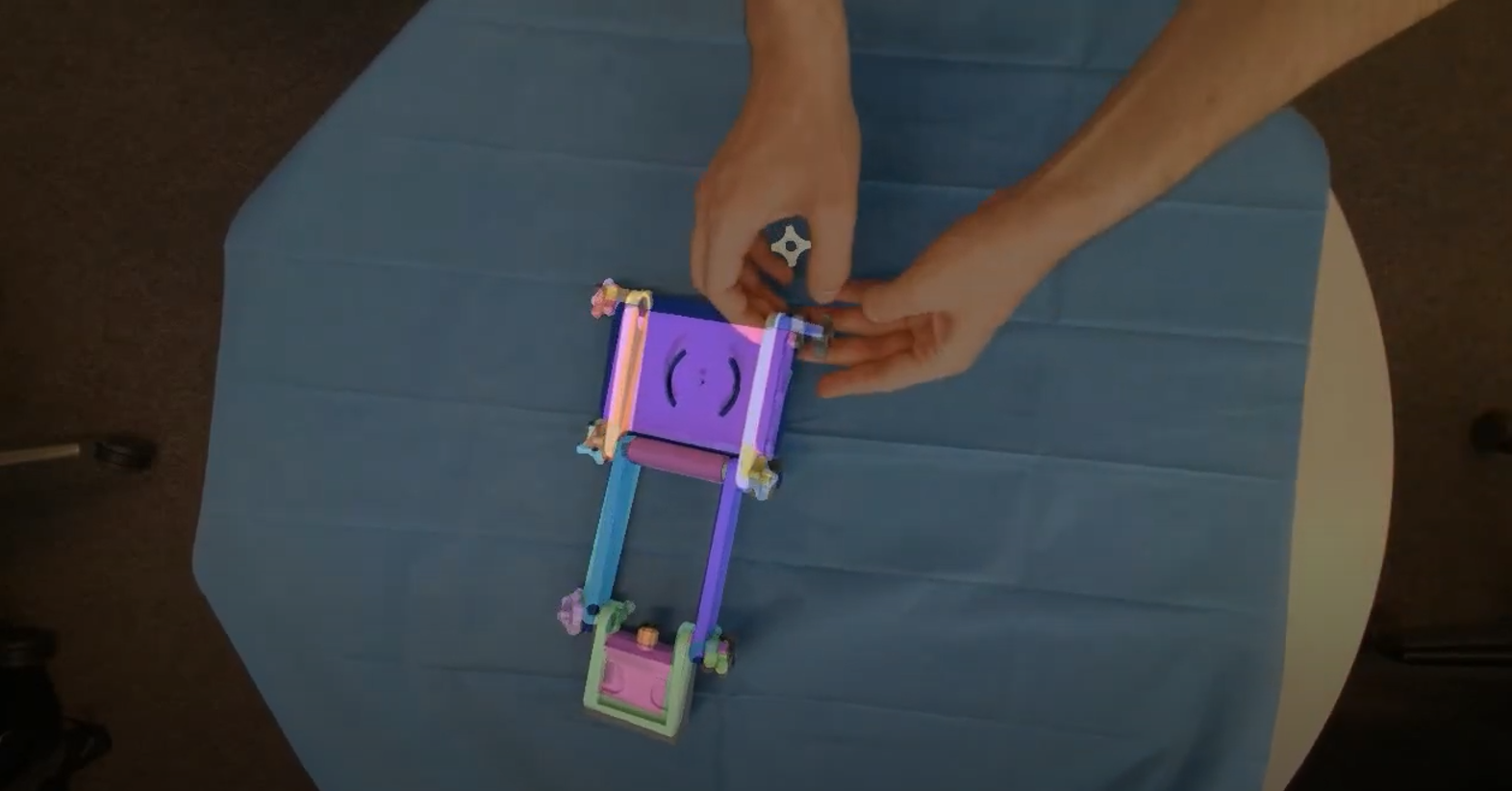}
        \caption{Success case (GBOT + re-init)}
    \end{subfigure}
    \hfill
    \begin{subfigure}{0.495\columnwidth}
        \centering
        \includegraphics[trim=3cm 0cm 5cm 0cm,clip, height=2.5cm, keepaspectratio]{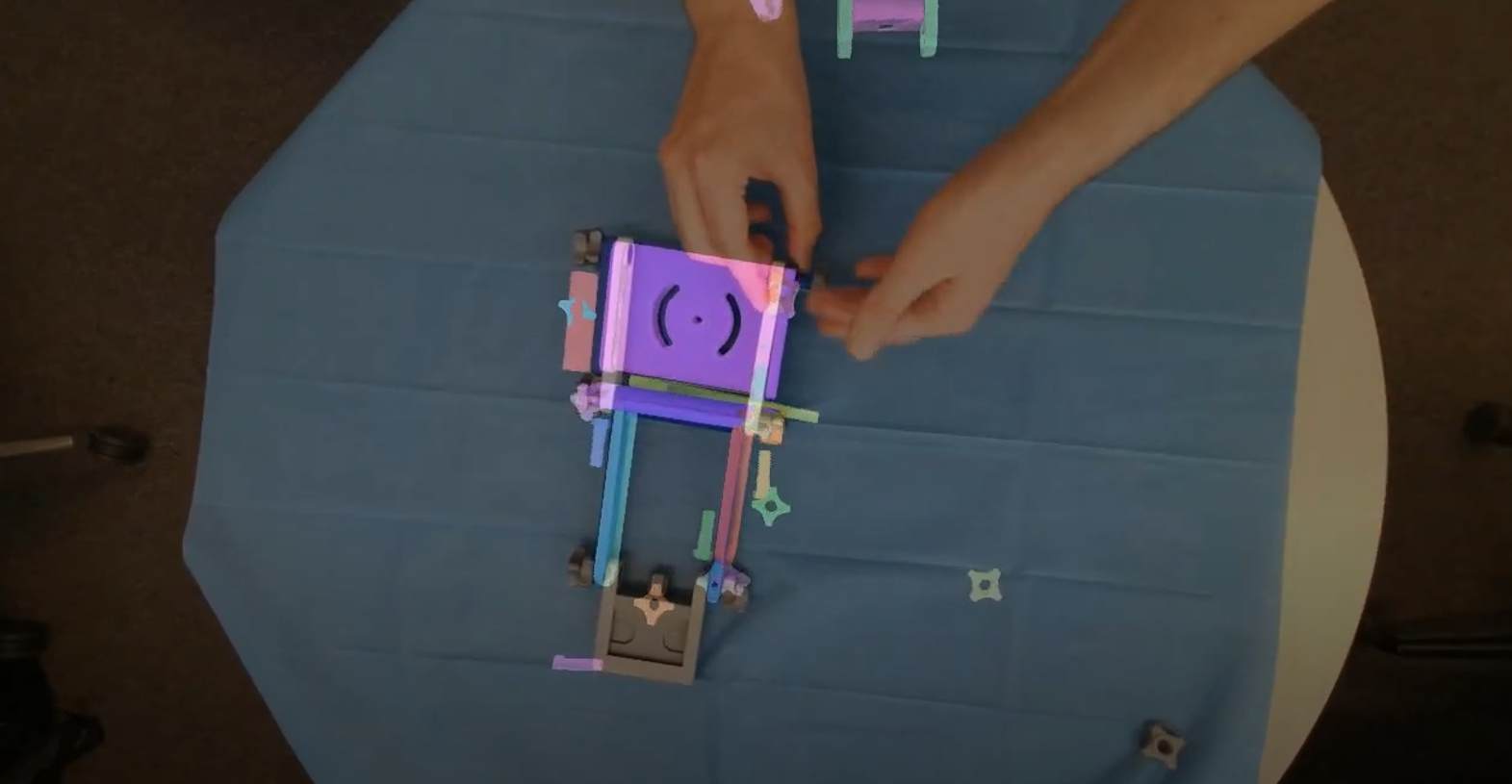}
        \caption{Fail case of (GBOT)}
    \end{subfigure}
    \caption{\textbf{Possible fail cases of GBOT due to hand occlusion during real assembly process.} Our tracking approach has significant improvement compared to the state-of-art approach. However, after heavy occlusion (top), the tracking without re-initialization suffers.
}
    \label{fig:failcases}
\end{figure}

To compare our approach with state-of-the-art tracking approaches, we used the metric-based comparison in \autoref{tab:obj_tracking} and compared qualitatively on synthetic and real scenes. We show the results on the synthetic scenes in \autoref{fig:qualitative_eval}. Qualitative results on the real-3D printing parts are shown in \autoref{fig:failcases} and \autoref{fig:qualitative_eval_real}. \autoref{fig:qualitative_eval} shows characteristic visualizations of the tracking for the three assembly assets. As shown in \autoref{fig:qualitative_eval},  \ac{gbot} is more robust compared with YOLOv8Pose and common tracking approaches. \ac{gbot} can especially track smaller parts. These parts are not lost during the assembly process based on the links to the frame parts. In the case of strong hand occlusion, the tracking suffers when the tracked object is highly occluded by a hand.

The tracking benefits from re-initialization in heavy occluded scenes, see \autoref{tab:quan_eval} and \autoref{fig:failcases}. Although our tracking approach works well with assembly assets, \ac{gbot} can also have fail cases. We show a qualitative breaking analysis in \autoref{fig:failcases}.

\subsection{Real Scenes}
Besides evaluating on synthetic scenes, we perform additional experiments on real data to show the effectiveness of our approach.

\begin{figure}
    \centering
        \centering
         \begin{subfigure}{0.49\columnwidth}
            \centering
            \includegraphics[trim=3cm 0cm 5cm 0cm,clip, height=2.5cm, keepaspectratio]{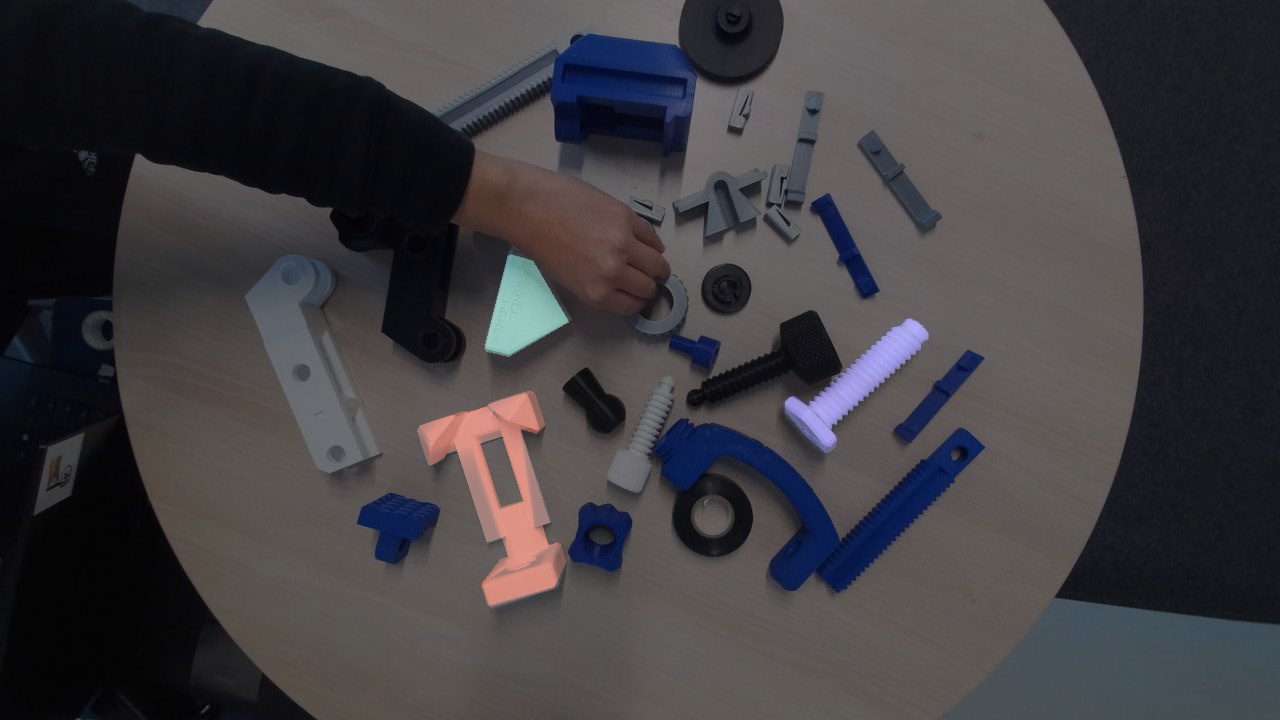}
            \caption{Hobby Corner Clamp}
        \end{subfigure}
        \hfill
         \begin{subfigure}{0.49\columnwidth}
            \centering
            \includegraphics[trim=3cm 0cm 5cm 0cm,clip, height=2.5cm, keepaspectratio]{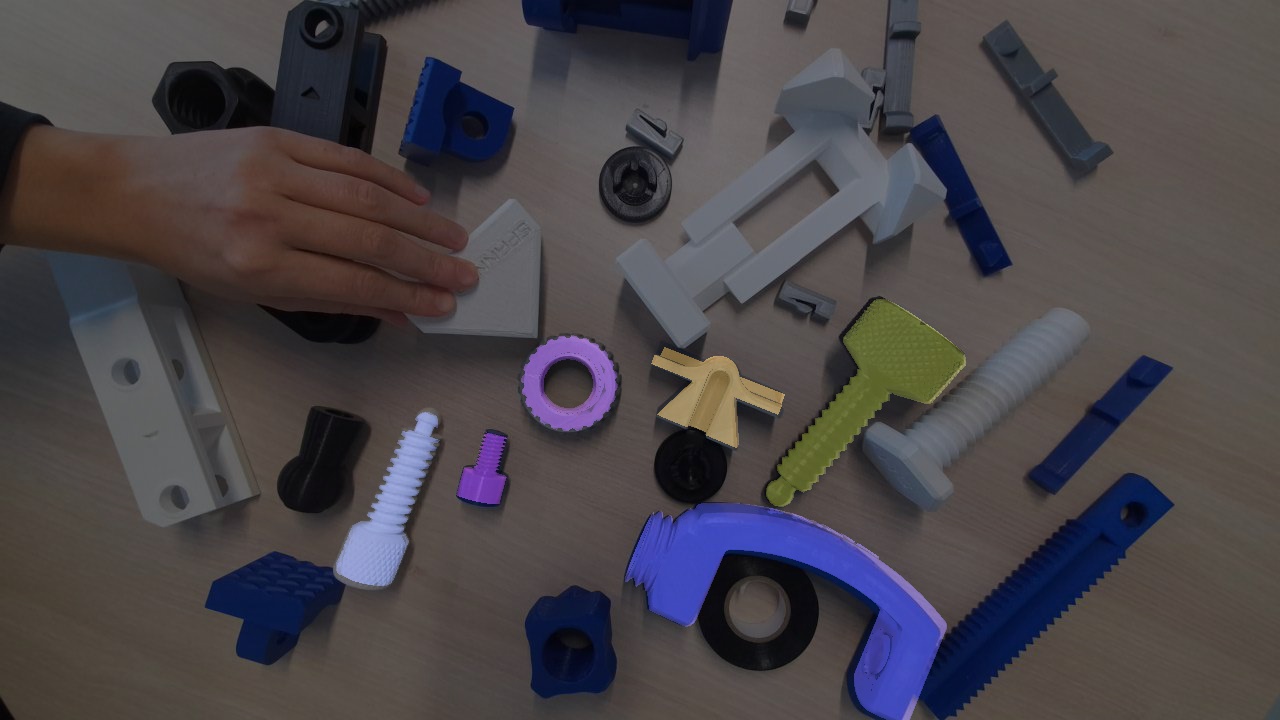}
            \caption{Nano Chuck by PRIma}
        \end{subfigure}
    \caption{\textbf{Evaluation on real cluttered scenes}: We randomly place GBOT assembly assets together with some distract objects to test the influence of cluttered scenes. Our training data with domain randomization helps to detect objects in the cluttered scenes.
}
    \label{fig:ablation_clutter}
\end{figure}

\textbf{Cluttered scenes with YOLOv8pose:} In each cluttered scene we select one assembly asset from the \ac{gbot} dataset and other 3D printing parts and randomly place them on a table. We find our YOLOv8pose can still robustly detect objects in clustered scenes, as shown in \autoref{fig:ablation_clutter}.

\begin{figure}
    \centering
        \centering
         \begin{subfigure}{0.49\columnwidth}
            \centering
            \includegraphics[trim=2cm 5cm 13cm 0cm,clip, height=2.5cm, keepaspectratio]{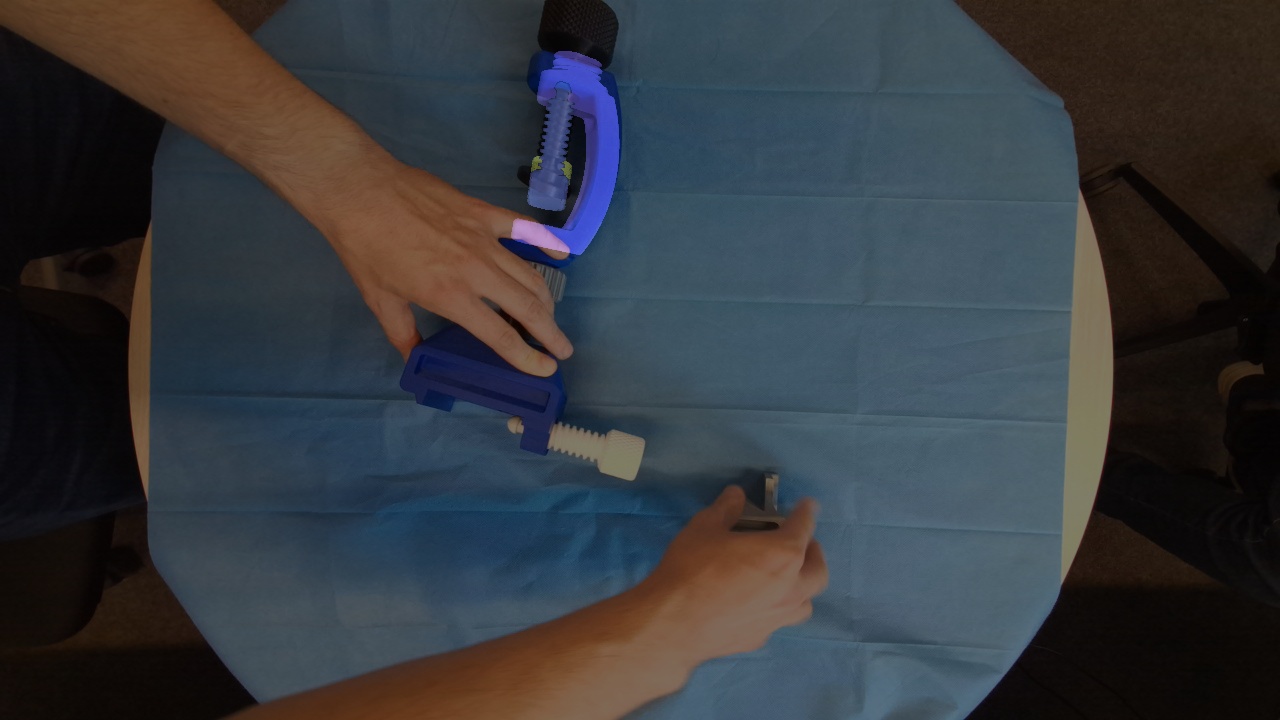}
            \caption{YOLOv8Pose w/o assembly data}
        \end{subfigure}
        \hfill
         \begin{subfigure}{0.49\columnwidth}
            \centering
            \includegraphics[trim=2cm 5cm 13cm 0cm,clip, height=2.5cm, keepaspectratio]{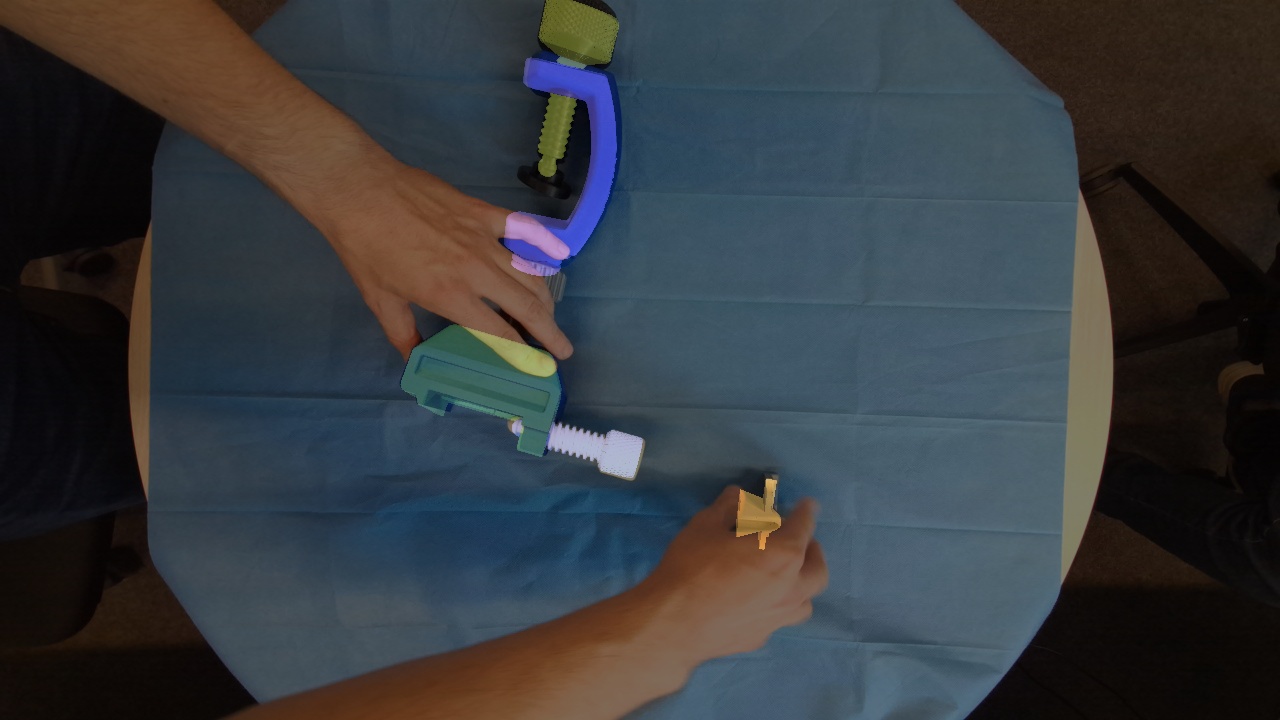}
            \caption{YOLOv8Pose w assembly data}
        \end{subfigure}
    \caption{\textbf{Assembly-aware training on synthetic scenes and evaluation on a real scene}: Our training data with assembly data helps to overcome occlusion during the assembly process.
}
    \label{fig:ablation_assembly}
\end{figure}

\textbf{Effect of assembly training data:} To validate the positive effects of assembly-aware training, we evaluate YOLOv8pose with and without assembly training, see \autoref{tab:quan_eval}. \autoref{fig:ablation_assembly} shows the qualitative comparison using training data with and without assembly assets. This can explain the improvement in \autoref{tab:quan_eval}. Compared with training of separate parts, training data with assembly parts can enhance the detection performance. Assembly assets usually include self-occlusion of separate parts. Considering assembled objects helps to reduce the domain gap between synthetic training data and real test data.

\begin{figure}[t!]
    \centering
        \centering
         \begin{subfigure}{0.49\columnwidth}
            \centering
            \includegraphics[trim=2cm 5cm 13cm 0cm,clip, height=2.5cm, keepaspectratio]{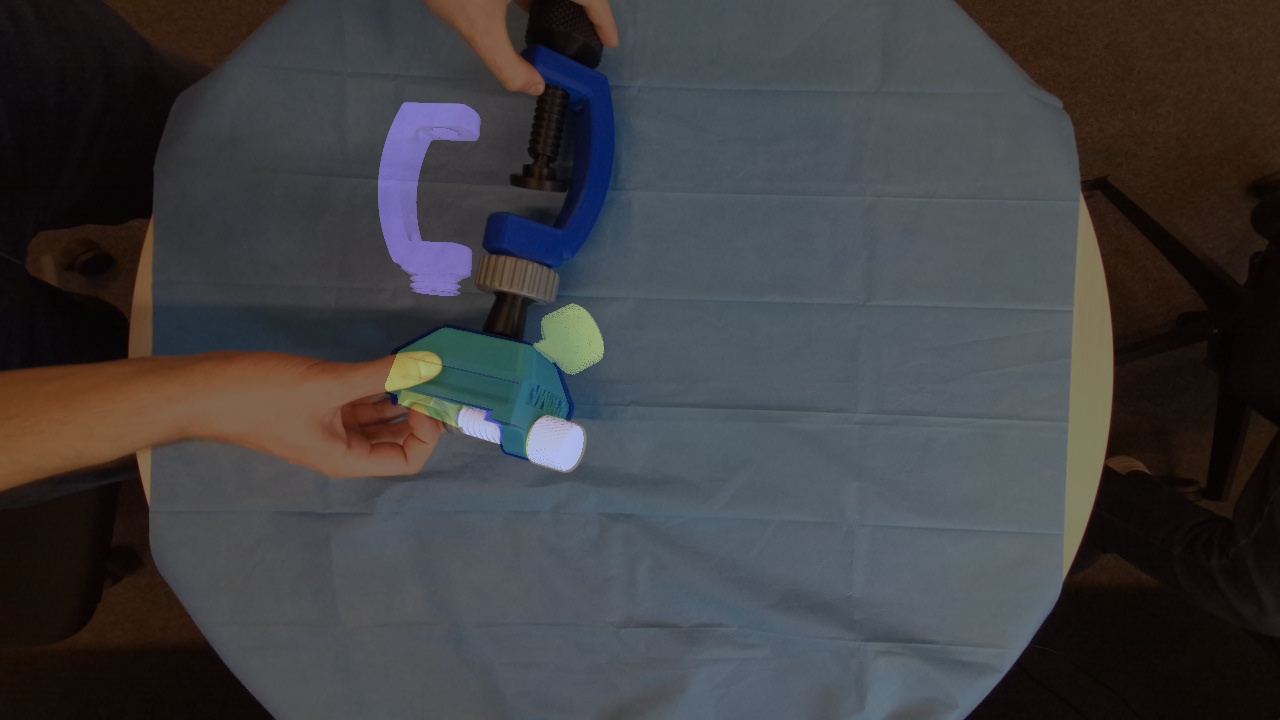}
            \caption{ICG\cite{stoiber_iterative_2022}}
        \end{subfigure}
         \begin{subfigure}{0.49\columnwidth}
            \centering
            \includegraphics[trim=2cm 5cm 13cm 0cm,clip, height=2.5cm, keepaspectratio]{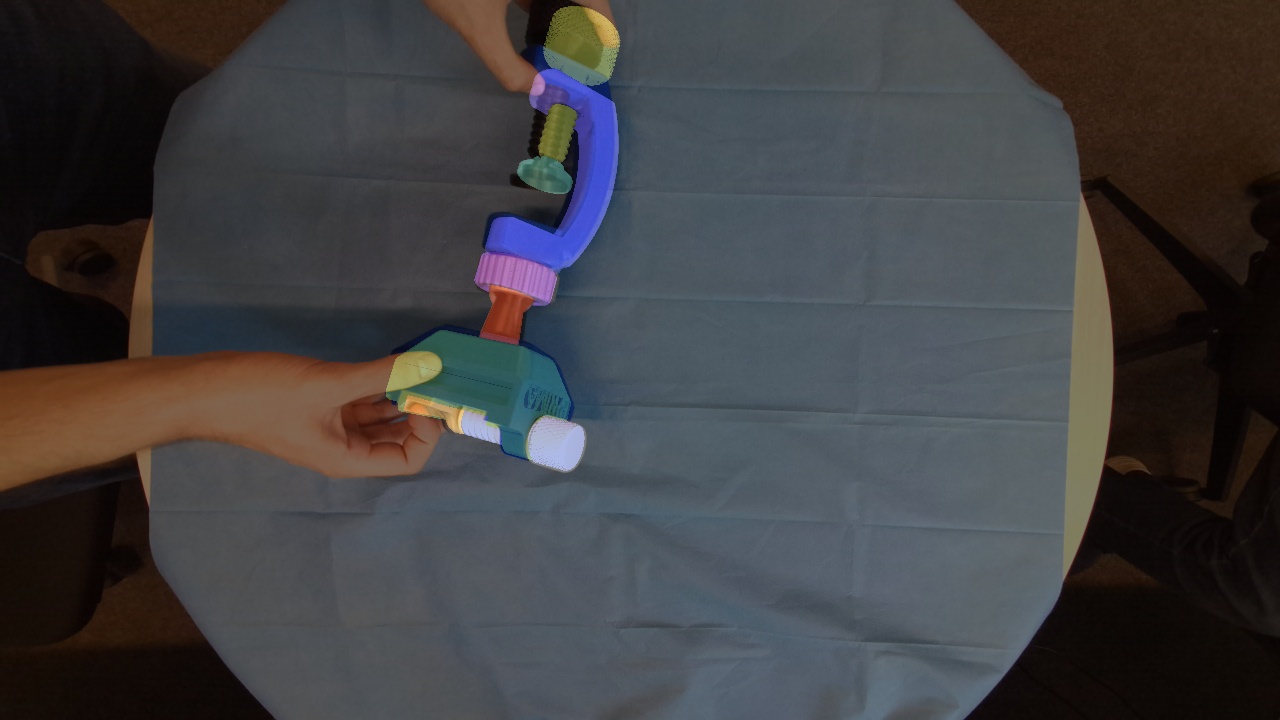}
            \caption{GBOT}
        \end{subfigure}
    \caption{\textbf{Tracking with assembly-state-graph in a real scene}: We extend our assembly-aware tracking based on work of MbICG\cite{stoiber_multi-body_2022}. The introduction of assembly state graph helps improve tracking accuracy considering assembly constraints.
}
    \label{fig:ablation_tracking}
\end{figure}

\textbf{Effect of assembly-state-graph on tracking:} In \autoref{fig:ablation_tracking}, we study the qualitative results of tracking with and without assembly graph. Using assembly constraints, the combined tracking is more accurate than separate tracking. This qualitative comparison is in accordance with the quantitative results in \autoref{tab:quan_eval}. Tracking with assembly constraints is more robust due to prior knowledge of relative poses between parts. If one of the separate tracking is lost, it can still be recovered from poses of other parts.

\subsection{Augmented Reality Integration}
\begin{figure}[t!]
     \centering
     \includegraphics[width=0.35\textwidth]{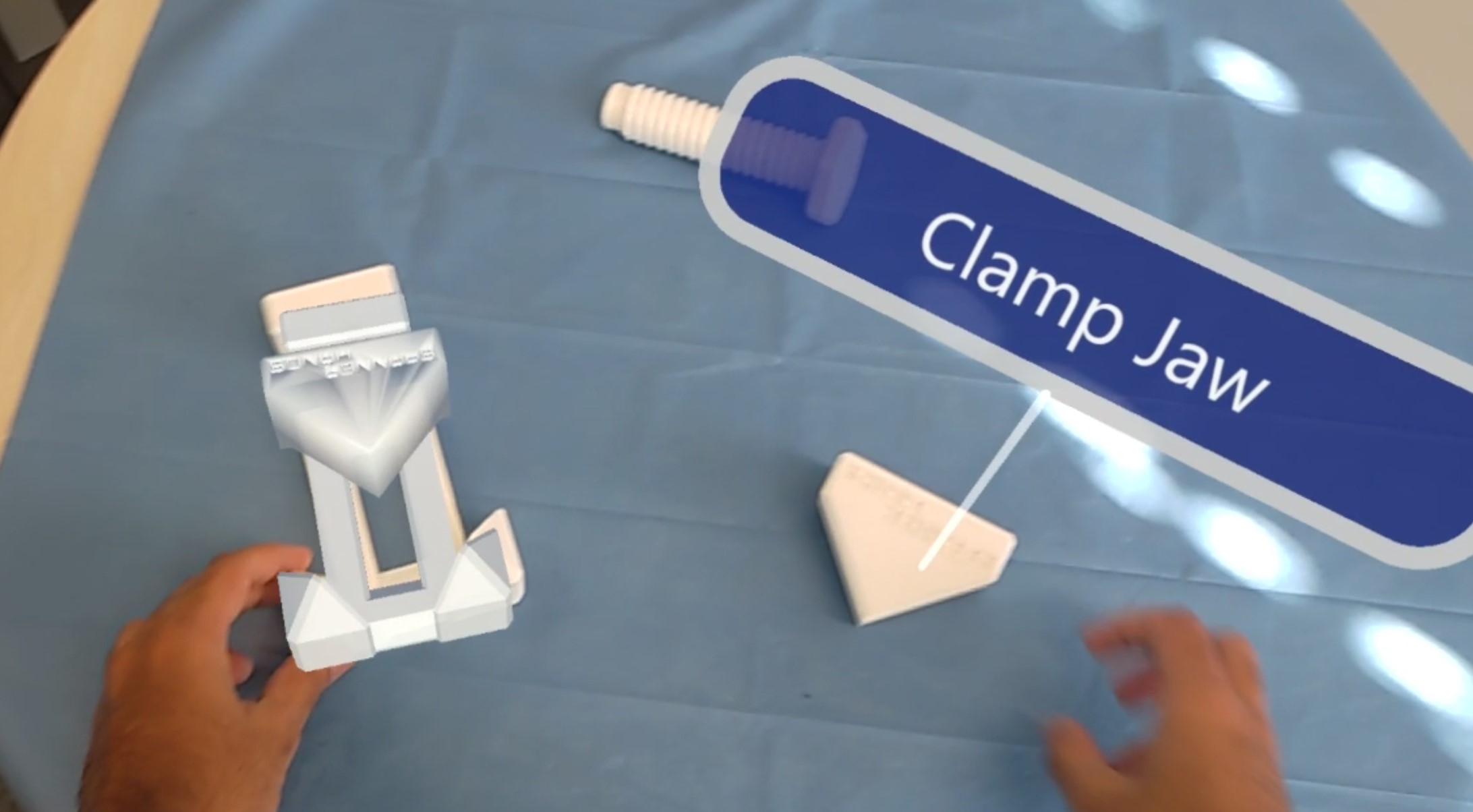}

     \caption{\textbf{Augmented reality integration}: Our GBOT tracking algorithm is able to feed a \ac{hmd} with 6D object pose information, enabling realtime visualization overlays or tooltips on real assembly parts.}
     \label{fig:demo}
\end{figure}

\ac{gbot} is able to be deployed in real-time to ensure its use in \ac{ar} applications. To prove this, we show an \ac{ar} application assembly guidance example of 3D printing parts from the GBOT dataset. As shown in \autoref{fig:demo}, GBOT enables \ac{ar} \acp{hmd} to visualize tooltips or overlays on real objects. Currently, we provide a client-server architecture to handle computational costly tracking on the server side and lightweight \ac{ar} visualization on the client side. The GBOT framework contains a RESTful API, to request the captured poses of our tracked objects by mobile devices. This pipeline can be utilized to visualize assembly instructions and guidance, as shown in \autoref{fig:demo}. For the current demo we used a Microsoft Hololens 2.

\section{Discussion}

\ac{gbot} utilizes a deep learning-based 6D pose estimation to initialize the tracking. It shows more robust results compared to purely deep-learning based 6D pose estimation and the state-of-the-art tracking. 

Our evaluation shows that pose estimation without assembly training data only works well at the beginning of tracking without assembly occlusion, but fails during the assembly process. With more context-aware training data, the performance of YOLOv8Pose in an assembly scene can be improved. Nevertheless, \ac{gbot} has more accurate results compared to YOLOv8Pose. This can be explained by the occlusion during the assembly process and the objects are difficult to be detected during each assembly step. However, the tracking utilizes the information from previous timestamps. With ongoing assembly states, as visualized in our quantitative results, the tracking works more robustly. Including the links between the individual objects showed better results than using an individual tracking like ICG~\cite{stoiber_iterative_2022}. % Continuous and stable tracking is essential for robust \ac{ar} applications.

% Regarding to the runtime, SRT3D is the fastest approach with about 12~ms per frame, ICG takes about 16~ms more than SRT3D. GBOT takes 0.57~ms more than ICG considering our proposed assembly graph. \ac{gbot} requires a little more time due to the time of calculating relative poses and determining assembly state. However, \ac{gbot} is more robust within the ongoing assembly process.

Generally, tracking smaller objects is difficult. Our strategy is to focus on the main assembly parts, the frame and base parts of the assembly objects and then link smaller objects to the frame parts according to the new assembly state information. This approach helps to overcome losing the tracking during the assembly process. Another important influence factor is the hand occlusion and dynamic light conditions. There are some fail cases, when the tracking is lost during the process and the assembly state cannot be switched to the next step. When the objects are fully occluded by hands or light condition changes swiftly, the tracking suffers. This can be supported by the re-initialization. As shown in Table~\ref{tab:obj_tracking}, with occlusion (hand), the re-initialization is beneficial. %the performance in hand is lowest in almost all assembly assets. 

Nevertheless, \ac{gbot} performs well in tracking assembly parts and we outperform the individual tracking (ICG) on the \ac{gbot} dataset. Although the performance in situations of hand occlusion, dynamic light and motion blur is lower than in the normal scene condition, the tracking is still more robust compared to the state-of-the-art. Especially, when looking at qualitative results on synthetic and real scenes, the robustness of our tracking becomes apparent. 

\subsection{Limitations}

In terms of deep learning-driven driven keypoint detection, symmetric objects are particular challenging. At the same time, smaller objects are challenging for object detectors like YOLO. With deeper convolutions, the smaller parts are at some point not anymore a part of the produced feature map of the network. Moreover, for our YOLO-based approach, we found limitations in occluded scenarios and dynamic light settings. In occluded scenes a re-initialization can be beneficial as indicated by our ablation study using GBOT + re-init. We implement re-initialization following the state-of-the-art approach~\cite{li2018deepim}, but the results show limitations in non-occluded scenes with less accuracy.

\section{Future work}

Our approach focuses on textureless printing parts. Future challenges could contain reflective or transparent objects like medical instruments to further test the boundaries of the tracking approach. Improving our 6D pose estimation algorithm with geometric priors could address robustly tracking target smaller objects with geometric ambiguities. Also, screws or similar objects could be detected more on a category-level basis to enable a more scalable approach for connecting parts of the individual assembly assets. To overcome occlusions, a multi camera setup could be useful, possibly also including \ac{ar} devices' cameras. To address more challenging assembly objects, a more robust tracking re-initialization might be necessary.

\section{Conclusion}
In this paper, we proposed a novel real-time capable graph-based tracking approach for \ac{ar}-assisted assembly tasks. \ac{gbot} tracks multiple assembly parts using kinematic links based on prior knowledge of assembly poses and combines the knowledge of 6D pose estimation with object tracking. Our tracking enables \ac{gbot} to continuously track the objects during the assembly processes in various conditions. To enable a comparison with the state-of-the-art in various scenarios, we propose the synthetic \ac{gbot} dataset and extra recorded real scenes. On this dataset, we evaluate our YOLOv8Pose, the tracking approaches SRT3D, ICG, ICG+SRT3D and \ac{gbot}. Our dataset contains five assembly assets each with three or more individual parts. The scenes of the datasets have four conditions, normal, dynamic light, motion blur, and hand occlusion. \ac{gbot} performs well in synthetic scenes with different lighting, hand occlusions and fast movement as well as real-recorded scenes. We show that tracking is more accurate compared to YOLOv8Pose and that using our dynamically created kinematic links is superior compared to individual tracking. \ac{gbot} outperforms the state-of-art tracking algorithms on the \ac{gbot} dataset which is easy to reproduce and aims as benchmark for assembly tasks. In conclusion, our approach and dataset are a promising step towards real-time and robust object tracking with \ac{ar}-guided assembly processes. %  useful to consider the kinematic structure for robust tracking of multi-state assembly parts. 

\acknowledgments{
The authors thank Philipp Stefan, Patrick Wucherer, and Matthew Lau from Medability GmbH for providing the 3D printed assembly parts. This work is funded by The German Federal Ministry of Education and Research (BMBF) with grant number 16SV8973.}

\bibliographystyle{abbrv-doi}

\bibliography{template}
\end{document}